\pdfoutput=1
\documentclass{article}
\usepackage{arxiv}
\usepackage[utf8]{inputenc} 
\usepackage[T1]{fontenc}    
\usepackage{hyperref}       
\usepackage{url}            
\usepackage{booktabs}       
\usepackage{amsfonts}       
\usepackage{nicefrac}       
\usepackage{microtype}      
\usepackage{lipsum}         
\usepackage{graphicx}
\usepackage[numbers,sort&compress]{natbib}
\usepackage{doi}
\usepackage{times} 
\usepackage{helvet} 
\usepackage{courier} 
\usepackage{graphicx}
\usepackage{caption}
\usepackage{algorithm}
\usepackage{algorithmic}
\usepackage[utf8]{inputenc}
\usepackage{newfloat}
\usepackage{listings}
\usepackage{verbatim}
\usepackage{latexsym}
\usepackage{booktabs}
\usepackage[english]{babel}
\usepackage{xspace}
\usepackage{psfrag}
\usepackage[misc]{ifsym}
\usepackage{xcolor}

\usepackage{amsmath}
\usepackage{amsfonts}
\usepackage{environ}

\usepackage{bbm}

\newenvironment{nalign}{
\begin{equation}
\begin{aligned}[b]
 }{
     \end{aligned}
     \end{equation}
     \ignorespacesafterend
 }

\newcommand{\equ}{\begin{nalign}}
\newcommand{\eque}{\end{nalign}}

\newcommand{\R}{\ensuremath{\mathbb{R}}\xspace}

\newcommand{\N}{\ensuremath{\mathbb{N}}\xspace}

\newtheorem{Definition}{Definition}[section]
\newtheorem{Lemma}{Lemma}[section]
\newtheorem{Corollar}{Corollar}[section]
\newtheorem{Theorem}{Theorem}[section]
\newtheorem{Example}{Example}[section]
\newtheorem{Note}{Remark}[section]

\newcommand{\1}{\mathbbm{1}}

\newcommand{\bea}{\begin{eqnarray}}
\newcommand{\ea}{\end{eqnarray}}
\newcommand{\be}{\begin{equation}}
\newcommand{\ee}{\end{equation}}

\newcommand{\Def}{\begin{Definition}}
\newcommand{\Defa}{\end{Definition}}
\newcommand{\Lem}{\begin{Lemma}}
\newcommand{\Lema}{\end{Lemma}}
\newcommand{\Cor}{\begin{Corollar}}
\newcommand{\Cora}{\end{Corollar}}

\newcommand{\Thea}{\begin{Theorem}}
\newcommand{\Thee}{\end{Theorem}}
\newcommand{\Ex}{\begin{Example}}
\newcommand{\Exa}{\end{Example}}
\newcommand{\Nb}{\begin{Note}}
\newcommand{\Ne}{\end{Note}}

\newcommand{\ala}{\begin{nalign}}
\newcommand{\ale}{\end{nalign}}

\title{Probabilistic Modelling of Operational Design Domains \\
\large A New Approach for Testing AI Systems}

\newif\ifuniqueAffiliation
\uniqueAffiliationtrue

\ifuniqueAffiliation 
\author{ \href{https://orcid.org/0000-0001-7421-6670}{Hans-Werner Wiesbrock} \\
	ITPower Solutions GmbH\\
	Kolonnenstrasse 26\\
	10829 Berlin, Germany \\
	\texttt{hans-werner.wiesbrock@itpower.de} \\
}
\else
\usepackage{authblk}

\fi

\renewcommand{\headeright}{Technical Report}
\renewcommand{\undertitle}{Technical Report}
\renewcommand{\shorttitle}{Probabilistic Modelling of ODDs}

\hypersetup{
pdftitle={Probabilistic Modelling of Operational Design Domains: A New Approach for Testing AI Systems},
pdfauthor={Hans-Werner Wiesbrock},
pdfkeywords={Testing ML-based Systems, Probabilistic Modelling, Training Data Evaluation},
}

\begin{document}
\maketitle

\begin{abstract}
    The conventional testing process quickly fails when applied to ML-based systems such as obstacle detection in vehicles: if an obstacle is not detected in a test, classical bug fixing is impossible -- and an AI system will always retain shortcomings.
    Test results can therefore only be interpreted statistically, which in turn requires test sets that are not only complete with respect to the operational design domain (ODD) of the system, but also representative of it.
    To this end, we introduce probabilistically extended ontologies (PEONs): ontologies describing the ODD, augmented with a probability distribution over the partitioning they induce.
    Instead of unmaintainable conditional probability tables, only marginal distributions and functionally described dependencies need to be specified; algorithms based on couplings and optimal transport complete this specification to a Bayesian network.
    From a PEON we derive the sampling of representative test cases, rigorous end-of-test criteria for given quality targets and significance levels, and methods for re-evaluating existing test results and for assessing the balance of training data.
    We demonstrate the practical modelling of a complex ODD using the example of automatic train operation.
\end{abstract}


\tableofcontents
\newpage

\section{Introduction}\label{section:Introduction}
Artificial intelligence (AI) is everywhere today -- as a topic in the media, as a buzzword in advertising, and above all as a hidden technology at work whenever we use our cell phones, start a web search, or activate the driver assistance in our cars.
The term AI covers a wide range of technologies.
Today it is mostly understood to mean the use of deep neural networks, known as deep learning.
This comprises generative systems such as ChatGPT or DALL-E, systems trained by reinforcement learning, such as AlphaGo, and feedforward networks.
The latter are typically used in classification and object recognition systems, and they are currently the first choice when it comes to detecting obstacles in autonomous driving.

Each of these system classes poses its own challenges for quality assurance.
What they have in common, however, is their statistical nature.
Deep learning networks infer the next word, the appropriate move, or the correct classification statistically: candidate outputs are first assigned probabilities, and the maximum likelihood solution is selected as the output.
Other variants exist, but they do not differ with respect to this statistical character.
We can never be certain that an answer is correct -- only that it is, in a well-defined sense, the most probable one.
It is therefore always to be expected that an AI will occasionally give a poor or wrong answer; its usefulness lies in being statistically better than alternative solutions.

\subsection{Testing AI?}\label{sec:TestingAI}
What does this mean for testing?
If a single test fails -- say, an image of a panda bear is classified as a gibbon, see \cite{Goodfellow.20.12.2014b} -- we cannot conclude that the AI is defective.
This and related work suggest that counterexamples can always be found.
If a misclassification occurs in only one out of 10,000 cases, we may well be satisfied.
Evaluating the functional quality of an AI system therefore requires many test cases, and only a statistical evaluation of the test results can provide the necessary criteria; individual counterexamples are not sufficient.
For testing, this means that the set of test cases must provide a valid basis for statistical evaluation.
We need enough test cases to absorb statistical outliers, and the test cases must be representative of the future application area of the system:
all important inputs must be reflected in the test cases -- and reflected in proportion to their frequency of occurrence.
While the first requirement applies equally to the testing of conventional systems, the second one is genuinely new.
Proven approaches such as combinatorial testing -- techniques designed to reduce the number of required tests -- fail to take the statistical nature of AI into account.
Without additional statistical information about the target environment, their results are not valid, as they lack any reference to the intended application domain, see \cite{OnSysTestExp}.

\subsection{Related work}
\label{section:relatedWork}
Testing safety-critical, security-critical, or mission-critical software faces the problem of determining those tests that assure the essential properties of the software and are able to unveil those failures that harm its critical functions.
Moreover, traditional software testing has major limitations with respect to the dynamics of machine learning, the sheer size of the problem domain, the underlying oracle problem, and the definition of completeness and reliable end-of-test criteria \cite{weyuker.oracle}.
The oracle problem is currently often addressed by metamorphic testing, see Murphy et al.\ \cite{murphy_improving_2008}, Segura et al.\ \cite{segura_survey_2016}, and Xie et al.\ \cite{xie_testing_2011}.
This approach increases the number of possible test cases, but does not take the statistical nature of the AI system into account.

To define end-of-test criteria and completeness, a number of proposals combine systematic testing of a Deep Neural Network (DNN) component with coverage criteria related to the structure of DNNs.
These include simple neuron coverage by Pei et al.\ \cite{pei_deepxplore:_2017}, which considers the activation of individual neurons in a network as a variant of statement coverage.
Ma et al.\ \cite{ma_deepgauge_2018} define additional coverage criteria that follow a similar logic to neuron coverage and focus on the relative strength of the activation of a neuron within its neighbourhood.
Motivated by the MC/DC criteria for traditional software, Sun et al.\ \cite{sun_structural_2019} propose an MC/DC variant for DNNs, which establishes a causal relationship between clusters of neurons, i.e.\ the features in DNNs.
The core idea is to ensure that not only the presence of a feature is tested, but also the composition of complex features from simple ones.
These whitebox tests check the structural integrity of the components and can provide indications of systematic errors.
They do not, however, address the proof of validity.

Wicker et al.\ \cite{wicker_feature-guided_2018} and Cheng et al.\ \cite{cheng_towards_2018} refer to partitions of the input space as coverage items, so that coverage measures are defined with respect to essential properties of the input data distribution.
While Wicker et al.\ discretize the input space into a set of hyper-rectangles, Cheng et al.\ assume that the input space can be partitioned along a set of weighted criteria describing the operating conditions.
Here, structural coverage criteria are combined with combinatorial approaches from conventional testing, but again the fact that AI draws statistical inferences is not taken into account.

A more recent paper \cite{Peleska.21.12.2023} also uses statistical methods to evaluate the quality of neural networks.
On the one hand, a model-agnostic approach is followed: as in Section~\ref{sec:SampleSize}, the test is modelled statistically as a Bernoulli experiment, yielding similar estimates of the required sample sizes.
Their second approach is based on a whitebox view of the model, in their case a convolutional classification network.
The preimages of the finitely many classes induce a partitioning of the input space.
With this partitioning, and using the training and validation data together with the classifications of the network, they obtain stronger estimates of the sample sizes than with their first approach.
It is unclear whether their way of partitioning the input space can be meaningfully applied to more complex recognition systems, and their evaluations rely on the training dataset.
Unlike our approach, they assume complete and representative training and validation data; bias or incompleteness are not captured.

\subsubsection{Ontologies}\label{sec:RelWorksontologies}
Ontologies have often been used successfully to model the Operational Design Domain (ODD), and thus the input domain, of autonomous agents such as automated vehicles and automated trains \cite{huang_ontology-based_2019, armand_ontology-based_2014}.
These ontologies represent the relevant contextual information about an agent's operation, such as environmental conditions, geographical features, traffic rules, and operational constraints.
This helps to define the boundaries and limitations of the automated agent's operation, to ensure that the agent stays within its operational design domain, and to provide a sound basis for evaluating test completeness.
Ontologies have by now also established themselves as a basis for systematic testing \cite{ASAMOpenSCENARIO,scenic,dissSchuldt,CoreOnt2015,Breitenstein.2020,Bogdoll.01.09.2022,Breitenstein.11.02.2021,Guneshka22,Zipfl.21.04.2023,UlHaq.2019}.
In this paper, we likewise use ontologies to test AI systems systematically.
However, we extend these concepts with probabilistic elements, since the previous approaches are incomplete when it comes to accounting for the statistical nature of AI systems.

In \cite{DecMaking} and \cite{SituationAssessment}, uncertainties of sensor data are modelled in a Bayesian approach for situation assessment and decision making.
In \cite{huang_ontology-based_2019}, this is extended with an ontology that captures different aspects of the driving scene, such as objects, events, and context.
The ontology contains logical rules and constraints that record the knowledge and experience of human drivers and experts.
Probabilistic dependencies among the various entities, conditions, and features are not captured in their models; apart from logical constraints, their selections remain uncorrelated.

\subsubsection{Bayesian Networks}\label{sec:BayesNets}
Bayesian networks (BNs) are directed acyclic graphical models \cite{Goodfellow-et-al-2016} -- or, in Judea Pearl's words, ``simply a factorization of a probability distribution'', as Koski and Noble \cite{Koski.2009} put it.
Pearl's work from 1988 laid the groundwork for probabilistic reasoning via Bayesian networks \cite{Pearl.2014}; for a modern treatment see \cite{Koski.2009}.
Subsequent research (Heckerman, Friedman, Korb \& Nicholson) focused on learning BNs from data and on combining statistical inference with expert input \cite{heckerman1995learning}.

In a Bayesian network, nodes represent sets of options.
A dependent child node contains selections whose probabilities depend on the previous selection at the parent node; the conditional probabilities for the associated selections are annotated at the nodes themselves.
This approach remains viable and manageable for smaller networks.
Its scalability, however, is inadequate for the highly complex networks that are essential for the probabilistic modelling of an ODD.

BNs are frequently integrated into ontologies to enable decision-making under uncertainty.

\subsubsection{Probabilistic Ontologies}
Traditional ontologies are limited in their ability to capture uncertainties and probabilistic dependencies, being restricted to logical relationships.
For instance, the statement ``it is warmer in summer than in winter'' expresses a probabilistic dependency that cannot be represented in a traditional ontology.
Probabilistic ontologies are a class of ontological frameworks that extend conventional ontologies by incorporating uncertainty.
This allows for more expressive and realistic knowledge representation, particularly in domains such as medicine, intelligence, and the Semantic Web.
Probabilistic ontologies integrate probability theory to express degrees of belief about axioms, relationships, and entities; they are based on Bayesian networks.
A pioneering framework that extends OWL using \textit{Multi-Entity Bayesian Networks} (MEBN) can be found in \cite{Laskey.2008}.
It allows for probabilistic reasoning within ontologies and supports uncertainty modelling in open-world settings; see also the foundational paper by Costa \& Laskey \cite{costa2005toward}.
Other languages adopt Markov Logic Networks for flexible ontology matching (Niepert et al.)\ \cite{niepert2008probabilistic}.

In a sense, we take up this line of work.
However, instead of annotating conditional probabilities at the nodes of the associated Bayesian networks, we specify only the far easier to ascertain marginal distributions of the associated selection probabilities -- where feasible, in functional form, as outlined in Section~\ref{section:PEON}.
Probabilistic dependencies between the selection nodes, where applicable, are likewise described functionally, and a complete Bayesian network is subsequently computed from the provided data.
This functional description is what makes the approach scale: adding, merging, or modifying selection ranges would entail a cascade of changes to the conditional probabilities, but these can be recomputed automatically by the underlying algorithms.
Consequently, even complex ontologies can be extended probabilistically.
Graphical modelling keeps such models readable and easy to create and maintain, even for novice users.

\subsection{Structure of this work}\label{section:Structure}
Owing to the high importance of recognition systems for autonomous driving and biometric access control, this article focuses on their quality assurance; the concepts, however, apply to more general settings.
For conventional systems, rigorous testing is the most effective means of detecting errors and forms the basis for confidence in the quality of a system.
To this end, the tests must cover the essential aspects of the system and its application domain (completeness).
As shown in Section~\ref{sec:MLTesting}, the statistical nature of machine learning (ML) systems demands more: proving the reliability of their results requires not just complete test coverage -- the tests must also be representative.
In short, the tests must be statistical.
This is in line with the prevailing quality criteria for ML systems, such as accuracy and precision, which are statistical metrics, see \cite{Geron.2019d}.
The test cases must therefore be not only complete but also representative of the application area, i.e.\ representative also in their statistical proportions.
Otherwise the system cannot be evaluated statistically: the statistics of the test results would not reflect what a customer means when 99.9\,\% accuracy in the target domain is promised, and the assessments of testers and developers could disagree even if both had done a good job.
Verifying completeness and representativity is the real challenge of testing AI systems.

Using a simple example, we first show in Section~\ref{sec:MLTesting} why statistical considerations are not just nice to have for the quality assurance of AI, but a must.

The basis of good testing is a thorough understanding of the target environment in which the system under development is to be used.
In Section~\ref{sec:ODD}, we use the example of obstacle detection in the track area to show what a formalized description of a target environment can look like.
It is based on the procedure in \cite{dissSchuldt}.
To ensure the completeness of the test cases, we use ontologies to describe all possible entities, their characteristics and relations, see Section~\ref{sec:OntTypeTracks}.
This step is very similar to partition testing and the classification tree method of the conventional test process, see \cite{GrimmGrocht,dunietz}.
What is new for the testing of AI systems, however, is the requirement that the test cases be statistically representative.
To achieve this, it is not enough to collect as many test cases as possible or to make a clever -- e.g.\ combinatorial -- selection; we must ensure that critical as well as normal situations enter the test set according to their probability of occurrence.
To this end, we extend the concept of ontologies and enrich it with occurrence probabilities, arriving at a PEON, see Section~\ref{sec:NeedPEON}.

These concepts can be used profitably in other contexts as well; Section~\ref{sec:UseCases} presents a number of further use cases.
At first sight, it does not seem practical to model a very complex, open ODD statistically.
Section~\ref{sec:PEONModelling} therefore shows in detail how such a complex model can be designed and developed in a manageable way.
The mathematical principles used along the way are collected in the appendix, Section~\ref{sec:Appendix}.

\textit{This publication makes the following contributions:}
\begin{itemize}
  \item We introduce the new concept of \textit{probabilistically extended ontologies} (PEONs). Instead of maintaining non-scalable conditional probability tables at the nodes of a network for the selection of the associated characteristics, we specify only the much simpler marginal distributions; probabilistic dependencies are described functionally. Suitable algorithms complete such a specification to a Bayesian network.
  \item As a basis for testing AI systems under their statistical interpretation, \textit{probabilistically extended ontologies} allow us to formulate a \textit{uniformity hypothesis} that serves as a termination criterion for the refinement of ontologies.
  \item From a PEON as a statistical model of an ODD, we derive end-of-test criteria for testing ML-based systems for given quality criteria and predetermined significance.
  \item We integrate these concepts into a systematic testing process that can be performed independently of development, using the black-box method.
\end{itemize}
If developers and testers have done their work well and independently of each other, this process is able to reproduce the statistical quality criteria offered by the developer and required by the user.

The concept of \textit{probabilistically extended ontologies} can moreover serve as a requirements document for ML systems, describing how the customer conceives of the ODD and which quality criteria are to be fulfilled.

\section{Testing ML-based Systems}\label{sec:MLTesting}
For conventional systems, testing is the most effective means of detecting software errors and forms the basis for confidence in the quality of the software.
What does this look like for a simple AI system that uses images to distinguish between dogs and cats?
\begin{figure}[htbp]
  \centering
  \includegraphics[width=0.6\textwidth]{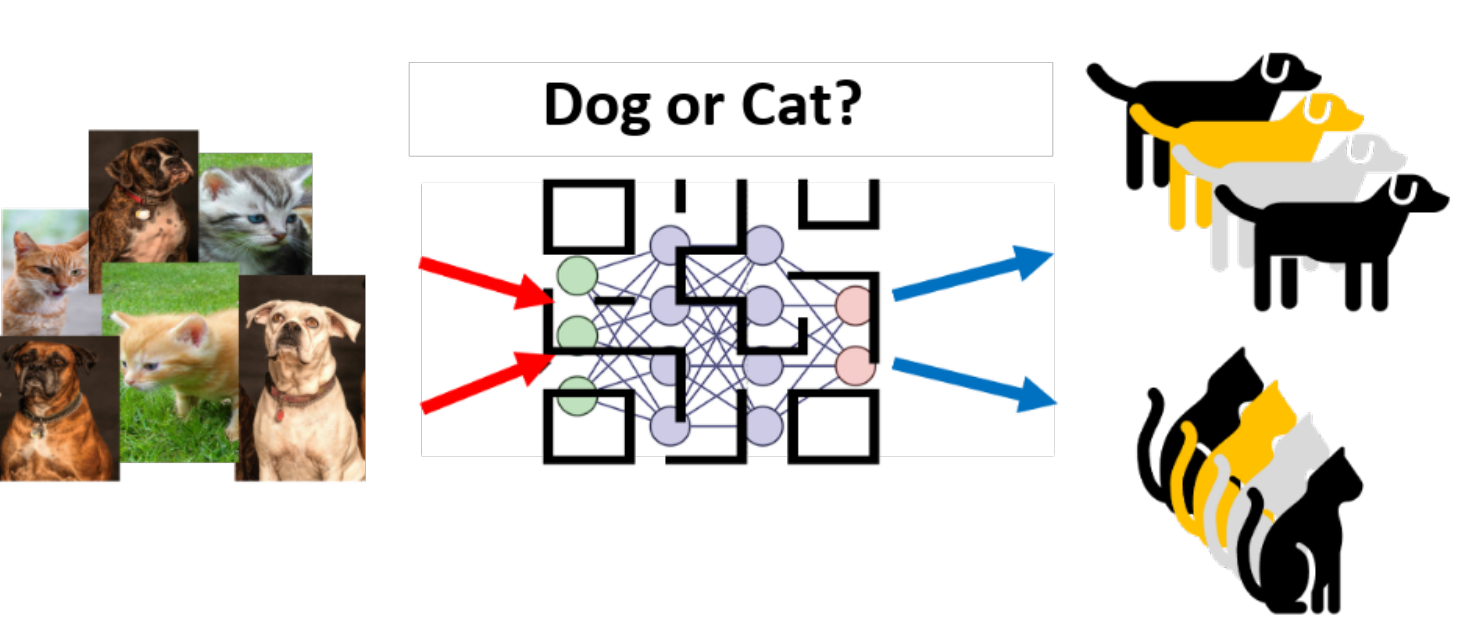}
  \caption{A simple AI system distinguishing pictures of dogs and cats}
  \label{fig:CatDogAI}
\end{figure}
Can we test AI systems like conventional systems?
The problems begin as soon as the test contains a picture of a cat that our system incorrectly recognizes as a dog.
We cannot simply locate the place in the code where the network took a wrong turn and fix it.
Deep neural networks have a multitude of parameters whose individual meanings we do not know.
The network has been trained on numerous examples, algorithmically adjusting these parameters so that image pixels are assigned to dogs or cats as accurately as possible.
This ensures that as many images as possible \emph{from the training data} are classified correctly.
We cannot expect all images to be classified correctly.
What does this mean?
It means that images will be sorted incorrectly time and again -- and that conventional bug fixing of our system is not possible.
And what does this mean for the test?

\subsection{Testing AI -- and statistics!}\label{sec:DogCatEx}
A recognition system is good if it classifies as many inputs as possible correctly.
Quantified, this means that a certain percentage of the test data is classified correctly.
From the results of all tests we calculate a mean value, the accuracy of the system, and use this number to compare quality.
Similar measures are precision, sensitivity, etc.; the problems discussed below are the same for all of them.
Suppose we want to deploy the system in Germany and guarantee the customer that our AI system will classify the camera images correctly in 90\,\% of all cases.
The customer commissions two experienced testers to check this quality.
One selects 1{,}000 test images, 100 of which depict dogs and 900 cats; the other also selects 1{,}000 test images, but with 900 dog images and 100 cat images.
Since the system has been on the market for a long time, we know that cats are recognized correctly with 95\,\% certainty, while dogs are recognized correctly only 85\,\% of the time.
What will the testers report?
The first tester will measure an average of
$$\frac{100 \cdot 85 + 900 \cdot 95}{1000} \  \% = 94\,\%,$$
the other
$$\frac{900 \cdot 85 + 100 \cdot 95}{1000}\, \% = 86\,\%.$$
Which of them is correct?
The ratio of dogs to cats in Germany is approximately 3:4.
If we randomly take 1{,}000 photos of dogs or cats in Germany, we will capture approximately 571 cats and 429 dogs, and our AI will correctly recognize a cat about 543 times and a dog about 365 times -- an accuracy of 90.7\,\%.
\begin{table}[h!]
  \begin{center}
    \begin{tabular}{l|c|r|r}
      \textbf{Test data} & \textbf{Dogs} & \textbf{Cats} & \textbf{Accuracy} \\
      \hline
      Tester 1           & 100           & 900           & 94 \%             \\
      Tester 2           & 900           & 100           & 86 \%             \\
      Reality            & 429           & 571           & 90.7 \%           \\
    \end{tabular}
    \caption{Test results, depending on the distribution of the test data}
    \label{tab:DogCat}
  \end{center}
\end{table}
Could the testers have arrived at this result?
Yes -- if they had weighted their results according to the actual ratio of cats to dogs in Germany.

These simple observations apply in general.
If the quality of a system is to be evaluated statistically, care must be taken that the sample is representative, i.e.\ that a sufficient number of samples has been drawn randomly from the population -- or that an accurate statistical model of the distribution of the population is available and used to weight the results.
This applies to accuracy, precision, and all other statistical criteria.

\subsection{The Need for Statistical Testing}\label{sec:NeedStatTest}
ML-based systems are currently employed in a broad spectrum of domains, with a growing trend towards safety-critical sectors, and the range of applications keeps expanding rapidly.
At the same time, a growing number of experts caution about the risks and uncertainties associated with the uncontrolled and hasty deployment of AI systems, see \cite{OpenLetter}.

There is thus an increasing need for methodologies and protocols to evaluate the functionality and the quality attributes of these systems.
Many approaches for assessing the quality of an AI system exist today.
Some evaluate its performance or its resilience to disturbances, employing techniques already established in conventional testing.
The spectrum includes formal verification, simulation approaches, classical testing, and new methods from the field of explainable artificial intelligence (XAI); see \cite{Albarghouthi.21.09.2021,Jackson.29.03.2021,VasuSingh.} for an overview, and \cite{Hoyer.30.07.2019,Guidotti.06.02.2018b} for analysis methods in the context of XAI.
These methods furnish evidence regarding the robustness and reliability of ML models and ML-based systems.
Further testing methodologies for ML systems are surveyed in the literature, see \cite{Marselis.b,Zhang.20.06.2019b}.

All these approaches and procedures are undoubtedly fruitful.
However, they miss the statistical nature of AI systems.
Most importantly, they do not establish the quality criteria necessary for a validation in specific application domains.
Furthermore, today's dynamic testing approaches often do not provide the end-of-test criteria necessary for a comprehensive evaluation -- and sometimes it is not even clear when a system should be rejected as unqualified.

Our claim is that the critical question of functional quality can only be answered by statistical analysis.
This follows from the very way these systems are built.
An ML-based recognition system does not implement rules that derive properties from pixel data.
Instead, it is presented with numerous labeled training examples, and optimization algorithms adjust its parameters so that as much of the training data as possible is subsequently classified correctly -- conceptually, a highly intricate regression model.
Such systems learn inductively, from examples; they do not operate deductively, according to predefined rules.
This also means that we cannot fix bugs in AI systems:
if a test finds a counterexample, we cannot tweak the parameters so that this particular ``bug'' disappears, because we simply do not know at which branch the network took a wrong turn.
The example in Section~\ref{sec:DogCatEx} points in the same direction:
the notion of a cat is not derived from a set of rules; the system estimates probabilities for the correctness of the candidate answers and selects the most probable one (maximum likelihood, or variants thereof).
Since the system draws statistical conclusions by design, the evaluation of these conclusions is inherently restricted to statistical methods.
For the test, this means that the set of test cases must be designed so that its evaluation provides a valid basis for statistical quality criteria:
the test data must be complete and representative of the target environment.

Some of the proposed test procedures suffer from yet another problem.
In traditional system development, tests are defined separately from design and development activities and are then executed on the system under test (SUT) to verify correct implementation.
Agreed-upon end-of-test criteria indicate how thoroughly the system has been tested and determine when the testing process can be finished.
In many of the approaches above, no such criteria can be formulated; in others, such as the metamorphic ones, see \cite{murphy_improving_2008}, Segura et al.\ \cite{segura_survey_2016}, and Xie et al.\ \cite{xie_testing_2011}, one relies on the quality of the training data and thereby loses the independence from development.
Biased training data cannot be detected there.

It is not only the test evaluation that must be based on balanced and complete data in order to produce trustworthy results.
Such data is equally essential for training.
Supervised learning requires large amounts of training data, and this data must be complete and representative with respect to the problem domain -- a non-trivial endeavor.
This is all the more significant since imbalanced training data leads to systematic distortions, see \cite{Chen.2024, Gao.13.02.2025}.
In this sense, the requirements for development-independent test data and for training data coincide.
The concept of a PEON, see Section~\ref{section:PEON}, can systematically support the fulfillment of these requirements in both cases, see Section~\ref{sec:DataEvaluation}.

\section{Operational Design Domain}\label{sec:ODD}
The target environment for which an AI system has been developed is referred to as its Operational Design Domain (ODD).
While the target environment is characterized by openness and high complexity, the ODD fixes the boundaries within which the system is required to work properly:
\textit{operating conditions under which a given automated driving system \dots\ or feature thereof is specifically designed to function, including, but not limited to, environmental, geographical, and time-of-day restrictions, and/or the requisite presence or absence of certain traffic or roadway characteristics},
see \cite{ISO2020}.
ODDs are prominently used in autonomous driving.

It is impossible to enumerate every conceivable driving situation.
For an effective quality assurance process, the first requirement on the test cases is therefore that they cover all essential scenarios delineated in the original design.
This is evidently ambitious, but it applies equally to the conventional testing process, and a range of methodologies has been developed to address it systematically.
One example is the classification tree method, which is employed extensively in the automotive sector \cite{GrimmGrocht}; it supports covering the input space as comprehensively as possible.
The high complexity of an ODD, however, calls for more general concepts than tree-like structures.
Ontologies have become established for this purpose in recent years, see \cite{Staab.2009, huang_ontology-based_2019,armand_ontology-based_2014}.
They allow the various entities, their logical relationships, and their properties, as they may occur in an ODD, to be captured formally.
A prominent example of the use of ontologies for testing can be found in \cite{ASAMOpenXOntology}, see also the references therein.
They have become an increasingly popular means of describing the domain of ML-based systems, see \cite{ASAMOpenSCENARIO,scenic,dissSchuldt,CoreOnt2015,Breitenstein.2020,Bogdoll.01.09.2022,Breitenstein.11.02.2021,Guneshka22,Zipfl.21.04.2023,UlHaq.2019}.

\subsection{A simplified version for Automatic Train Operation (ATO)}\label{sec:SimpVersATO}
\begin{figure}[ht]
  \centering
  \begin{minipage}{0.35\textwidth}
    \centering
    \includegraphics[width=0.8\textwidth]{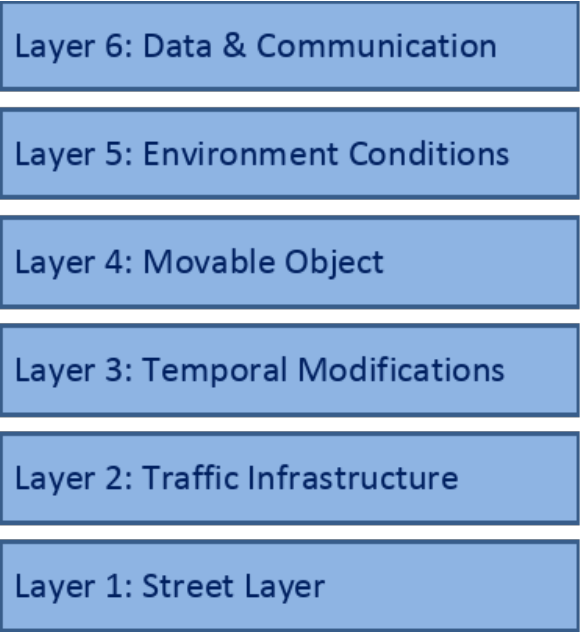}
    \caption{Six Layer Model}
    \label{fig:SixLayer}
  \end{minipage}\hfill
  \begin{minipage}{0.65\textwidth}
    \centering
    \includegraphics[width=1.0\textwidth]{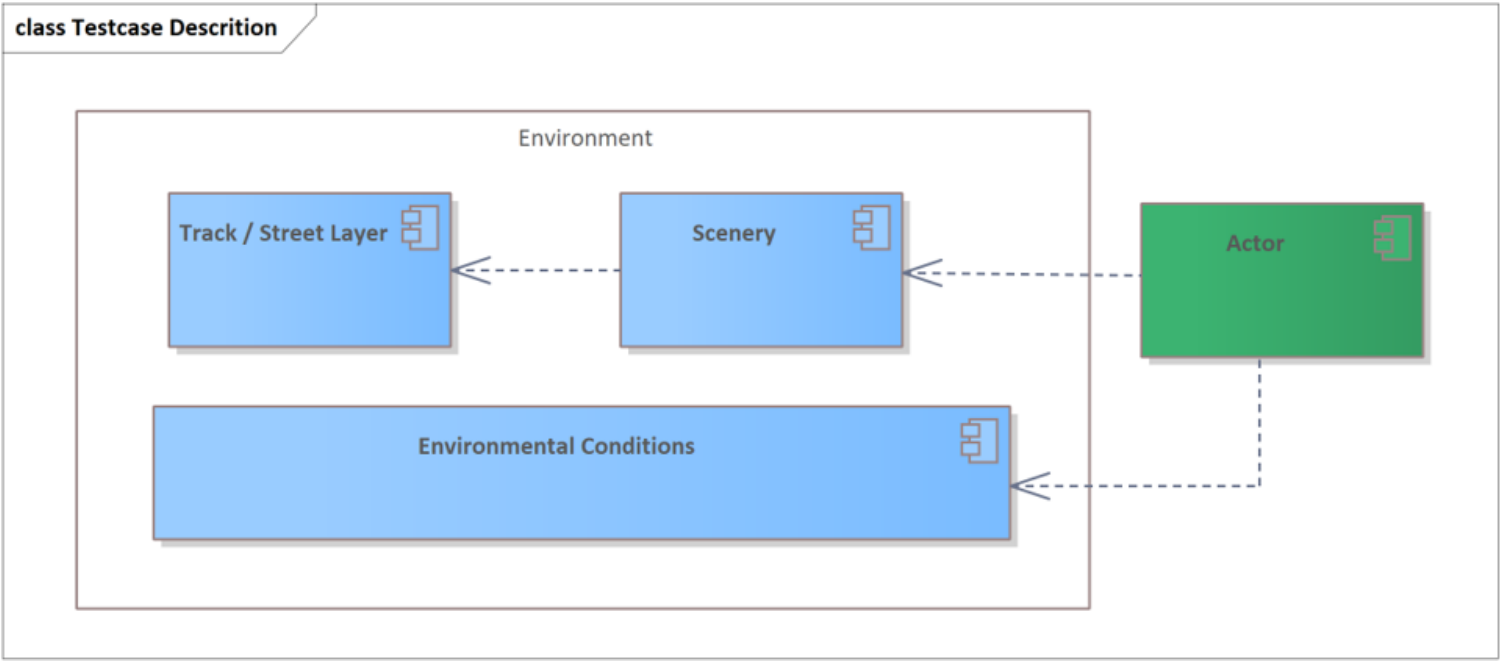}
    \caption{Simplified probabilistic model tailored to the railway domain}
    \label{fig:SixLayerProbModelLayer}
  \end{minipage}
\end{figure}

Given the high complexity of an ODD for autonomous driving, it makes sense to structure its description with an ontology.
The first thing to note is that a driving scene is essentially determined by the possible traffic routes -- infrastructure, signals, traffic signs, etc.
Together they delimit the movements of the actors.
The various entities that play a role are commonly organized into layers:
in \cite{ASAMOpenXOntology,dissSchuldt,ScholtesEatAl,Weber.2019}, a 6-layer model is proposed within the ASAM standard for autonomous driving, see fig.~\ref{fig:SixLayer}.

In the railway sector, the clearance gauge -- the area close to the rail track -- plays the decisive role in autonomous driving.
This region is easier to describe, and we can simplify the 6-layer model somewhat.
Furthermore, since we want to model the ODD probabilistically, we do not structure it into layers but into areas that may depend on each other probabilistically -- actors on environmental conditions, for instance, or sceneries on tracks.
To give a simple example, we will not expect skiers in hot, dry regions; and driving situations, i.e.\ scenarios, depend to a certain degree on the route structure.
We therefore describe these areas from the outset by a probabilistic graph, see Appendix~\ref{sec:StructProb}.
Slimmed down and tailored to the railway domain, we use fig.~\ref{fig:SixLayerProbModelLayer}.
In the following, only some of the sub-areas will be modelled by way of example: the environment with the rail line and its signals, streets, and some points of interest, as well as a few actors, such as a train driving on the track and vehicles on a street.
We will also model simple environmental conditions such as the weather.
The aim is to show the interested reader how diverse entities, with their characteristics and relations, can be modelled.

\subsubsection{An Ontology for Tracks and Signals}\label{sec:OntTypeTracks}
The clearance profile is central to train traffic: everything within it must be localized and categorized.
As in the six-layer model, we start by demonstrating what an ontology for the track layer and infrastructure could look like.

\begin{figure}[ht]
  \centering
  \begin{minipage}{0.45\textwidth}
    \centering
    \includegraphics[width=\textwidth]{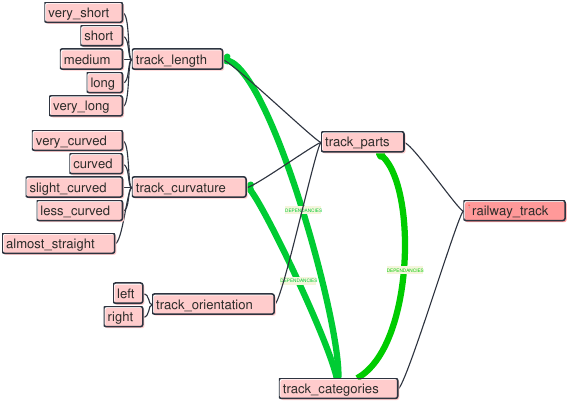}
    \caption{railway track}
    \label{fig:RegionTrack}
  \end{minipage}\hfill
  \begin{minipage}{0.45\textwidth}
    \centering
    \includegraphics[width=\textwidth]{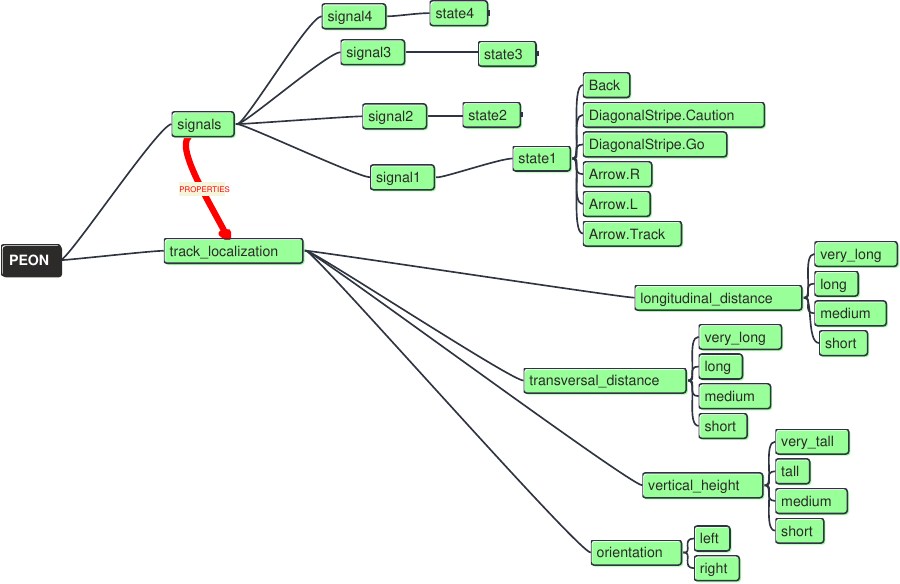}
    \caption{Signals, localized along the track}
    \label{fig:SignalsLoc}
  \end{minipage}
\end{figure}

There are regional and high-speed lines, freight lines, etc.
They are characterized by the different curve behaviour to be expected on them on average -- there will be no tight curves on high-speed lines, for example.
To describe different tracks, we therefore start by selecting the type, i.e.\ the track category, see fig.~\ref{fig:RegionTrack}.

Instead of describing an entire track section at once, we assemble it from individual curve parts.
Each part has a length, a radius of curvature, and an orientation to the left or right; in the case of rail, the parts can be connected by clothoid pieces.
We thus describe a track by a list of $N$ track parts, each consisting of a length, a curvature, and an orientation.
Together with the clothoid connections, this unambiguously specifies a track course, where we neglect any vertical profile in the following.
Regional rail lines generally have many more curves than high-speed rail lines. Depending on the type of line, a route will therefore be described by more or fewer segments, with sharper or gentler curves.
In anticipation of the explanations that follow (see \ref{sec:EntAtt}), it should be noted here that such dependencies are modeled on the left using \textit{DEPENDANCIES}; see Fig.~\ref{fig:RegionTrack}.
(Strictly speaking, we are already extending the concept of an ontology here: ontologies were originally designed to capture logical dependencies, not looser dependencies such as correlations.)

We want to characterize the entities and their characteristics by finite, discrete descriptions.
To this end, we partition the possible part lengths into suitable sub-ranges, and likewise the radii of curvature, named as classes \textit{very short, short,\dots}\ and \textit{very curved, curved,\dots}.
Selecting the number $N$ of parts and a list of $N$ tuples of length and curvature classes together with orientations then defines an \textit{abstract track}.
To obtain a concrete one, we select reals uniformly from the underlying intervals and arrive at a list of $N$ tuples of concrete lengths, curvature radii, and left-right orientations of the track parts, which, connected by clothoids, uniquely specifies the track.
A similar procedure is already in use in test specification via partitioning and the classification tree method.

Later on, the track will be parameterized by arc length.
Being central to the railway domain, this reference will be used to locate signals, points of interest, and so on along the route.
The position of a signal along the track, for example, is then uniquely specified by its distance along the track (longitudinal), its transversal distance to the left or right, and its height, see fig.~\ref{fig:SignalsLoc}.

In this simple example, we consider four different signals, each of which can be in different states.
All signals have a location.
We describe this by signals with states, where each signal carries a localization, represented by the red \textit{PROPERTIES} links in fig.~\ref{fig:SignalsLoc}, (see \ref{sec:EntAtt}).
Once more, behind the leaves of the subtree \textit{localization} lie ranges of distances -- classes \textit{very long, long, medium, short} forming a partition of the distance domains.

We now form equivalence classes of situations involving signals.
Consider a situation with a signal.
We group the situations by the type of the signal, refined by its state.
Next, we look at the longitudinal, transversal, and vertical distances relative to the track and, owing to the partitioning of the distance domains, classify the situation by the distance classes \textit{very long, long, medium, short} it belongs to.
One easily sees that this grouping is canonically defined by the classification of the properties as given by the ontology.
The informed reader will be reminded of the classification tree, see \cite{GrimmGrocht}.
In short: the ontology of signals defines, in a canonical way, an equivalence relation on the set of all situations.

We can proceed similarly with the tracks in a situation.
Given a concrete track, we can ask: How many parts does it have? In which classes do their lengths and curvatures lie? Which orientations do they have?
We then group the situations according to the answers.
It is easy to see that this grouping stems from an equivalence relation:
given the classes \textit{very short, short,\dots}, \textit{very curved, curved,\dots}, combined with the orientations, two situations are called equivalent iff their characteristics agree with respect to the entities and property classes defined in the ontology.
So, with a track in hand, we can map it to its equivalence class.

Even though these ontologies are very simple, they already allow the description of a wide variety of routes and signals along the rail.
Later, we will describe other objects relevant for the clearance gauge in a similar way.

\subsection{The Need for Extending Ontologies, and Our Approach}\label{sec:NeedPEON}
Before we develop further ontologies to capture the ODD for automated train operation, let us recall our actual goal.
Ontologies have primarily been used to identify corner cases, see \cite{Bogdoll.01.09.2022}.
This works well, but we want to use ontologies more pervasively, in a wider range of applications:
we want to extend them to cover all of the ODD, and to extend them into a statistical model.
As motivated in Section~\ref{sec:NeedStatTest}, we need test cases that not only cover an ODD completely but are also representative, i.e.\ reflect the statistical relationships in the ODD.
This means we have to go beyond previous approaches and add probabilistic data.

In general, we will only employ ontologies that carry a canonical equivalence relation in the sense of Section~\ref{sec:OntTypeTracks}.
We call it the \textbf{equivalence relation associated with the ontology} when it is clear from the context how the classes are defined.
Equivalently, such an ontology defines, in a canonical way, a partition of the ODD it comprises: Does this entity occur in the situation, and if so, in which embodiment? And so on.
We can thus unambiguously map a specific situation of the ODD to the entities present in it, together with their characteristics.
And if the ontology captures the ODD, all these possibilities and combinations together describe the ODD.
We call this subdivision the \textbf{partition associated with the ontology}.
Equivalence relations lead to unique partitions, and partitions in turn define unique equivalence relations; we use both descriptions synonymously.

As noted above, a choice of the number of track parts and of the property classes may be viewed as an abstract track, and, similarly, a signal with specified state and localization classes as an abstract signal.
For an ontology whose associated equivalence relation, or partition, captures the ODD, we read an \textbf{equivalence class as an abstract test case} for the ODD.
Again, the informed reader will be reminded of the classification tree method, see \cite{GrimmGrocht}.

\begin{figure}[ht]
  \centering
  \begin{minipage}{0.45\textwidth}
    \centering
    \includegraphics[width=\textwidth]{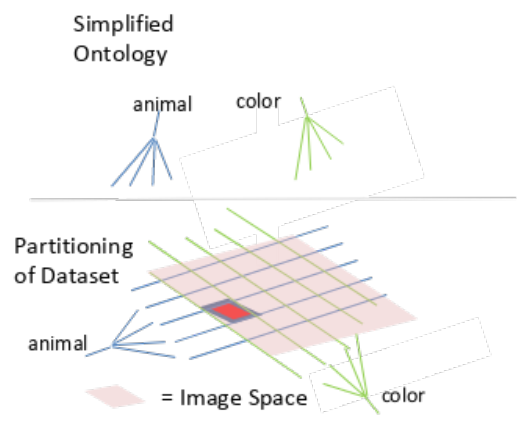}
    \caption{The partitioning associated with an ontology}
    \label{fig:OntologyPartitioning}
  \end{minipage}\hfill
  \begin{minipage}{0.45\textwidth}
    \centering
    \includegraphics[width=\textwidth]{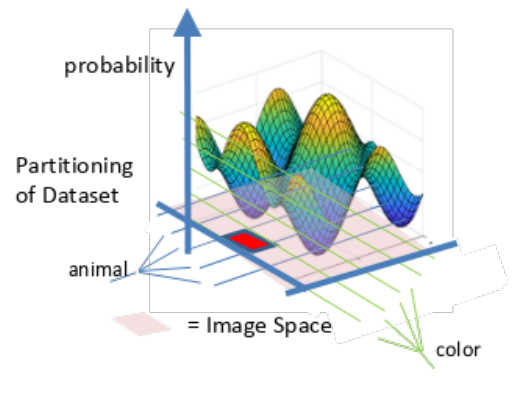}
    \caption{Probability distribution over the partition space}
    \label{fig:SketchPEON}
  \end{minipage}
\end{figure}

In our example of an AI that is supposed to distinguish between cats and dogs in images, we consider dogs and cats as the possible entities, see Section~\ref{sec:DogCatEx}.
They can have long or short snouts and, in simplified terms, be brown/black or yellowish.
Based on these descriptions, we divide the images into classes -- \textit{\{dog, long snout, brown\}}, \textit{\{cat, short snout, yellowish\}}, etc.\ -- and in this way obtain a partition of all images, see the left of fig.~\ref{fig:OntologyPartitioning}.
The selection \textit{\{dog, brown\}} would be an abstract test case for the setting of Section~\ref{sec:DogCatEx}.
If we now apply combinatorial testing, we select approximately the same number of cats with long and with short snouts, and of dogs that are yellowish and black.
But this does not correspond to the distribution of the appearance of dogs and cats in Germany -- our test results will probably not be representative!
We were able to demonstrate this experimentally with a simple example, see \cite{OnSysTestExp}.
What is missing are the occurrence probabilities of the image classes, see fig.~\ref{fig:SketchPEON}:
How often are dogs in Germany black or yellowish? How often do they have long or short snouts?
Only with these figures can we make valid statistical quality statements.
So we consult the relevant statistics and learn that, restricting our view to dogs and cats, a dog is to be seen with probability $p=\frac{3}{7}$ in Germany, a cat with $p=\frac{4}{7}$.
Furthermore, we read in the statistics that perhaps $p=\frac{1}{3}$ of all dogs are black.
Then the probability of capturing a black dog in a picture is $p=\frac{3}{7}\cdot \frac{1}{3} = \frac{1}{7}$, and we assign the occurrence probability $p=\frac{1}{7}$ to the abstract test case \textit{\{dog, black\}}.
We proceed analogously for the other classes.
Then, by construction, our test statistics should be reliable.

In order to define a statistical model for the test, see Section~\ref{sec:NeedStatTest}, our approach is thus to define a probability distribution over the set of equivalence classes -- over the set of all abstract test cases, see fig.~\ref{fig:SketchPEON}.
That is, the ontology is augmented with occurrence probabilities for each partition class.
These structures will serve us as a replacement for category or classification trees, with abstract test cases (equivalence classes) sampled according to their probability instead of being selected combinatorially, see \cite{GrimmGrocht}.

\subsection{PEON: probabilistically extended ontologies}\label{section:PEON}
We start with an ontology that describes our ODD as completely as possible.
We now want to define the occurrence probabilities for the classes associated with it -- which, at first sight, seems an unrealistic undertaking.
Thinking of all the possible routes and signals described by the ontologies of Section~\ref{sec:OntTypeTracks}, figures \ref{fig:RegionTrack} and \ref{fig:SignalsLoc}, it might still work in this case.
But there are many more signals, switches, and crossings -- how can we capture all these combinations together with their probabilities of occurrence?
And had we made this effort once, how could we maintain the result?
On the one hand, our experiments in \cite{OnSysTestExp} show that even rough statistical models are significantly better than combinatorial ones.
On the other hand, Section~\ref{sec:PEONModelling} will develop techniques by which this modelling can be carried out and maintained in practice.

Before we move on, we make a few general remarks, starting with the definition.

\Def (Preliminary definition of a PEON)
A \textit{probabilistically extended ontology} (PEON) is an ontology augmented with a probability distribution on the associated partitioning, reflecting the occurrence of the partition classes in the ODD.
\Defa

Formulated mathematically:
let $\mathcal{P}$ be the set of partition classes of the ontology, and let $\Sigma_{\mathcal{P}}$ be the $\sigma$-algebra of the partitioning, i.e.\ the set of all subsets of $\mathcal{P}$.
Then a PEON is given by a probability distribution $p$ with respect to this $\sigma$-algebra.
(Later, see Section~\ref{sec:PEONModelling}, we will describe such distributions using Bayesian networks. Since ontologies may contain cyclic subgraphs, this approach requires the restriction to tree- or forest-like graphs.)

To rephrase: a \textit{probabilistically extended ontology} provides us with a finite probabilistic model of the ODD.
The interpretation goes as follows.
Let a PEON be given for an ODD, and let the ontology encompass the ODD completely.
Then each situation in the ODD can be unambiguously assigned the partition class -- its equivalence class -- to which it belongs, and the probability of occurrence of the situation in the ODD is modelled by the probability measure of its equivalence class.
We thereby obtain a discrete statistical model of the ODD.
Sampling test cases accordingly, we should obtain valuable test results, consistent with our expectations about the functioning of the AI system in the ODD.

\vspace{0.2cm}
\noindent \textbf{Test evaluation.}
When the test cases are executed on the test object, we can compute various statistics of the results against the probability model given by the \textit{probabilistically extended ontology}.

If we return to the example of recognizing dogs and cats, Section~\ref{sec:DogCatEx}, we observe: a cat can be spotted, striped, plain, black, dark gray, \dots\ -- and further subdivisions lead to more and more classes, making our model ever finer.
How far should we go?
The same question already arises when testing conventional systems, and there the uniformity hypothesis of partition testing answers it as follows:
the test space must be subdivided to such an extent that two different test cases from the same partition class are equally suitable for detecting a given error.
Due to the statistical nature of our use case, we extend this requirement.

\vspace{0.2cm}
\noindent\textbf{Uniformity hypothesis for a PEON:}\label{sec:UniHyp}\footnote{\scriptsize I would like to express my sincere thanks to S. Sadeghipour, who pointed out this connection to me}
the concept of a PEON suggests a criterion for which entities and refined properties should be modelled in the ontology.
The \textbf{uniformity hypothesis for a PEON} states that the refinement of an (ontology) partition should continue until the failure rate of the test item is likely to be equal across samples from the same partition class.

This hypothesis allows us to derive a statistical model of the test performance from the PEON, and thereby a more precise evaluation of the test, see \cite{ProbExtOnt,OutlinePEON}.
Admittedly, with realistic capacities it is hardly possible to define such a refinement systematically.
From a statistical point of view, however, any approximation of the underlying population is better than none, and so a PEON should simply be modelled at as fine a granularity as possible.
Our experiments \cite{OnSysTestExp} confirm this.

Now that we have a PEON, a discrete statistical model of the ODD, what can we say about the testing process?

\vspace{0.2cm}
\noindent \textbf{Sampling, sample size, and test completion criteria:} \label{sec:SampleSize}
given an accuracy target with specified significance that is to be established by testing, the probability model helps us sample an appropriate set of abstract test cases.
If these abstract test cases can be enriched to feasible, i.e.\ executable, test cases, we can apply reliable statistics to the test results with respect to the probability model of the ODD.
In particular, given numerous test cases distributed randomly according to the model, we obtain significant estimates of the accuracy of our system under test.
Classical statistics tells us how many samples are needed to measure the mean of a distribution empirically up to a given significance level.
This provides the basis for defining rigorous end-of-test criteria, both qualitatively and quantitatively.

In its simplest form, the test may be regarded as a Bernoulli experiment.
A test case is selected from the totality of all possible test cases; during execution, a series of checks ascertains the functionality of the system in this particular case, and the outcome is designated as either \textit{passed} or \textit{failed}, with failure probability $p$.
This failure probability is directly tied to the accuracy to be verified by the testing process: accuracy $= 1-p$.

Let the required quality, i.e.\ the accuracy level, be $1-p_0$:
we want to be able to rely on the system working correctly with probability at least $1-p_0$.
The test is intended to prove this quality, so we must show that the actual failure probability $p$ is less than or equal to $p_0$.

We can attempt to establish this statistically through the test.
Assume $p_0 \ll 1$, a premise that undoubtedly holds for safety-related applications, and let us estimate $p$ with significance $\alpha$.
Let $z_{1-\alpha}$ denote the quantile of the standard normal distribution for significance level $\alpha$.
We approximate the Bernoulli distribution by the Poisson distribution with parameter $p$.
Then, for the empirical mean $\hat{p}$ at large sample size $n$ (the number of test cases), we obtain the confidence interval $\hat{p}\pm z_{1-\alpha/2}\sqrt{\frac{\hat{p}}{n}}$, and if we bound the interval half-width by $d$:
\bea
n\geq \Big(\frac{z_{1-\alpha/2}}{d}\Big)^2 \hat{p}
\ea
see \cite{SheinChungChow.2017}.
That is, depending on the significance and the allowed deviation $d$, we obtain an estimate of the number of randomly chosen test cases needed for a confident estimate of the mean failure probability $p$.
If the resulting estimate satisfies $\hat p < p_0 - d$, the test has provided evidence of the required quality within the specified confidence.

Alternatively, we can make more specific use of the fact that we are dealing with a Bernoulli experiment.
Suppose the true failure probability is expected to be around $p_0/2$, and the objective is to support this statistically.
Two errors may be committed here: the hypothesis $p\approx p_0/2$ may be true but rejected on the sample (type~1 error), or false but accepted (type~2 error).
We limit both by specifying $\alpha$, the statistical significance level (type~1), and the type~2 error rate $\beta$.
Power analysis, see \cite{SheinChungChow.2017}, then yields an estimate of the required number $n$ of test cases:
with $z_{1-\alpha}$ the quantile of the standard normal distribution for significance level $\alpha$, and $z_{1-\beta}$ the corresponding quantile for $\beta$, we obtain
\bea
n \geq \frac{\big(z_{1-\alpha}\sqrt{p(1-p)}+z_{1-\beta}\sqrt{p_0(1-p_0)}\big)^2}{(p-p_0)^2}
\ea

In both cases, we obtain an estimate of the number of test cases that must be executed to establish the desired quality.

What does this mean for our test process?
Given a PEON, we obtain clearly defined end-of-test criteria.
However, obtaining the number of labeled test data required for a given significance is usually very time-consuming; the approach is therefore best used in combination with the generation of test simulations, see \cite{TDGenRailwayDomain} and Section \ref{sec:SysTP}.

Often, a number of test cases already exist but are not distributed according to the PEON.
We can then relate them to the partitioning of the PEON: we assign to each test case the equivalence class of the PEON to which it belongs.
For each class we then have its relative frequency in the set of all test cases as well as its modelled probability of occurrence, and we can use these figures to re-weight the test results so as to relate them to the modelled PEON distribution. (The implementation of more precise labelling, such as that which is required for the PEON partition, is a more straightforward process than the development of entirely new representative test data.
It should be noted, however, that this approach also changes the significances, i.e. p-values.)

Nevertheless, at first sight, modelling such a distribution of occurrence probabilities seems impossible, at least in a manageable way.
We have developed a number of techniques, language concepts, and algorithms to achieve this goal for a wide range of applications; they are presented in Section~\ref{sec:PEONModelling}.
Before that, we describe a few further use cases in which these concepts provide systematic support.

\section{Use Cases}\label{sec:UseCases}
The concept of a PEON was developed to account for the statistical nature of AI systems in testing -- to design tests in such a way that their results are statistically valid.
Underlying this is the idea that a statement such as ``the system detects a person on the tracks with a sensitivity of 99.9999\,\%'' refers to the expected operating environment: a person \emph{in that environment} will fail to be detected in at most $10^{-6}$ of all cases.
Beyond testing, however, the concept supports a number of further activities, which we sketch in this section.

\subsection{Risk Analysis}
A preceding risk analysis may have shown that, over a given number of operating hours, a certain number of people are to be expected on the tracks on average -- and the safety requirements then state the percentage at which the system must recognize them.
Let us take a closer look at this.

What can happen during system operation?
Possible malfunctions that could lead to damage are identified, for example by means of a fault tree analysis, and the possible severity of their consequences is estimated.
Risk is then defined as the product of the possible severity and the probability of the failure:

\textsf{
Risk is a probability or threat of damage, injury, liability, loss, or any
other negative occurrence that is caused by external or internal vulnerabilities,
and that may be avoided through preventive action.} \cite{Tzanakakis.2021}, p.~7

This means that we have to estimate the probability of malfunctions occurring during operation.
If the causes of failure are all independent of each other, they can be estimated independently, and the risks can be calculated from them.
Otherwise, Markov models, for example, are commonly used to compute the occurrence probabilities of dependent causes of failure; these models record logical dependencies.

Consider now the example of lightning and storm damage.
Lightning can occur without a storm, and storms can occur without lightning -- but we also frequently have thunderstorms, which can cause greater damage.
This dependency is probabilistic rather than logical, and it is often not taken into account in such models.
Here, too, a PEON can help to achieve more accurate estimates.

\subsection{Requirements Document}
The starting point of conventional system development is a requirements or functional specification: it states what the system is supposed to do, see \cite{Rupp.@2014}.
Part of it are safety requirements that specify the required quality with regard to possible hazards, e.g.\ derived from a risk analysis.

\begin{figure}[htbp]
    \centering
    \includegraphics[width=0.8\textwidth]{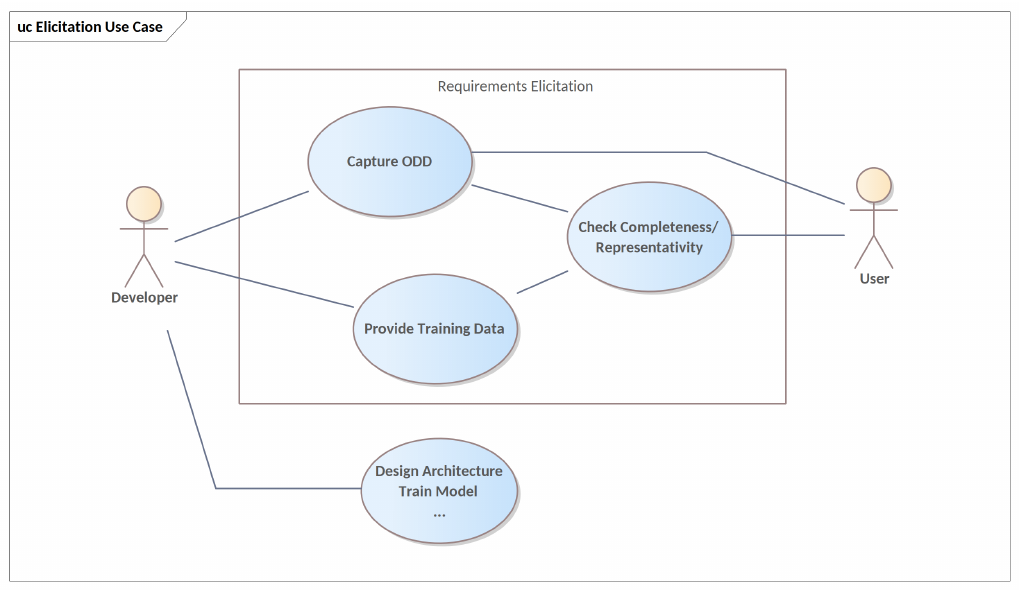}
    \caption{Requirement Elicitation}
    \label{fig:RequElic}
\end{figure}

For an AI system, this is not so easy.
The target environment is so complex and diverse that a description of what the system should be able to do can no longer be readily cast into requirements.
This corresponds to the fact that these systems are not implemented imperatively and do not work deductively according to rules, but are trained, see Section~\ref{sec:NeedStatTest}.
This raises the question of how requirements for such a system can be formulated in a comprehensible, verifiable way, see fig.~\ref{fig:RequElic}.
After all, such consultation with the user is essential for a valid and successful development.

We can turn the problem around and start from traceability through testing -- a kind of test-driven requirements engineering -- beginning with a PEON for the test.
Such a description of the ODD makes the customer's idea of the target environment explicit: which objects can occur in it, with which characteristics, and how the customer sees the statistical relationships.
Statistical quality criteria can then be attached directly: the system shall distinguish between dogs and cats in Germany from images with a reliability of 90\,\% -- and the PEON tells you what the dogs and cats will statistically look like.
A tester can sample and specify test cases from these requirements and evaluate the system in question.

In this sense, a probabilistic extension of an ontology can serve as a requirements document for the quality of an AI application.
The developer must then demonstrate that the tests exhibit the defined quality, at the given significance, with respect to this probability model.

The standardization of ontologies for describing the ODD of autonomous driving, see \cite{ASAMOpenXOntology}, can be regarded as a first step in this direction.

\subsection{Ethical Demands}\label{sec:DethicalDemands}\footnote{\scriptsize I would like to extend my sincere thanks to N. Grube for initiating the discussion on this topic.}
Consider a scenario in which a person on the tracks must be detected reliably: the failure rate is required to be at most $p=10^{-8}$.
Suppose the system has been optimized to a point where it misses persons with light skin in only $p=10^{-10}$ of cases, while its failure rate for persons with dark skin is $p=2\cdot10^{-8}$.
Assuming an equal distribution of individuals across the light-to-dark skin spectrum in the population, the system misses a person in approximately $10^{-8}$ of cases on average -- and would thereby fulfill the requirement.
Clearly, this is not in accordance with ethical standards and legislation, which stipulate that the quality of products and services must not depend on skin colour: ``No person shall be favoured or disfavoured because of sex, parentage, race, language, homeland and origin, faith or religious or political opinions. No person shall be disfavoured because of disability.'' (Basic Law for the Federal Republic of Germany, Art.~3(3))

What is actually required is that a person be missed in at most $p=10^{-8}$ of cases \emph{regardless of skin colour}.
We therefore have to prove by testing that the failure rate is at most $10^{-8}$ for \emph{each} skin colour group.
In Section~\ref{sec:SampleSize}, we discussed the statistical estimation of system accuracy from test cases based on a PEON, and we derived test termination criteria for this purpose.
Evidently, in specific domains, ethical considerations require refining these statements to sub-domains: we must be able to verify the quality criteria for specific sub-populations.
In the context of person detection, these correspond to individuals with different skin colours.
To this end, a classification \textit{skin colour}, denoted $C_{skin}$, is introduced in the ontology.
The set of situations involving a person with a specific skin colour, say \textit{skin\_colour}, is then given by the set of all partition classes with $C_{skin}\cong$ \textit{skin\_colour}; for short, let $C_{others}$ denote all other classifications.
Let $P$ be the probability distribution on the set of all characteristics (partition classes) as specified by the PEON.
The Bayesian conditional expectation given $C_{skin}=$ \textit{skin\_colour} then provides the appropriate distribution over all remaining characteristics:
\bea
\mathcal{P}_{skin\_colour}(C_{others}):= P(C_{others} \,|\, C_{skin}= skin\_colour) \ \ \mbox{distribution for the domain with fixed skin colour}
\ea
This approach applies in full generality.
To establish the accuracy of the system under specific boundary conditions, we compute the Bayesian conditional expectation defined by those boundary conditions, i.e.\ we restrict the set of partition classes to those satisfying them.
We thereby obtain the adequate distribution over all characteristics beyond the given boundary conditions.
Taking this distribution as the starting point in Section~\ref{sec:SampleSize} allows the required quality criteria to be evaluated under the given boundary conditions.

\subsection{Evaluation of Training Data for ML-based Systems}\label{sec:DataEvaluation}
To check the completeness of data, it is advisable to create an ontology recording the essential entities with their characteristics, see \cite{Staab.2009,KIAbsicherung}.
The ontology underlying a PEON helps to make the description of the ODD as complete as possible, see \cite{ASAMOpenSCENARIO,scenic,dissSchuldt,CoreOnt2015,Breitenstein.2020,Bogdoll.01.09.2022,Breitenstein.11.02.2021,Guneshka22,Zipfl.21.04.2023,UlHaq.2019}.
As argued above, these approaches fall short in that they do not capture the required statistics.
This requirement must already be imposed on the training data: it must be statistically balanced.
Otherwise, systematic distortions are to be expected, see \cite{AIBias, Gao.13.02.2025}.

Training data collected haphazardly from the internet, user forums, etc.\ often carries a bias.
It is for the same reason that, for example, the German microcensus invests considerable effort in obtaining a representative sample of respondents.
The concept of \textit{probabilistically extended ontologies} can support the development of complete and balanced training data, as it enables a systematic specification of that data.
By making the statistical assumptions explicit -- in a model that can be evaluated independently and is modelled graphically -- it helps customer and engineer discuss their ideas about the ODD.
Given in a readable and verifiable form, the description also becomes accessible to an audit.
If in doubt, the model can be checked against empirical, descriptive statistics.
And given both training data and a PEON, we can compare their statistics, which is able to reveal hidden assumptions and prejudices.
PEONs thus offer quantifiable criteria for the quality of training data, see Section \ref{sec:TestTD}.

Conversely, given a PEON, we can sample abstract training data from it.
Rendered into concrete training data, this may yield a complete, balanced training set.

\subsection{Comparison of a PEON with Empirical Statistics}\label{sec:CompStatistic}
A PEON contains, in particular, an ontology.
As we will see in Section~\ref{sec:PEONModelling}, the statistics of the individual selections are described by probability distributions annotated at the nodes.
We can, of course, apply empirical descriptive statistics to check the reliability of this modelling.
Dependencies between different selections are mainly described functionally, with the help of a few parameters, and these parameters, too, can be tested empirically -- using correlation or variance tests, for example.

These techniques allow us to re-engineer PEONs from given labelled training data, see Section~\ref{sec:Reconstruction}.
We arrive at a readable representation of the completeness and balance of the training data, which can then be evaluated.

\subsection{Re-Evaluating Test Results from a Development Process}\label{sec:TestCorrection}
In many concrete projects, predetermined, labeled test data will already be available -- and perhaps also test results and the statistics associated with them.
One example is a test process in which the test data was determined combinatorially.
In Section~\ref{sec:TestingAI} we asked how such results can be related to the customer's expectation.
We can now answer this question, provided we have a PEON describing our unadulterated idea of the ODD as the customer sees it: we want to relate the test results to this PEON.

To do so, we start from the various labels of our test data and structure them in a new ontology $\mathcal{O}_{label}$.
We assume that the existing labels of the training and test data can be modelled well by such an ontology, i.e.\ that their partitioning is indexed by the labels; if necessary, the labeling of a selected sample should be extended accordingly.
We end up with an ontology such that every test case lies in a unique partition class.
Counting the relative number of elements in each class gives us a PEON.

We thus obtain two PEONs: one for the ontology $\mathcal{O}_{label}$ describing the labels, with probability distribution $P_{labels}$, and our starting PEON with distribution $P_{ODD}$, living on its own partitioning and reflecting our idea of the ODD.

Let us make the crucial assumption that the PEON describing the ODD is finer than the one describing the training and test data.
This technical assumption is usually fulfilled, or can be fulfilled with little effort.
Being a refinement, $P_{ODD}$ induces, according to Section~\ref{sec:Refinements}, a probability distribution over $\mathcal{O}_{label}$, denoted by $P'_{labels}$.
We therefore obtain two probability measures over the partitioning of $\mathcal{O}_{label}$.
Completeness of the underlying PEON for $P_{ODD}$ implies that the distribution $P_{labels}$ is absolutely continuous with respect to $P'_{labels}$.
We can then compute the Radon--Nikodym derivative of the two distributions, see Section~\ref{sec:2PEONs2Onts}, and relate the expectation values of functions with respect to both, see Lemma~\ref{lem:Jacobian}.
In particular, we can relate test statistics -- being statistical expectation values of suitable functions -- across the two distributions: the statistics of the given test results are translated into statistics with respect to $P'_{labels}$, and thereby related to the ODD as modelled by our PEON.
In this way, existing test results can be given a meaning relative to the ODD.
Note that the significance of the results can be tracked along the way.

If test data exists for every partition class of the PEON describing the ODD -- i.e.\ for every constellation of entities that occurs in the ODD with non-negligible probability according to $P_{ODD}$ (completeness of the tests!) -- then the converse holds as well: $P'_{labels}$ is absolutely continuous with respect to $P_{labels}$, and we can compute which test statistics, e.g.\ of combinatorial testing, are to be expected.

Note, however, that the significance levels must be recalibrated in order to evaluate the quality of the tests in this way.

\section{Modelling a PEON}\label{sec:PEONModelling}
Perusal of the elementary example set out in Section~\ref{sec:DogCatEx} suggests that the modelling process is relatively straightforward.
The ontology under consideration is of a simple nature and consists of a single entity, \textit{Animals}, with two instances, \textit{cat} and \textit{dog}, which may be supplemented by colours.
The image world is divided into two classes, determined by whether a dog or a cat is depicted; this classification is further refined by distinct colours, yielding a more nuanced categorization.
It is evident from the data of the Federal Statistical Office that the occurrence probability of dogs in photos from Germany is approximately $3/7$, and $4/7$ for cats.
For a further breakdown by colour, we need to delve a little deeper into the statistics.

In this case, it is still easy to assign the various appearances of animals and colours to their respective probabilities of appearing in pictures.
For autonomous driving, however, matters are much more complicated.
Even if we could identify the various partitions, describing their occurrence probabilities appears hopelessly complicated:
in the best case, we would end up with huge posters covered in numbers that are neither easy to understand nor easy to read in their entirety.

In general, ontologies can also contain cyclic dependencies, which makes them even more difficult to understand.
On the other hand, various techniques exist for breaking up such cyclic dependencies by introducing common influencing factors, see \cite{Pearl.b, Koller.2009}, so we restrict ourselves to ontologies that consist only of tree- or, more generally, forest-like dependency graphs.
As directed acyclic graphs, they admit a graphical modelling that remains understandable even for readers unfamiliar with the formalism.

A notable benefit of this restriction is that it lets us employ the Bayesian network technique for acyclic graphs.
Applying Kahn's algorithm topologically sorts the nodes, so that for every dependency the parent node is processed before its dependent children.
Later, we will construct a Bayesian network from the PEON data along these lines, and use it to describe complex distributions on the partitions.

In a Bayesian network, nodes represent sets of options.
A dependent child node contains selections whose probabilities depend on the previous selection at the parent node; the conditional probabilities for the associated selections are annotated at the nodes themselves.
This technique works well for smaller networks, where hierarchical structures keep the description organized -- but for a mature statistical description of an ODD, even this is not maintainable.

Nonetheless, the approach of describing complex distributions via Bayesian networks is sound, and we build on it.
We will, however, first develop a different model, a PEON, which can then be used to compute a Bayesian network -- and thus a distribution on the space of partitions -- algorithmically.

This study aims at developing a more sophisticated ontology that can serve as a point of departure for testing automatic train operation (ATO).
The following sections demonstrate the extension of an ontology through a probabilistic lens.

Rather than starting with the ontology for tracks and signals of Section~\ref{sec:OntTypeTracks}, however, we begin with an ontology of weather conditions, as it is better suited to illustrate the issues that motivate extending an ontology to a PEON.

\subsection{Entities and Attributes}\label{sec:EntAtt}
A graphical model of a PEON is, first and foremost, an acyclic graph whose essential elements are nodes and edges, together with typed, directed arrows describing special relations between the nodes.
We distinguish three kinds of nodes: \textit{branch nodes}, \textit{classification nodes}, and \textit{class nodes}.

Classification and class nodes correspond to the nodes of the classification tree method:
classification nodes describe properties -- such as the curvature of a track section or a wind speed -- while class nodes represent the corresponding feature classes, or equivalence classes, e.g.\ \textit{strong} or \textit{weak} wind.
Branch nodes, to a first approximation, correspond to the compositions mentioned there.

\Ex
Consider modelling weather conditions, with wind speed, precipitation, and temperature among the phenomena of interest.
Wind speed may be categorized as \textit{strong, moderate, weak}, or \textit{calm}; temperature ranges from \textit{very cold} to \textit{chilly}, and so on.

We model \textit{weather conditions} as a branch node, with its characteristics \textit{wind}, \textit{temperature}, and \textit{precipitation} as classification nodes, and the associated values \textit{strong, moderate, weak, calm}, etc.\ as their class nodes.
\Exa

\begin{figure}[htbp]
    \centering
    \includegraphics[width=0.4\textwidth]{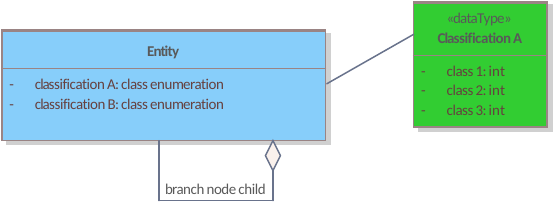}
    \caption{Branch nodes $\simeq$ tables; their classification child nodes $\simeq$ attributes; and the class-node children of a classification node specify its data type}
    \label{fig:DataModel}
\end{figure}

In a first approximation, a branch node can thus be read as an entity that carries certain characteristics, given by its classification nodes.
The classification nodes of a branch node are modelled as its children, and the classes attached to a classification -- its own children -- define the possible values of that characteristic.
In particular, class nodes are the leaves of the PEON, see fig.~\ref{fig:DataModel}

Some properties, such as localization, are shared by many entities and are therefore modelled as a separate subtree to which the relevant entity nodes simply refer, using \textit{PROPERTIES} links.

With this interpretation in hand, we can give a first, preliminary description of a PEON.
We begin by identifying the entities -- objects or actors -- that may appear in a scene, and model each as a branch node.
Where a fixed or bounded number of instances of an entity can occur in the scenery, we record this in a field \textit{COUNT} attached to the branch node: if between $n_0$ and $n_1$ objects of that type may occur, we write \textit{COUNT} $=[n_0,n_1]$.
The occurrence probability for a given count need not be uniform -- for instance, fewer pedestrians are to be expected on rainy days, and lower counts may generally be more probable -- so we may instead specify a distribution over the counts in a field \textit{Marginal Distribution} attached to the branch node; the uniform distribution is taken by default, see fig.~\ref{fig:railroadtrack}.

It is also conceivable that only a limited selection of a set of possible objects may occur in a scene -- e.g.\ at most three of four possible signal types.
This is modelled by a further field, \textit{SELECTION}: writing \textit{SELECTION}~$=$~XOR$(2)$, for example, states that exactly two of the child nodes are to be selected, see fig.~\ref{fig:signals}.

Note that we frequently have compositions of several objects assembling another one -- a \textit{railway track}, for instance, composed of several \textit{track parts}.
A branch node can therefore have other branch nodes as children, corresponding to the $n$-to-$1$ relations of tables in a data model.

Classification nodes describe attributes of their parent entity, and their class nodes the possible manifestations of these attributes.
A key aspect of the probabilistic description of an ODD is to specify these properties in probabilistic terms: the distribution of the corresponding values is given in the classification node's \textit{Marginal Distribution} field.

Initially, the categories of a classification are merely abstract terms, such as \textit{large, small, long,\dots}.
These terms, however, usually refer to concrete values or intervals -- e.g.\ \textit{large} $=$ between one and two metres -- which are recorded in a \textit{Value} field of the class node; for a concrete test case, specific values are then sampled from this field.

The remaining challenge is to specify the dependencies between the entities and attributes.
How, for instance, should we express that it does not snow at high temperatures, or that high-speed rail lines have fewer curves?
For this purpose we introduce a dedicated link type, indicated by \textit{DEPENDENCIES} links.
The following chapter investigates various such dependencies and shows how to specify them.

\subsection{Weather Conditions as a simple Example of Environmental Conditions}\label{sec:EnvConditions}
\begin{figure}[ht]
  \centering
  \begin{minipage}{0.45\textwidth}
    \centering
    \includegraphics[width=\textwidth]{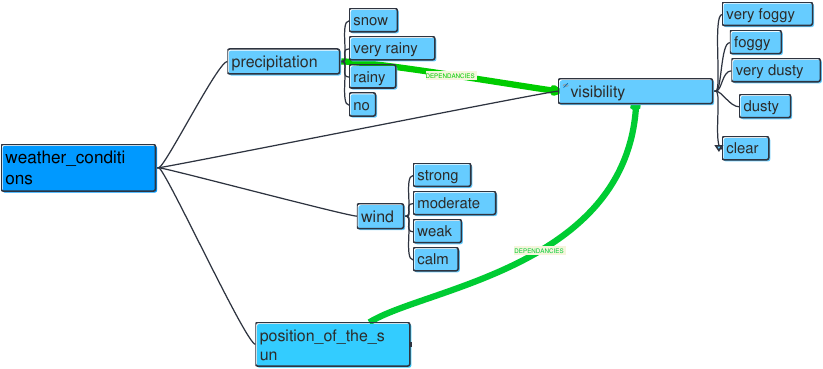}
    \caption{Various weather conditions which impact recognition}
    \label{fig:weather}
  \end{minipage}\hfill
  \begin{minipage}{0.45\textwidth}
    \centering
    \includegraphics[width=\textwidth]{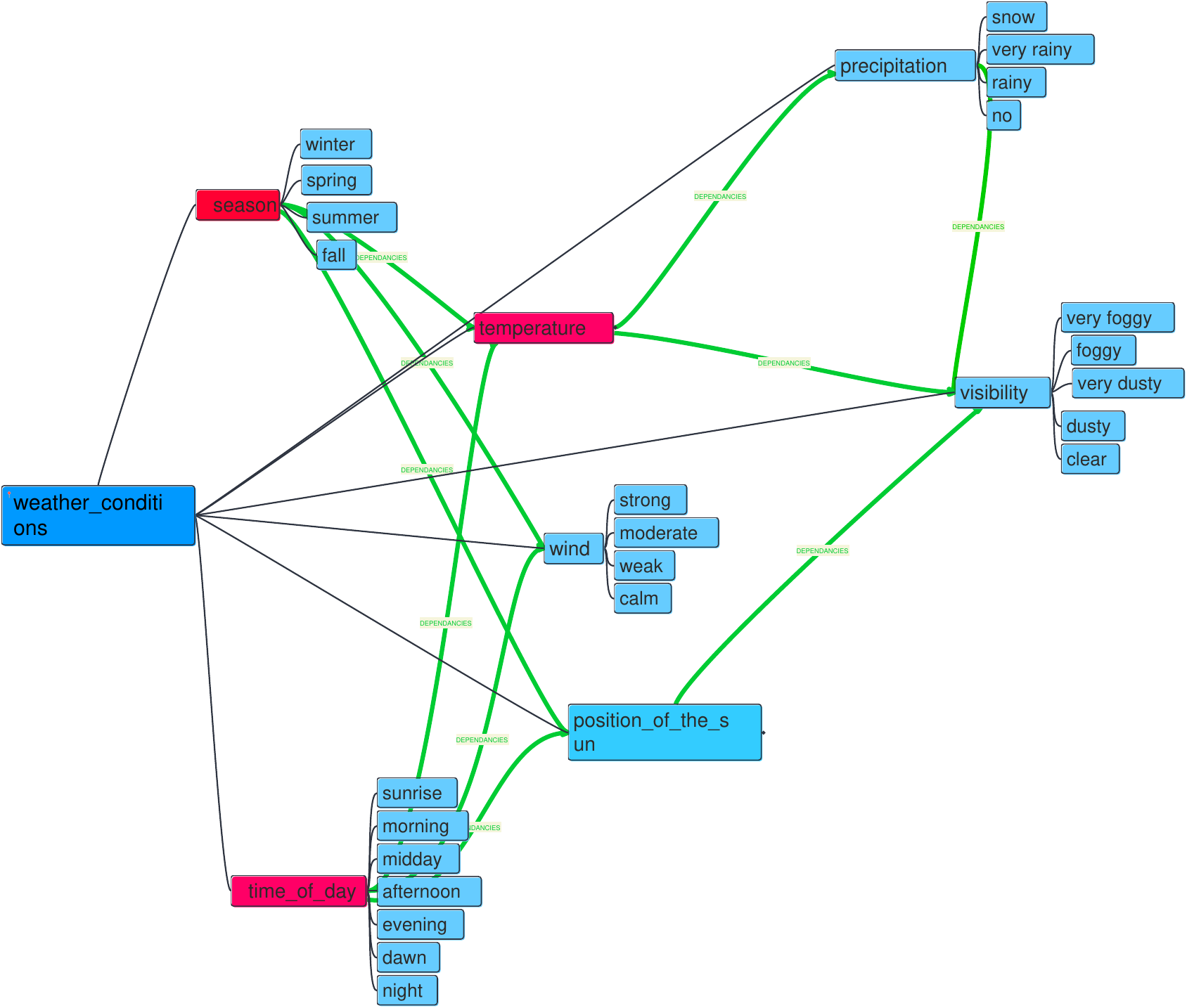}
    \caption{Extended ontology for better modelling of dependencies}
    \label{fig:ext_weather}
  \end{minipage}
\end{figure}

Let us start with the weather conditions, as these form the backdrop for many AI systems -- especially obstacle detection systems, see fig.~\ref{fig:weather}.

First, we identify the most important weather factors influencing recognition in images or videos: rainfall, visibility, wind, and the position of the sun, with manifestations such as \textit{snow, heavy rain,\dots} and \textit{no rain, foggy, dusty,\dots,clear}, etc., they depend on, see fig.~\ref{fig:weather}.

Next, we look at the dependencies between these factors.
It is obvious that heavy rain, for example, limits visibility -- visibility depends on rainfall.
But there are hidden dependencies as well: we would expect snow mainly in winter, when the sun stands a little lower in the sky (position of the sun $\Leftrightarrow$ rainfall), and fog is more likely in the morning and evening than at midday, when the sun is high (position of the sun $\Leftrightarrow$ visibility).
These observations suggest that we should consider not only these factors, but also common factors such as the season, the time of day, and the temperature, see fig.~\ref{fig:ext_weather}.

This is in line with our restriction to tree- or forest-like graph structures, which is what enables the use of Bayesian networks in the first place, see Section~\ref{sec:PEONModelling}.

\subsubsection{Probability Distributions for Independent Selections (Marginal Distribution)}
The model in fig.~\ref{fig:ext_weather} is somewhat more complex, but it helps us greatly in modelling the dependencies.
With this model at hand, we now describe the occurrence probabilities with respect to the corresponding partitioning.
Instead of working directly with conditional probabilities, as required for a full Bayesian network, we first neglect any graphical structure -- any dependency between the individual factors -- entirely, and examine the distribution of each selection independently of the others.
How often does it rain heavily in Germany, regardless of the season and time of day? How often is it foggy on an annual average, and how strongly does the wind typically blow?
A quick search on the internet, or an inquiry at a meteorological institute, gives us the average rainfall probability for Germany, as well as the annual frequency of fog or dust; we store these numbers at the corresponding nodes.
Assuming the seasons are of equal length, winter occurs with probability $0.25$, as do the others -- the distribution over seasons is uniform.
For the time of day, we can likewise use the average duration of each period to compute its probability of occurrence.

\begin{figure}[ht]
  \centering
  \begin{minipage}{0.45\textwidth}
    \centering
    \includegraphics[width=\textwidth]{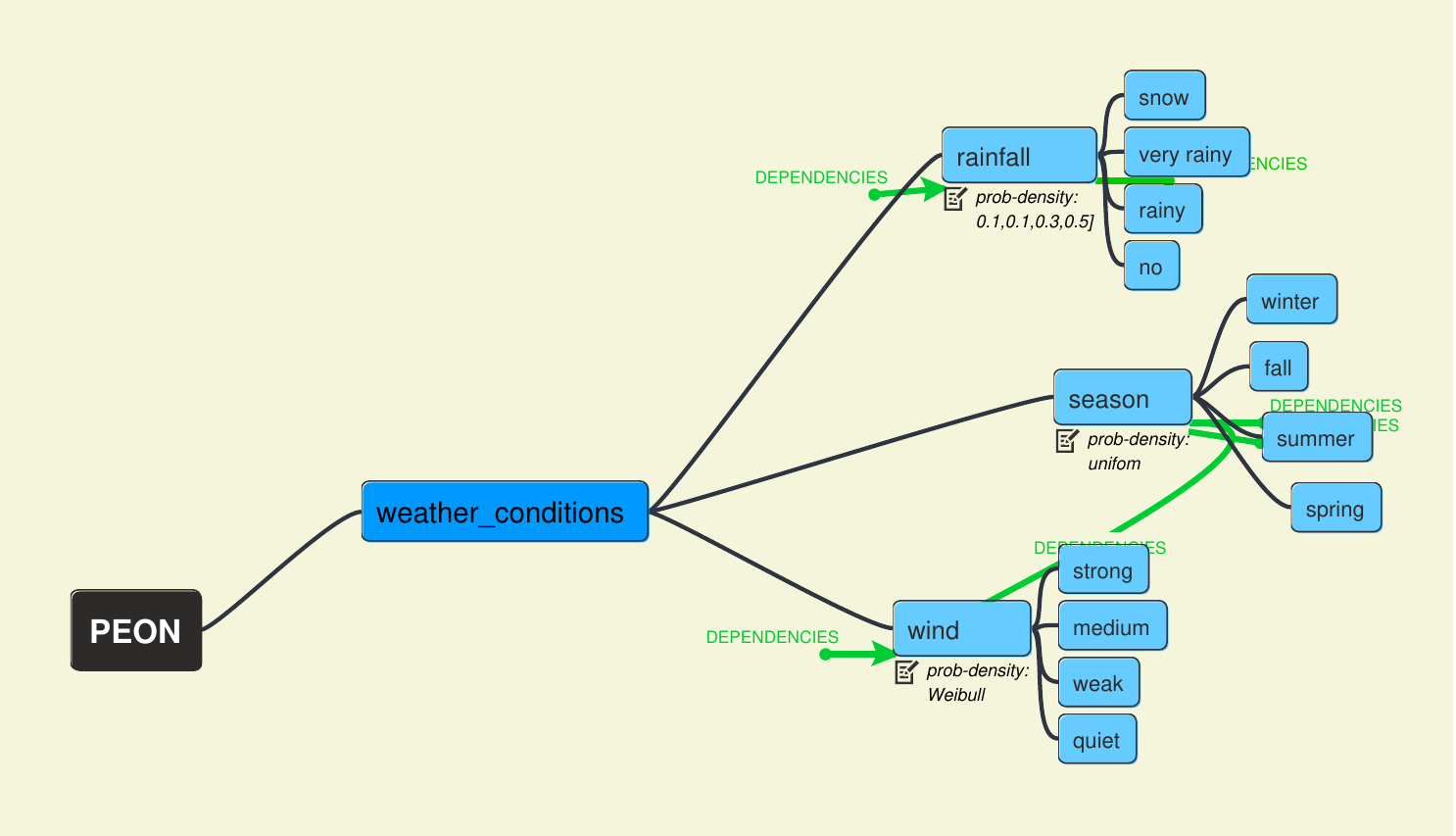}
    \caption{Specification of probability distributions}
    \label{fig:WeatherProbs}
  \end{minipage}\hfill
  \begin{minipage}{0.45\textwidth}
    \centering
    \includegraphics[width=\textwidth]{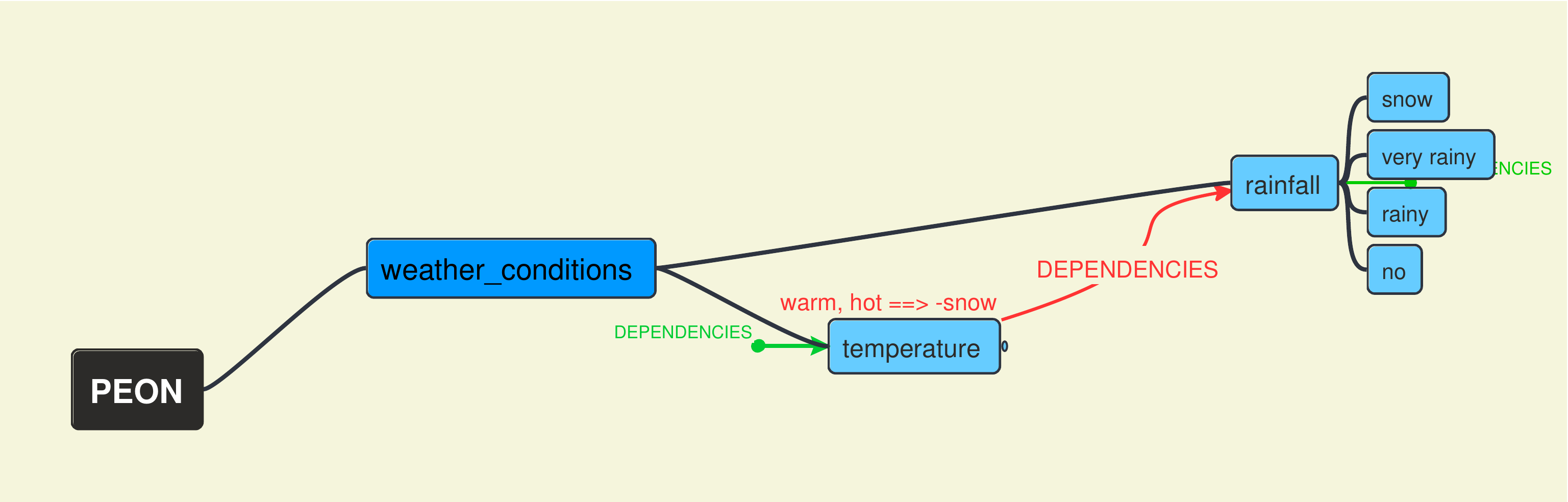}
    \caption{Logical dependency between temperature and rainfall}
    \label{fig:DepTempRain}
  \end{minipage}
\end{figure}

As a slightly more involved example, consider the distribution of wind speed.
Behind the individual selections we place the speed ranges distinguishing strong from weak winds.
From meteorological tables we know that wind speed in Germany follows a Weibull distribution, a well-known family of probability distributions; with the appropriate parameters, we can then compute the occurrence probability of strong winds, and so on.
Instead of storing the various occurrence probabilities as a table, it therefore suffices, in such cases, to specify the general distribution family, see fig.~\ref{fig:WeatherProbs}; concrete values can then be computed algorithmically by taking advantage of the specified ranges for class values and performing integration.

We proceed in this way for all branch and classification nodes.

Where the probability distribution can be specified functionally, using numerical values and intervals to delineate the selection options, maintaining the description is significantly simplified:
rather than readjusting the distribution whenever classes are refined or coarse-grained, this can be done algorithmically.

The problem of dependencies, however, remains.

\subsubsection{Logical Dependencies}\label{sec:LogDep}
In the context of driving, weather conditions are typically characterised by warm or cold temperatures, combined with more or less frequent precipitation.
We want to estimate the probability of the various combinations -- warm and rainy, cold and dry, and so on -- which requires modelling the joint distribution on the product space of temperature and rainfall.
While the individual distributions of temperature and rainfall are readily available, we still need the probability of their joint occurrence.
One simple observation applies throughout: at high temperatures, it does not snow.
Our goal is to express this fact probabilistically.

Let $p_{rain}(R)$ describe the distribution of rainfall $R\in\{snow,\dots,no\}$, and $p_{temp}(T)$ that of temperature, $T\in\{very\ cold,\dots,hot\}$, as stored at the respective nodes.
We are looking for a distribution on the product space, i.e.
\be
p_{rain\times temp}(R\times T)\ \ \mbox{ on }\ \ \{snow,\dots,no\}\times\{very\ cold,\dots,hot\}.
\ee
By Bayes' theorem this can be reformulated via the conditional probability:
\be
p_{rain\times temp}(R\times T) = P_{rain|temp}(R|T)\cdot p_{temp}(T) \ \ \ \mbox{(Bayes' theorem)}
\ee
with $P_{rain|temp}(R|T)$ the conditional probability of rainfall $(R)$ given temperature $(T)$.

In this formulation, our simple observation can be stated directly as a restriction on the conditional probability:
\bea
&&\mbox{If the temperature is warm or hot, then the probability of snow is }0 \ \Leftrightarrow \nonumber \\
&&\hspace{1cm} P_{rain|temp}(snow \mid T\in\{\mbox{warm or hot}\}) = 0
\ea

For simplicity, we assume that, apart from this logical constraint, the selections are stochastically independent.
With the mathematics of Appendix Section~\ref{sec:coupling}, we can then find an optimal solution on the product space that satisfies both the given marginal distributions and the logical constraint; explicit algorithms allow us to compute the resulting joint distribution $p_{rain\times temp}$.

In our graphical model, we draw a DEPENDENCIES arrow from the temperature node to the rainfall node, marking it informally by $\{warm, hot\}\Rightarrow \lnot snow$, see fig.~\ref{fig:DepTempRain}.
Given this information, suitable algorithms compute the joint distribution of rainfall and temperature for us.

\subsubsection{Functional Dependencies}\label{sec:FuncDep}
\begin{figure}[ht]
  \centering
  \begin{minipage}{0.45\textwidth}
    \centering
    \includegraphics[width=\textwidth]{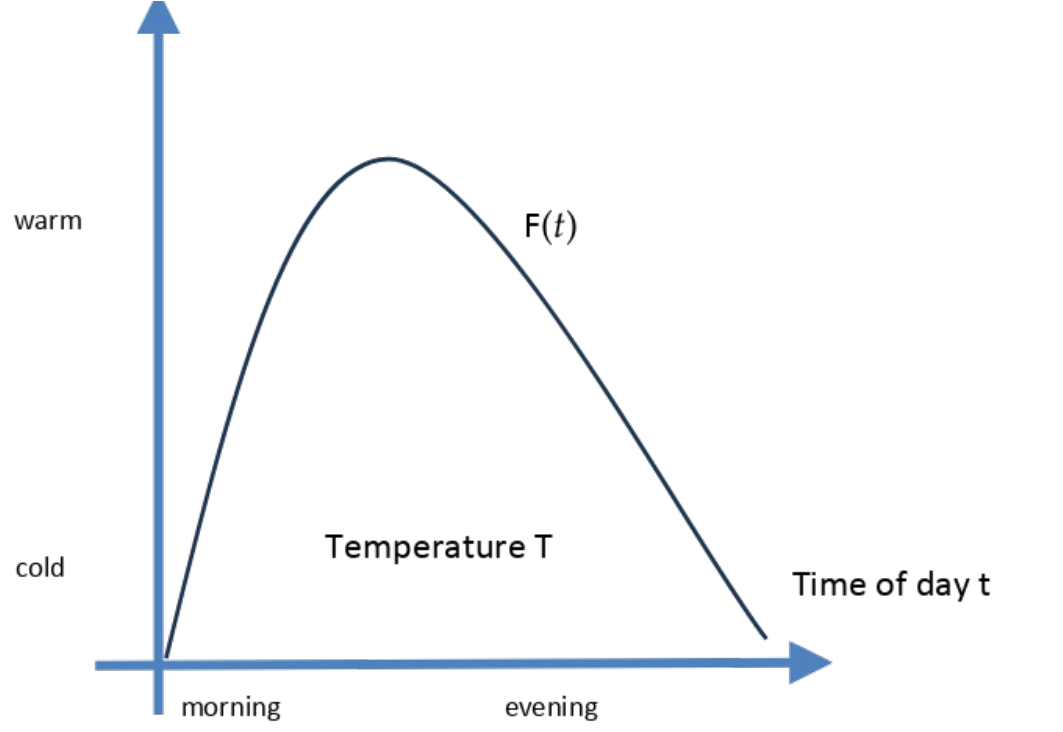}
    \caption{Temperature distribution over the time of day}
    \label{fig:TempTime}
  \end{minipage}\hfill
  \begin{minipage}{0.45\textwidth}
    \centering
    \includegraphics[width=\textwidth]{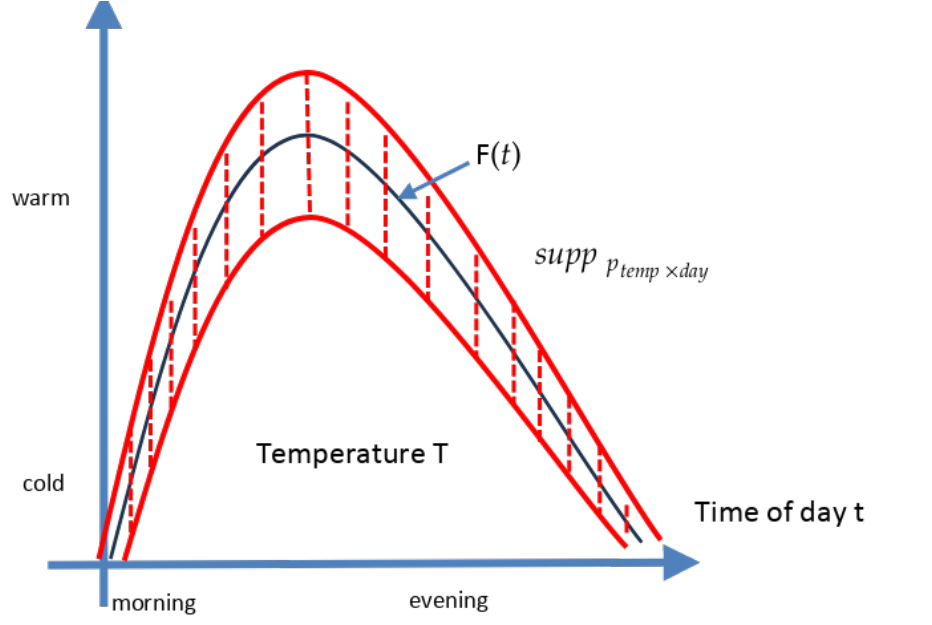}
    \caption{Essential support of the probability distribution, close to the function graph}
    \label{fig:TempDaySupp}
  \end{minipage}
\end{figure}

Next, we turn to dependencies of a more general kind, going beyond those captured by ontologies alone: functional dependencies.
Consider once more temperature and time of day.
The marginal distributions are again obtained from meteorological data, or determined over an annual average day.
For simplicity, assume these distributions are given as densities $p_{temp}(T)$ over the temperature range $T\in[T_{min},T_{max}]$ and $p_{time}(t)$ over the time of day $t\in[t_0,t_{24}]$.
Dividing the temperature and time ranges into finitely many classes, the distribution over these classes is obtained by integration.
The dependence of temperature on time of day, in a first approximation, follows a parabola open at the bottom: the average temperature is higher at midday than in the early morning or at night, see fig.~\ref{fig:TempTime}.
We are looking for the joint distribution of time of day and temperature, $p_{temp\times day}(T,t)$.
Using Bayes' formula, we write informally
\be \label{equ:FuncDep}
ess.supp (P_{temp\ \times\ day}(T|t)) \approx \mbox{graph } F
\ee
with $F$ the function of fig.~\ref{fig:TempTime}.
This informal description can be made precise: our objective is to describe probabilities of occurrence, not functional values, so equation \ref{equ:FuncDep} should be read as stating that, for a given time of day, only temperature values close to $F(t)$ carry non-negligible probability, i.e.
\be
P_{temp\ \times\ day}(T|t) \approx \delta_{T==F(t)}(T,t) \ \ \mbox{with } \delta_{T==T_1}(T)=1 \mbox{ iff } T==T_1, \mbox{ else } 0.
\ee
By the normalization $\int_T P_{temp\ \times\ day}(T|t)=1$, and since
\be
\int_T \delta_{T==F(t)}(T,t)\, p_{temp}(T) = p_{temp}(F(t)),
\ee
it follows that
\be
P_{temp\ \times\ day}(T|t) \approx \delta_{T==F(t)}(T,t)\, \frac{p_{temp}(T)}{p_{temp}(F(t))}
\ee
where, for the moment, we assume $p_{temp}(F(t))>0$.
(The formulas hold in general, but a fully rigorous treatment would need to address null sets.)

With this conditional probability, we obtain the following distribution on the product space:
\be
p_{temp\times day}(T,t) \approx \delta_{T==F(t)}(T,t)\, \frac{p_{temp}(T)\cdot p_{day}(t)}{p_{temp}(F(t))}
\ee
By construction, the first marginal condition holds, i.e.\ $\int_T p_{temp\times day}(T,t) = p_{day}(t)$.
For the second condition, the $p_{temp}$ factors cancel, and we obtain
\bea
\int_t p_{temp\times day}(T,t) = p_{day}(t\in F^{-1}(T))
\ea
The right-hand side is a familiar construction:
the function $F:\{\mbox{time of day}\}\rightarrow\{\mbox{temperature}\}$ induces, besides $p_{temp}$, a further distribution on the space of temperatures -- the pushforward measure
\be
p^F_{temp}(T):= F_\#(p_{day})(T) := p_{day}(F^{-1}(T))
\ee
This equation is stated here for discrete probability spaces, but it holds generally for measurable functions, see \cite{Villani.2009} or Appendix Lemma~\ref{lem:pushforward_measure}.
We thus obtain two distributions on the space of temperatures.

Assume for a moment the ideal case of a functional equality $T=F(t)$ between temperature and time of day, i.e.\ that the temperature followed the curve $F$ exactly throughout the day.
The assignment $F$ could then be read as a reparametrization of the temperature, and the substitution rule shows that the two measures are equal, merely expressed with respect to different parametrizations.
The essential support of the conditional probability is then exactly
\be
\mbox{ess\ supp}(P_{temp|day}) = \{(F(t),t) \in \{temperature\}\times\{time\ of\ day\}\},
\ee
i.e.\ the graph of $F$.
And indeed in this case the distributions $p_{temp}$ and $p_{temp}^F=F_{\#}(p_{day})$ must coincide.

In practice, however, there is no simple, physical functional relationship between temperature and time of day, and indeed $p_{temp}\neq F_\#(p_{day})$.
The functional dependence is only approximate and probabilistic, and can be formulated as such:
we look for a distribution on the product space that respects the given marginal distributions $p_{temp}$ and $p_{day}$, while placing the essential support of its conditional probability as close as possible to the graph of $F$, see fig.~\ref{fig:TempDaySupp}.
Using the pushforward measure $F_\#(p_{day})$, this can be reformulated as seeking a coupling $\widehat{p}_{temp\times temp}(T_1,T_2)$ of the two measures $p_{temp}$ and $F_\#(p_{day})$ on the space of temperatures (Appendix Section~\ref{sec:coupling}) -- respecting the given marginals, with essential support close to the diagonal.
We thus consider
\be
\int_{T_1,T_2} |T_1-T_2|\ \widehat{p}_{temp\times temp}(T_1,T_2)
\ee
as a measure of the essential support's spread around the diagonal, which we minimize.
Formulated this way, the problem is a transport problem, see Appendix Section~\ref{sec:TransportProblem}, and a minimal solution can be found with the help of standard tools, see \cite{Villani.2009,Peyre.,Cuturi.c} -- an optimal coupling with minimal spread.
Generically we obtain a unique coupling on $\{temperature\}^2$ for $p_{temp}, F_\#(p_{day})$,
\be
p_{sol.\ of\ transport\ problem}(T_1,T_2) = \underset{\mbox{coupling}\ \widehat{p}}{\min}\ \  \int_{T_1,T_2} |T_1-T_2|\ \widehat{p}_{temp\times temp}(T_1,T_2)
\ee
which pulls back to a coupling for $\{temperature\} \times \{time\ of\ day\}$ with respect to $p_{temp},p_{day}$.
Equivalently, we can look directly at the solution for
\be \label{equ:TransProb}
p_F(T,t) := p_{sol.\ of\ transport\ problem}(T,t) := \underset{\mbox{coupling}\ \widehat{p}}{\min}\ \ \int_{t,T} dist(T,F(t))\ \ \widehat{p}_{temp\times day}(T,t)
\ee
looking at the minimal couplings over $p_{temp},p_{day}$.
This is again an instance of the classical transport problem, in which any nonnegative distance function may be used, see Appendix Section~\ref{sec:TransportProblem} for details.
This connection allows very general probabilistic relationships to be modelled: by choosing suitable distance functions and functions $F$, the solution to the transport problem serves as a model for a wide range of probabilistic dependencies.

Our aim was to specify a distribution on the product space of temperature and time of day that is as close as possible to the informal description of equation \ref{equ:FuncDep}.
We have now obtained such a distribution -- the solution of the transport problem with $dist(T,F(t))=|F(t)-t|$.

\subsection{Noise}\label{sec:Noise}
The construction above specifies the distribution on the product space via an informal functional description, but this optimal implementation of a functional dependency does not always match our intuition exactly.
In most cases, such relationships are noisy: we do not expect the narrowest possible band around the diagonal, but a certain width around it.
The dependence of temperature on time of day, for instance, is certainly not simply distributed optimally according to the functional description, but with a certain amount of noise.

The type of noise present in such data can vary considerably, but thermal noise is arguably the most natural choice here.
Its strength is governed by an inverse temperature $\beta$: the larger $\beta$, the sharper the dependency; the smaller $\beta$, the noisier.
Consider an ordinal set of classes $\{0,1,\dots,n\}$ -- e.g.\ the four classes $(0,1,2,3) = \textit{clear, dusty, very dusty, foggy}$ used for the visibility condition.
Without noise, each class $n_i$ would retain its full probability mass; thermal noise blurs this picture by reassigning part of the mass of class $n_i$ to a neighbouring class $n_j$, with a share that decays with the squared class distance, $\exp(-\beta\,(n_i-n_j)^2)$.
Even with the sun high in the sky, it may thus still occasionally be foggy -- just less frequently, the larger $\beta$ is chosen.

Crucially, we do not simply reweight the marginals with this kernel, as that would destroy the prescribed marginal distributions.
Instead, the kernel biases the choice of coupling among all couplings of the two given marginals: it is rescaled row- and column-wise until the resulting coupling again reproduces the prescribed marginals exactly, while remaining concentrated -- according to the thermal kernel -- around the diagonal $n_i\approx n_j$.
In this way, we model noise for dependent, ordinal data as a function of the inverse temperature $\beta$; the formal construction is given by Definition~\ref{def:ThermalNoise} in Appendix Section~\ref{sec:NoiseModelling}.

\begin{figure}[htbp]
  \centering
  \includegraphics[width=0.9\textwidth]{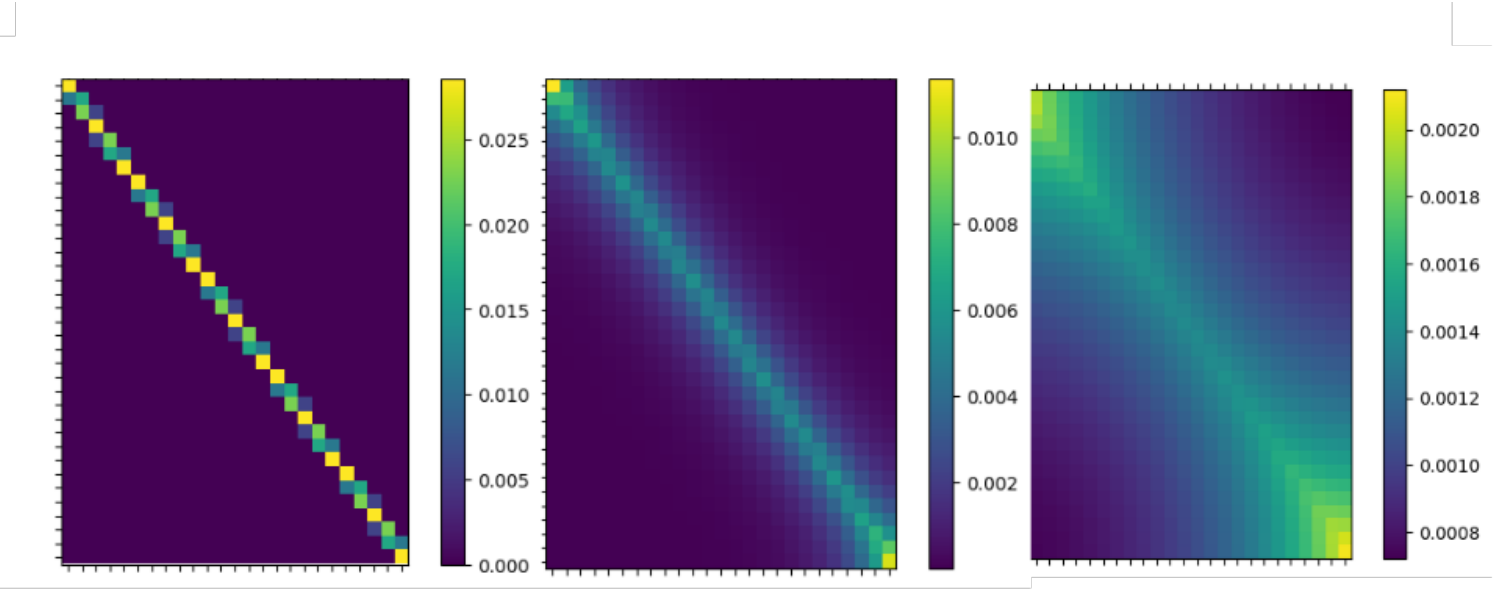}
  \caption{Linear coupling displaying the properties one would expect of a noise coupling ($\beta = 1, 0.9,0.1$)}
  \label{fig:NoiseCoupling}
\end{figure}

\Ex
Consider two uniform distributions on 35 and 25 classes, respectively, coupled linearly and with some noise.
At zero temperature there is no noise; as the temperature increases, so does the noise, until at very high temperature the coupling approaches complete noise, i.e.\ the two variables become uncorrelated (their product measure).
This is illustrated in fig.~\ref{fig:NoiseCoupling}, where the inverse temperature is varied from $1$ to $0.9$ to $0.1$.
\Exa

This construction generalizes readily to functionally dependent couplings, as the following example shows.

\begin{figure}[htbp]
  \centering
  \includegraphics[width=0.7\textwidth]{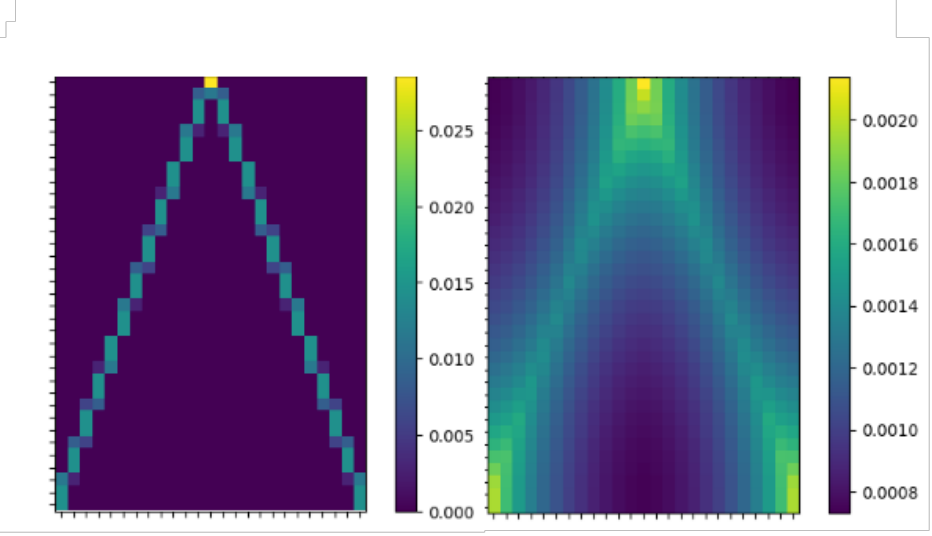}
  \caption{Simple functional dependency $x\mapsto abs(x-12)$ with noise (inverse temperature $=1\to 0.1$)}
  \label{fig:NoisyAbsXmin5}
\end{figure}

\Ex
Consider two uniform distributions on a set of 10 classes.
The coupling is described functionally by $x\mapsto abs(x-12),\ x\in\{0,\dots,9\}$.
The construction of this chapter results in the distribution shown on the left of fig.~\ref{fig:NoisyAbsXmin5}; the corresponding temperature-dependent noisy distribution is shown on the right (note the value scale!).
The shape of the functional dependency is clearly visible, washed out under noise.
\Exa

\subsection{Environmental Conditions Continued}
Having laid out these general considerations, we now turn to how environmental conditions can be modelled in practice.

\subsubsection{Logical Dependency Revisited}
Using the techniques developed so far, let us take another look at logical dependencies.
First, a small observation: if two or more selection options (classes) are equivalent with regard to testing or further consequences, they can be grouped into a summary class, with the sum of the individual probabilities assigned to it.
We already used this implicitly when summarizing value ranges in an ontology.
As an example, consider rainfall where we are only interested in whether it is snowing or not:
we summarize the classes \{very rainy, rainy, no\} into one class, and \{snow\} as the second.
The probability of the summarized class is then the sum of the individual probabilities, $p_{strong\ rainy}+p_{rainy}+p_{no}$.

Conversely, once a selection has been made in this new, coarser class, it still corresponds to a specific selection from one of the merged classes:
we can sample the finer selection according to the individual probabilities of the merged classes.
In our example, assume it is not snowing; we then select \textit{rainy} with probability $p_{rainy}$, and \textit{no} rain with probability $p_{no}$, and so on. Mathematically:
\bea \label{equ:MergeClasses}
p_{\{class\ A_1,\dots,class\ A_n\}} &=& \sum_i p_{class\ A_i} \nonumber \\
x\in\{class\ A_1,\dots,class\ A_n\} &\Rightarrow& \mbox{select } x\in class\ A_i \mbox{ with probability } p_{class\ A_i}
\ea
In this way, we can model, in larger contexts, that a specific selection within a subclass is not relevant, while still allowing a specific selection to be sampled afterwards, without distorting the statistical relationships.
Appendix Section~\ref{sec:EquivRel} generalizes this construction and makes it mathematically precise.

Let us apply this observation to the logical dependency of Section~\ref{sec:LogDep}, between temperature and rainfall.
Under the logical constraint, the relevant classes are $\widehat{cold}=\{cold, very\ cold\}$, $\widehat{warm}=\{chilly,warm,hot\}$, and, for precipitation, $\widehat{rainy}=\{very\ rainy,rainy,no\}$, $\widehat{snow}=\{snow\}$.
We condense the selection options into two classes each:
\[
  \widehat{rainfall} = \{\widehat{rainy},\widehat{snow}\}, \ \ \ \widehat{temperature} = \{\widehat{cold},\widehat{warm}\}
\]
This gives two probability distributions on two-element sets: $p_{\widehat{temperature}}$ and $p_{\widehat{rainy}}$.

\begin{figure}[htbp]
  \centering
  \includegraphics[width=0.4\textwidth]{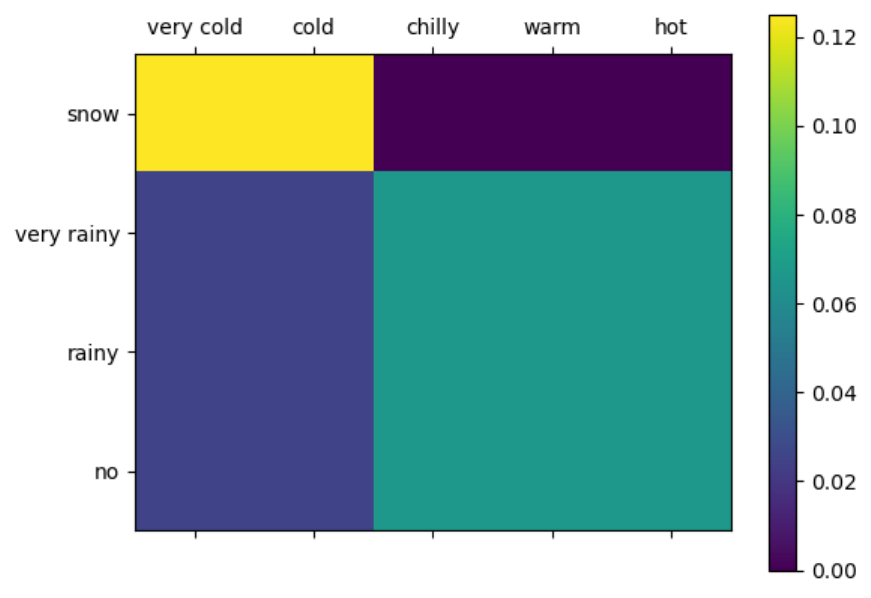}
  \caption{Logical constraint: whenever snow falls it cannot be warm or hot}
  \label{fig:LocConstrRainTemp}
\end{figure}

We want $\widehat{snow}$ to occur only alongside $\widehat{cold}$.
To this end, we define a mapping $F(\widehat{snow})=\widehat{cold}$, $F(\widehat{rainy})=\widehat{warm}$, and look for an optimal coupling of $p_{\widehat{temperature}}, p_{\widehat{rainy}}$ with the graph of $F$ as essential support.
Given the constraints imposed by the marginal distributions, we do not expect a coupling supported purely on this graph, but the best possible coupling can still be found, see fig.~\ref{fig:LocConstrRainTemp} (for simplicity, we take uniform distributions for the classifications there).

In this way, more complex logical dependencies can also be modelled probabilistically, see Appendix Section~\ref{sec:LogConstr}, adding noise if desired.

\subsubsection{Modelling Functional Dependencies}\label{sec:ModFuncDep}
To describe noisy functional dependencies, it suffices to write a functional description $F(t)$ on the DEPENDENCIES arrow; this description is then depicted graphically, and a further parameter, \textit{noise}, models the noise level.

The classes involved, however, do not always have continuous value ranges but often categorical or mixed forms, as with rainfall.
In that case we typically have no functional description of the probability distribution, though sets of classes often carry a natural order.
Consider, for example, the types of railway line: urban, regional, freight, and high-speed -- the latter always permitting higher speeds.
Independently of this order, the classes are indexed from top to bottom in the graphic: the topmost class is indexed with $0$, the one below with $1$, and so on, see fig.~\ref{fig:railroadtrack}.
This indexing lets us describe the occurrence probabilities of the individual classes as a list of nonnegative numbers, which are normalized and assigned to the classes in order of appearance, read as probabilities.

With this indexing in place, functional dependencies can again be described functionally, now with $F$ defined on ranges of natural numbers -- the index ranges of the classifications.
In equation \ref{equ:TransProb}, the distance function $dist(F(i),j)$ is then defined with $i$ indexing the source classes and $j$ the indices of the dependent target classification, and the coupling is generically unique.
Noise can again be added, using the same notation as in the continuous case.

Alternatively, the map can be given explicitly by a list $l$, defining the function by $\iota\mapsto l[\iota]$.

\begin{figure}[htbp]
  \centering
  \includegraphics[width=0.8\textwidth]{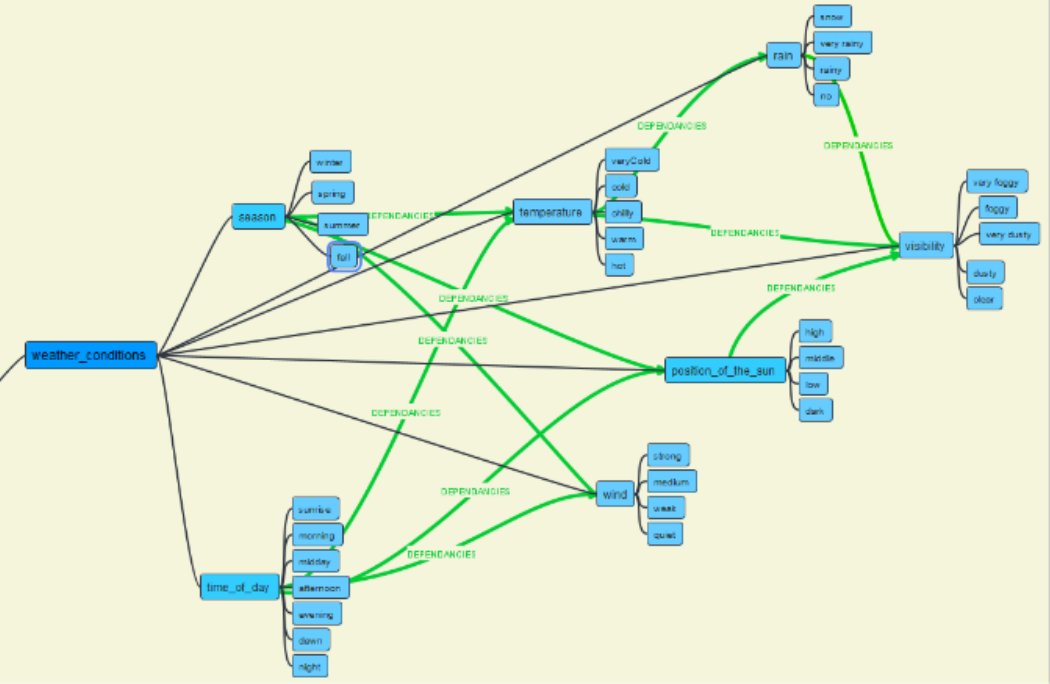}
  \caption{PEON for weather conditions}
  \label{fig:WeatherConditionsPEON}
\end{figure}

We apply maps of subsets to generalise such descriptions and specify logical constraints.
For example, in Section \ref{sec:LogDep}, this looks like the logical constraint $(\{snow\}\Rightarrow\{very\ cold, cold\})$. See Fig.~\ref{fig:WeatherConditionsPEON} and Appendix Section~\ref{sec:LogConstr}. A list of logical constraints can be given in the same notation, even in a higher-dimensional case.

More often than not, we face two or more dependencies rather than a single one.
Temperature, for instance, depends not only on the time of year but also on the time of day.
For simplicity, we assume seasons and times of day vary independently of each other; from the point of view of couplings, this means taking the product measure of their two marginal distributions on the product space -- a trivial coupling that simply models their stochastic independence.
The dependency of temperature can now once more be described functionally, this time via a function of the two parameters \textit{season} and \textit{time of day}.
The term \textit{reparametrization} is used for this function $F$, with the resulting dependency indicated as in fig.~\ref{fig:WeatherConditionsPEON}.

\subsection{Railroad Track}\label{sec:railwayTracks}
We began our investigation with the attempt to formally capture an ODD for autonomous driving, and to extend it with occurrence probabilities for the sceneries in the ODD, so as to provide a statistical basis for quality assurance.
As mentioned in Section~\ref{sec:SimpVersATO}, the ODD for ATO in the railway domain is centred on the track.
A track can be conceptualised as a sequence of sections, each with a defined length and curvature, directing the train to move left or right according to its orientation.
The specific details are not what matters here; rather, the fundamental observation is that a track consists of a finite number of sections, delineated by their length, curvature, and orientation, see fig.~\ref{fig:RegionTrack}.
In this way we obtain an ontology of railway tracks.

We designate the different sizes abstractly, as \textit{short, medium, long}, etc.
Behind these abstract names, however, lie concrete intervals with units -- the corresponding length and curvature-radius intervals, which we annotate at the respective nodes.\label{sec:Abst2Conc}
For a conceptual simulation, we proceed in two steps: first, we sample an abstract description, such as \textit{5 sections, [(short, curved, left),(long, almost\_straight, left),\dots]}; then, according to the \textbf{uniformity hypothesis} of Section~\ref{sec:UniHyp}, we uniformly select a concrete length from the real interval underlying \textit{short}, and so on.
In this way we obtain concrete line segments with definite lengths, radii of curvature, and orientations, from which unique tracks can be computed by various algorithms -- concatenated clothoids or spline interpolation, for instance.
We do not elaborate on this step further, as the transition from abstract to concrete data is highly application-specific.

\begin{figure}[htbp]
  \centering
  \includegraphics[width=0.8\textwidth]{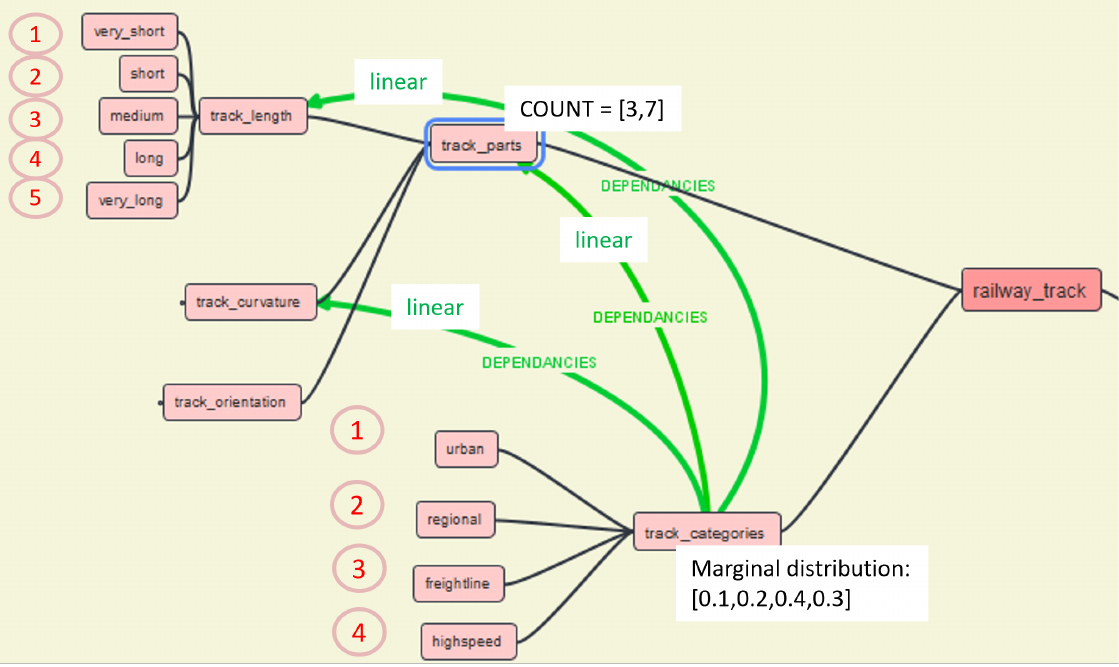}
  \caption{Section of a \textit{probabilistically extended ontology} for railway tracks}
  \label{fig:railroadtrack}
\end{figure}

There are various categories of railway line -- urban, regional, and high-speed -- whose distribution we obtain from railway institutes.
We annotate all marginal distributions at the corresponding classification node.
The possible characteristics are indexed along their natural order: \textit{very\_short}$\rightarrow 1$, \textit{short}$\rightarrow 2,\dots$, and similarly \textit{urban}$\rightarrow 1$, \textit{regional}$\rightarrow 2,\dots$.
The occurrence probability is then annotated at the classification node, see fig.~\ref{fig:railroadtrack}, node \textit{track\_categories}, for example: the \textit{urban} category, with the first index, corresponds to probability $0.1$, the \textit{freightline} category, with the third index, to $0.4$, and so on.

Assuming we again obtain statistical data on the distribution of curvatures and lengths from a railway institute, we annotate them at the classification nodes \textit{track\_curvature} and \textit{track\_length}.
Given the symmetry of running back and forth, we take the orientation to be uniformly distributed -- $0.5$ each for a left- and a right-hand curve.
This completes the specification of the occurrence probability of a \textit{track\_part}.

The track is composed of $3$--$7$ sections, annotated by \textit{COUNT} $= [3,7]$, see fig.~\ref{fig:railroadtrack}; for simplicity, we model the distribution of \textit{COUNT} as uniform.
This completes the specification of the marginal distributions for \textit{railway\_track}, see fig.~\ref{fig:railroadtrack}.

Next, consider the dependencies.
High-speed lines typically comprise long sections with minimal curvature, whereas urban lines feature comparatively short sections with sharper curves -- the curvatures and lengths of the track sections thus depend statistically on the previously selected route category.
To model this dependency, we exploit the following observation:
the route categories carry a natural order, \textit{urban, regional,\dots, highspeed}, and so do the curvatures and lengths, \textit{highly curved,\dots, almost straight} and \textit{very short,\dots, very long}.
This lets us describe the statistical dependency informally as linear: the higher the index of the category, the more probable a higher index of curvature, see Section~\ref{sec:ModFuncDep}.
In the terminology of Section~\ref{sec:FuncDep}, their coupling should exhibit a noisy linear functional dependency, which we note simply by marking it \textit{linear}, see fig.~\ref{fig:railroadtrack}.
Adding noise would, of course, be possible if desired.

\Nb \label{sec:GenReparam}
As in Section~\ref{sec:ModFuncDep}, we have modelled various marginal distributions by listing occurrence probabilities according to the class indexing.
Equivalently, these can be described by relative occurrence probabilities, normalized in the background.

As noted in Section~\ref{sec:ModFuncDep}, we may specify value intervals behind abstract names, as illustrated on p.~\pageref{sec:Abst2Conc}.
In the absence of relevant data -- on reputable websites or in statistical references -- the central limit theorem, see \cite{Dudley.}, or other theoretical results from statistics, see \cite{Bijma.2017}, can be used to obtain a reliable estimate of the distribution of continuous variables.
The volume associated with the variables then indicates their probability of occurrence, and this specification is annotated at the classification node; the effective discrete distribution is computed automatically in the background.
\Ne

\begin{figure}[htbp]
  \centering
  \includegraphics[width=0.4\textwidth]{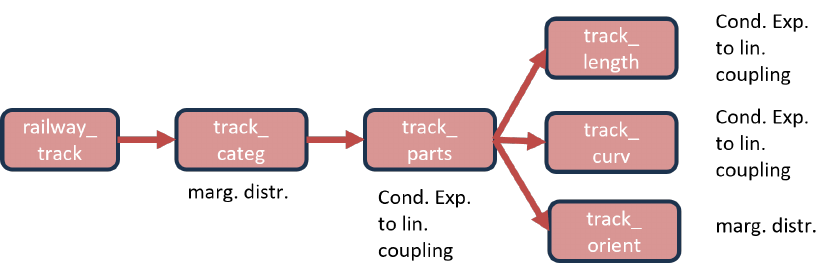}
  \caption{Bayesian network for the subtree PEON \textit{railway\_track}}
  \label{fig:BayesNet}
\end{figure}

We can now use this information to specify a Bayesian network uniquely.
We take \textit{railway\_track} as the root node, with child nodes \textit{track\_categories} and \textit{track\_parts} -- the latter depending on the selected category.
We therefore select \textit{track\_categories} as the next node in the Bayesian network below the root, and \textit{track\_parts} below it.
The distribution at \textit{track\_categories} is the specified marginal distribution; for \textit{track\_parts}, we define the conditional probability, via Bayes, from the previous selection of \textit{track\_category}, and similarly for the remaining nodes, see fig.~\ref{fig:BayesNet}.

\begin{figure}[htbp]
  \centering
  \includegraphics[width=0.8\textwidth]{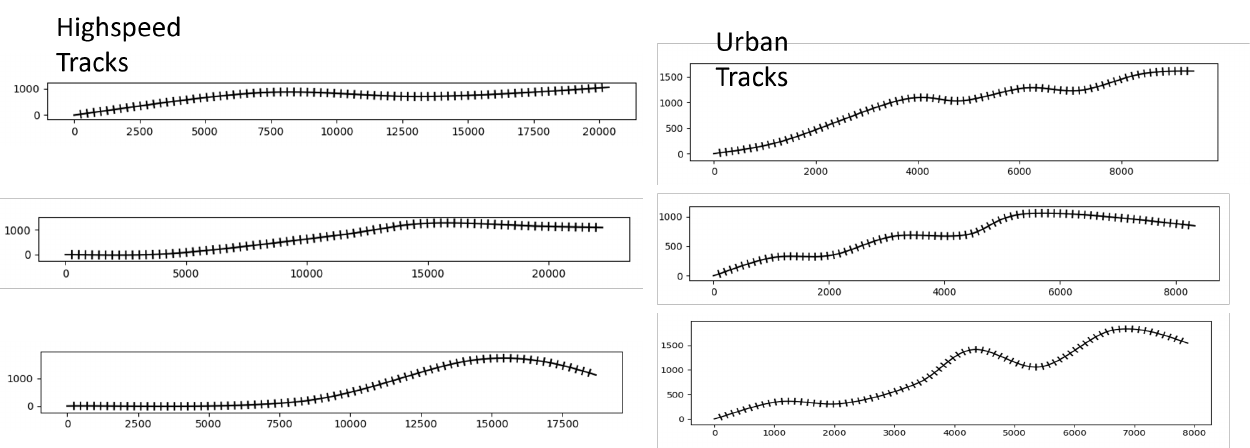}
  \caption{Typically sampled high-speed and urban tracks -- note the different scales!}
  \label{fig:HighUrban}
\end{figure}

With this Bayesian network we have uniquely specified a probability measure on the \textit{railway\_track} subtree, and we can sample railway tracks accordingly.
We implemented a conceptual simulation rendering the abstract samples into concrete railway tracks, connecting the parts by clothoids, see fig.~\ref{fig:HighUrban}.

This section has shown how the set of railway tracks can be described in a discrete, finite manner using the concepts of a PEON, and how the method can be used to sample abstract track descriptions and render them into concrete, simulatable form.

\subsection{Roads Crossing the Track}\label{sec:RoadCrossing}
\begin{figure}[ht]
  \centering
  \begin{minipage}{0.45\textwidth}
    \centering
    \includegraphics[width=\textwidth]{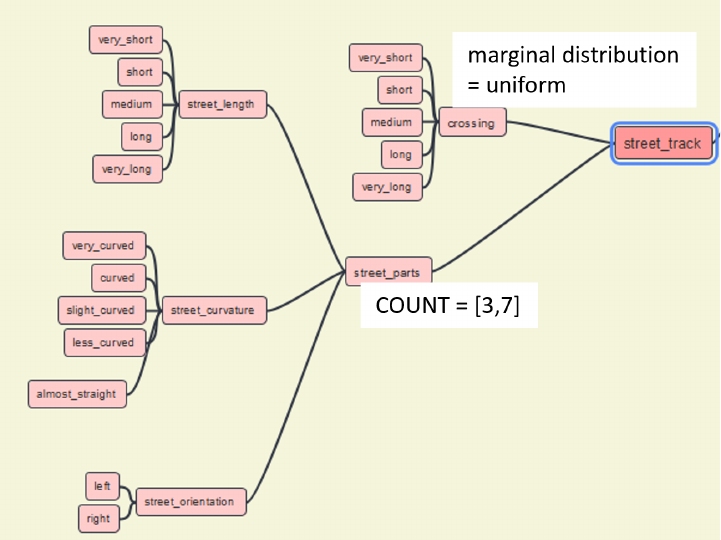}
    \caption{PEON for streets crossing the track}
    \label{fig:PEONRoads}
  \end{minipage}\hfill
  \begin{minipage}{0.45\textwidth}
    \centering
    \includegraphics[width=\textwidth]{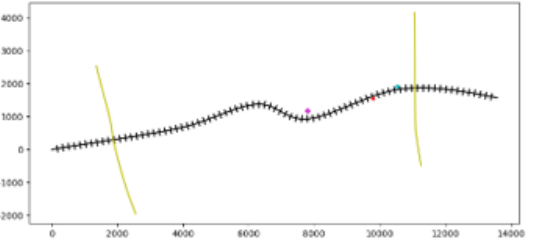}
    \caption{Simulation of the track with two crossing streets}
    \label{fig:TrackSigsRoads}
  \end{minipage}
\end{figure}

A notable challenge for ATOs is the crossing of roads.
We therefore define an ontology for road routes, closely resembling that of the railway tracks; the only additional element is the crossing with the railway track itself (\textit{crossing}).
Since the ego train moves along the railway track, we again assume, by symmetry, that the crossing distances are uniformly distributed, see fig.~\ref{fig:PEONRoads}; for clarity, we have dispensed with the various possible dependencies in this ontology.

This again uniquely specifies a probability distribution over streets crossing the track, from which we can sample abstract data and render it into concrete form, see fig.~\ref{fig:TrackSigsRoads}.

\subsection{Signals along the Track}\label{sec:signals}
\begin{figure}[ht]
  \centering
  \begin{minipage}{0.45\textwidth}
    \centering
    \includegraphics[width=1.0\textwidth]{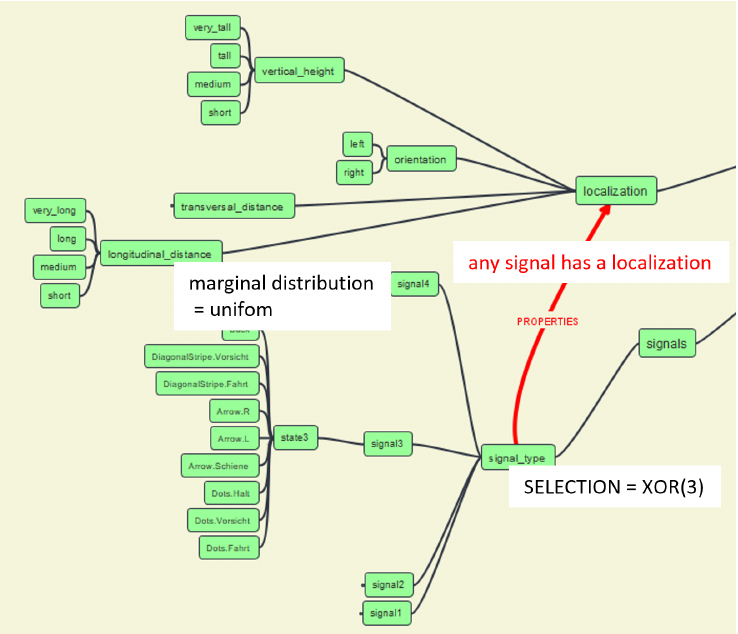}
    \caption{PEON for trackside signals}
    \label{fig:signals}
  \end{minipage}\hfill
  \begin{minipage}{0.45\textwidth}
    \centering
    \includegraphics[width=0.6\textwidth]{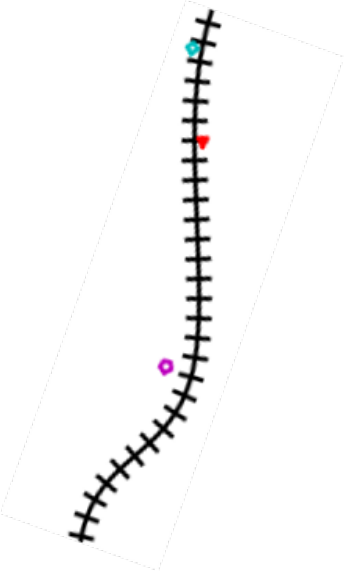}
    \caption{Detailed presentation of signal types and states}
    \label{fig:SignalsDetails}
  \end{minipage}
\end{figure}

A pivotal function of object detection systems, such as those used in ATO, is the accurate identification of signals and their statuses along the route.
Let us therefore look at how signals along the railroad line can be modelled.
There are several distinct signal types, including the dwarf signal and the Andreas cross, each closely tied to different states and to a location relative to the ego train, see fig.~\ref{fig:signals}.
We use abstract identifiers \textit{signal1, signal2,\dots} in what follows.

In this example, we consider four possible signal types, of which only three are actually to be expected at any given position.
This is specified in the node text and annotated accordingly:
\textit{SELECTION} $=$ XOR$(3)$ selects three of the four child nodes, as illustrated in fig.~\ref{fig:signals}, using the uniform distribution over the resulting subsets.
(The number of signals could, in principle, depend on the route category, though this is not modelled in the present example.)

Furthermore, every selected signal must be localized along the track.
This is annotated via a link, \textit{PROPERTIES}, stating that a localization must be sampled for any selected child of the node.
Since a localization is required not just for signals but for a variety of objects along the route, it is convenient to consolidate this attribute into a single subtree, rather than documenting it independently under each entity -- a common convention in ontologies, see fig.~\ref{fig:signals}.
The \textit{PROPERTIES} link type was introduced precisely to represent this graphically.

As the ego train moves along the track, the relative position of the signals changes continuously; we therefore assume the longitudinal spacing of the signals to be uniformly distributed, see Section~\ref{sec:RoadCrossing}.

We do not elaborate further here, as the procedure parallels that of the railway track: a sample can be obtained from this abstract setting by placing signals along the track, and the data is subsequently rendered into concrete form.
As illustrated in fig.~\ref{fig:SignalsDetails}, different signal types are represented by different geometric figures, and different signal states by different colours.

\subsection{Points of Interest}\label{sec:pois}
Of course, a railway line involves more than signals and road crossings -- there are houses, train stations, and forests, for instance.
Here we focus, by way of example, on stations and forests along the line.
For simplicity, we characterise a station solely by the length of its platform, since this correlates directly with the time the train needs to pass it.
Forests, in turn, are characterised by two factors: their extent -- given by the \textit{height} and \textit{width} of the wooded area -- and their density, measured as the number of trees per unit area (\textit{n\_trees}).

\begin{figure}[ht]
  \centering
  \begin{minipage}{0.45\textwidth}
    \centering
    \includegraphics[width=\textwidth]{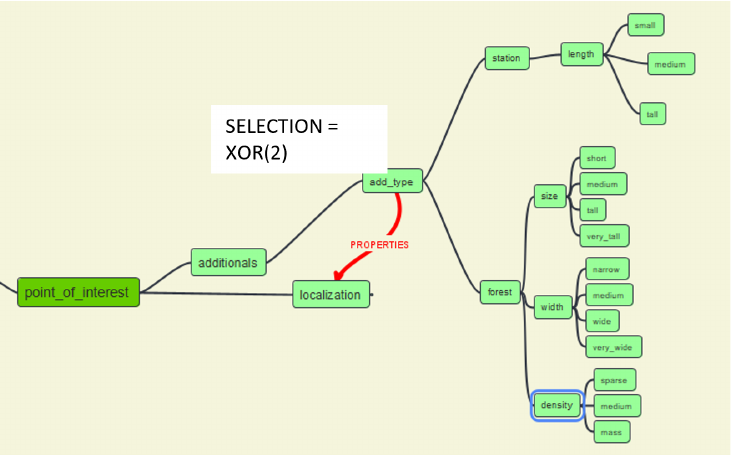}
    \caption{PEON for point of interest}
    \label{fig:PEONPoi}
  \end{minipage}\hfill
  \begin{minipage}{0.45\textwidth}
    \centering
    \includegraphics[width=\textwidth]{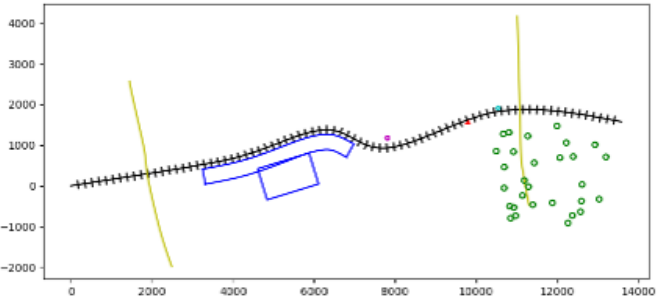}
    \caption{Simulated track with a station (blue) and a wood (green)}
    \label{fig:TrackPois}
  \end{minipage}
\end{figure}

As before, the \textit{PROPERTIES} link denotes the location of every point of interest, ensuring that it is correctly placed relative to the route, as in Section~\ref{sec:signals}.
The height of an object -- forest or station alike -- is a further factor we disregard in what follows.

Further entity types can be added as children of a dedicated \textit{add\_type} node.
To select among such typed objects, the node carries \textit{SELECTION} $=$ XOR$(n)$ for an integer $n$, see Section~\ref{sec:signals}, and, to fix the number of instances entering the scene, a node labelled \textit{COUNT} $=[k]$, see Section~\ref{sec:railwayTracks}; neither is used in the present example, see fig.~\ref{fig:PEONPoi}.

As before, we incorporate marginal distributions from internet sources or reputable institutions, and model dependencies where necessary.
The result is a PEON from which a Bayesian network, and thus a probability distribution over the \textit{point\_of\_interest} subtree, is derived, see fig.~\ref{fig:PEONPoi}.

\subsubsection{Putting it Together}\label{sec:PuttingTogether}
\begin{figure}[htbp]
  \centering
  \includegraphics[width=0.8\textwidth]{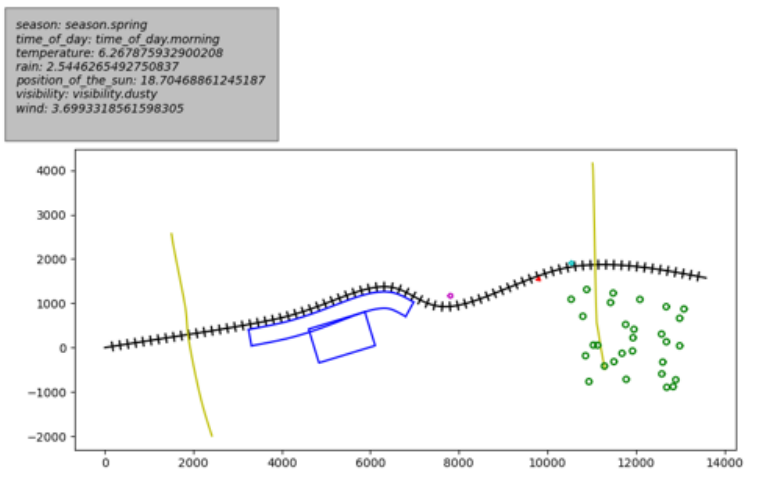}
  \caption{Sampled weather, annotated at the top of the figure}
  \label{fig:TrackWeather}
\end{figure}

We have used the framework above to build up a number of subordinate branches -- \textit{weather\_conditions, railway\_tracks,\dots} and so on -- each with its supplementary statistical data, together forming the PEON that captures the statistical interrelationships within the ODD for our railway example.
The preceding sections outlined the computation of the corresponding Bayesian networks and their use in sampling.
As illustrated in fig.~\ref{fig:TrackWeather}, the resulting model can be understood as a scene rendered randomly from the abstract data.

\subsection{Actors}\label{sec:Actors}
Here we introduce only the most elementary actors: the ego train and cars.
The fundamental principle governing their movement is that they traverse curves -- in this case, along the designated tracks or roads.
In the general case, e.g.\ for pedestrians, suitable curves would need to be generated for each actor individually, in a manner resembling the construction of railway and street tracks, albeit somewhat more intricate; we leave this to the interested reader.

For simplicity, we consider only the ego train, moving along the rails, and cars, moving along the roads, each at a certain speed along its route.

The movement profiles are simplified further by assuming constant velocities over given time intervals, omitting any acceleration phase.
More realistic motion profiles could be obtained via suitable interpolation methods, or quite general stochastic processes.
Rather than specifying speeds directly as absolute values, we first select whether a speed is \textit{very\_fast, fast,\dots}; and rather than choosing distinct durations for each velocity, we select, for illustration, a single number from $[1,n]$ and divide the scenario's total time interval into $k$ equal parts.

To model different types of objects moving on the road, we add a \textit{color} attribute, which so far has no effect beyond colouring the conceptual simulation.

\begin{figure}[ht]
  \centering
  \begin{minipage}{0.45\textwidth}
    \centering
    \includegraphics[width=\textwidth]{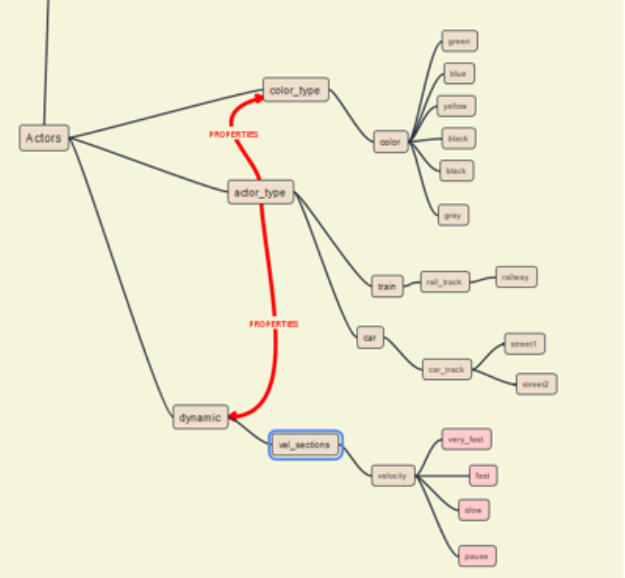}
    \caption{PEON for actors}
    \label{fig:PEONActor}
  \end{minipage}\hfill
  \begin{minipage}{0.45\textwidth}
    \centering
    \includegraphics[width=\textwidth]{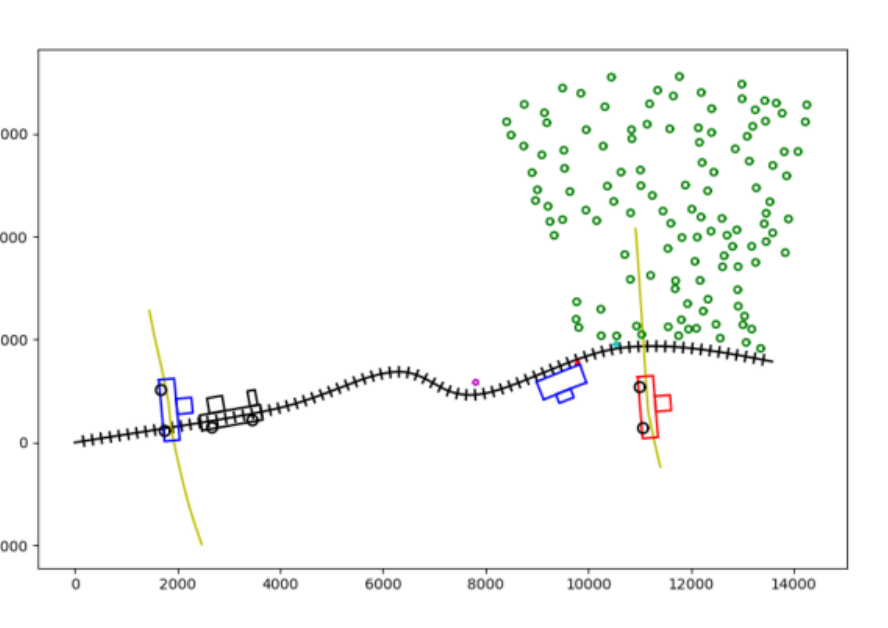}
    \caption{Simulation with an ego train,two cars, a few signals (coloured figures close to the track), a station and a wood}
    \label{fig:railwaytrack}
  \end{minipage}
\end{figure}

The velocities are then recalibrated so that each actor starts at the beginning of its trajectory and arrives at its end.
While these simplifications are evidently coarse, they already yield a range of interesting general scenarios, see fig.~\ref{fig:PEONActor}; a more realistic treatment of the actors is left to the interested reader.

Using the \textit{street\_tracks} subtree of Section~\ref{sec:RoadCrossing}, we add two roads crossing the railway track, \textit{street1} and \textit{street2}, fixing the trajectories of our three actors: the ego train along the railway track, and cars~1 and~2 along street1 or street2, respectively.
This could, of course, be modelled more dynamically and generally.

Once more, the \textit{PROPERTIES} link is used to attach a specific \textit{dynamic} profile, as well as a \textit{color\_type}, to any given actor, see fig.~\ref{fig:PEONActor}.

In the model considered here, all dependencies between the actors have been neglected -- for instance, that a vehicle typically slows down the closer it gets to a crossing.
(Modelling this would be a design decision, best captured by interpolation or a stochastic process; here, an interpolation would likely reflect driver behaviour more faithfully.)
For the marginal distributions, we simply take the uniform distribution.

The chapter provided a comprehensive overview of the methodology for creating a PEON, incorporating probabilistic information, and introducing several constructs for probabilistically modelling even highly complex relationships and ODDs in a manageable and maintainable way.
In the final section, these approaches will be linked to possible quality assurance processes.

\section{A Proposal for a Systematic Test Process}\label{sec:SysTP}
\begin{figure}[htbp]
    \centering
    \includegraphics[width=0.7\textwidth]{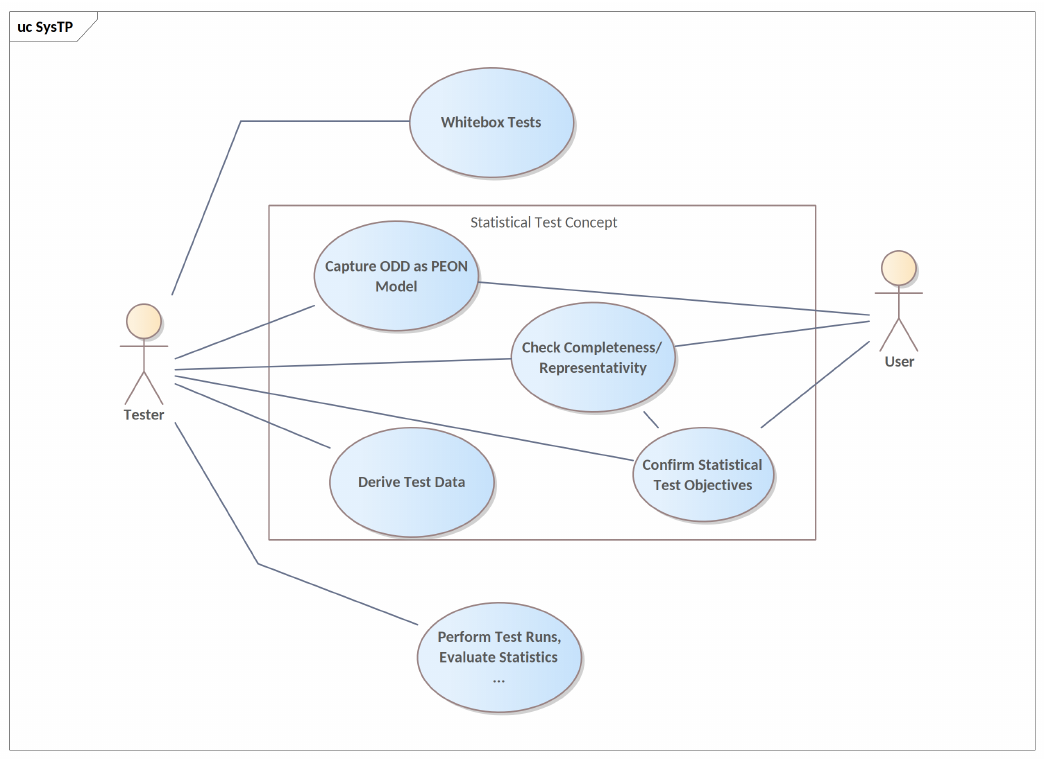}
    \caption{Various use cases relevant to AI testing.}
    \label{fig:SysTPUse}
\end{figure}

Reaching an understanding with the future user about the scope and quality of an application is essential for sound and efficient development -- and providing evidence that this quality has been achieved is equally urgent.
For safety-critical applications, standards-based quality criteria must be met as well, to ensure the system's future usability.
A detailed analysis of the intended scope of application and of the ODD is therefore essential; for AI systems, the required statistical quality criteria must be considered in addition.
A PEON model of the ODD supports exactly this understanding: its graphical representation makes it accessible even to users without a technical background, and ensures that the underlying decisions and modelling choices remain transparent.
It can thus serve as a suitable, shared baseline model for describing the ODD, see fig.~\ref{fig:SysTPUse}.

As in the systematic testing process for classical systems, see \cite{GrimmGrocht}, potential implementation weaknesses should first be examined through whitebox testing, in coordination with the user.
Subsequently, the system should undergo exhaustive blackbox testing to verify its functionality; for an AI application, this means conducting statistically significant tests.
To this end, a complete and representative PEON model of the ODD must be created, and close collaboration with future users is strongly recommended to verify that the model is indeed complete and representative.
Our experiments have shown that a conceptual simulation of exploratorily derived test cases is very helpful in building this shared understanding, and we recommend it accordingly.
To this end, the abstract, generated test cases are first enriched with concrete data -- for example, a \textit{fast} velocity $v$ ($\widehat{=}\ [50\ km/h,\ 150\ km/h]$) is sampled uniformly to $v=87\ km/h$.
In this way we obtain concrete test cases that can be simulated conceptually, so that the various scenarios can be conveyed using simple graphical means, see fig.~\ref{fig:railwaytrack}.

Once a PEON model of the ODD has been agreed upon with the various stakeholders, specific testing objectives can be defined:
What level of significance should the test results have? What ethical requirements must they satisfy? Which aspects require particularly thorough verification?
This also allows for a more accurate estimate of the testing capacity required to provide validation.

Subsequent to the attainment of these results, the actual testing process can be initiated.
Firstly, abstract test cases are derived from the PEON model in accordance with the established criteria. The number of tests is determined by their required significance (see Section ~\ref{sec:SampleSize}), along with any specific requirements, such as ethical considerations (see Section ~\ref{sec:DethicalDemands}), and any restrictions on the appropriate conditional distributions from which they are to be generated.

In a subsequent step, the abstract test cases derived from the PEON are expanded into concrete test cases. Depending on the test environment, these are then further processed, for example rendered to realistic images, until they can finally be applied to the system under test.
The derivation of test cases from the validated PEON ensures the applicability of statistical methods for their evaluation.

\begin{figure}[htbp]
    \centering
    \includegraphics[width=0.7\textwidth]{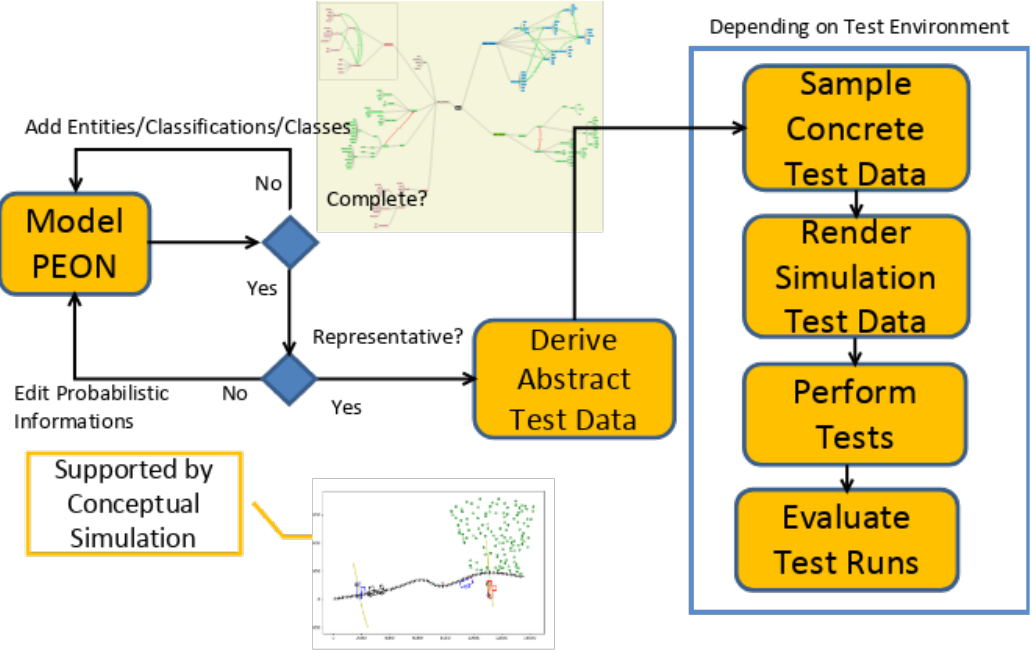}
    \caption{Various activities required for statistical testing.}
    \label{fig:SysTP}
\end{figure}

Once all tests have been run, the results can be evaluated against the agreed-upon statistical quality criteria.
This yields clear metrics reflecting the statistical quality of the AI, and with them, clear criteria for assessing whether the requirements have been met.

These concepts were implemented on an exploratory basis in the AI-LOK research project.
A rudimentary PEON model was formulated for the railway sector, from which abstract test cases were derived and subsequently augmented with concrete data, enabling their conceptual simulation.
These concrete test cases proved instrumental for the effective realization of the test cases, which were then rendered and visually implemented in suitable simulators -- an approach that allowed the test cases to serve as prototypes for testing an object detection system in the railway sector.
For further information, see \cite{TDGenRailwayDomain}.

\subsection{Conformance Test of Training Data}\label{sec:TestTD}
It should be briefly noted that this approach can also be used to evaluate the quality of training data:
its quality can be assessed by comparing its distribution statistically against the one defined by an agreed-upon PEON.
This requires that both distributions reside in the same space.
Ideally, we select a sufficient number of training data points and label them as finely as our PEON requires, i.e.\ assign each to a partition of the tuned PEON.

This is very time-consuming, however, and may not even be necessary if only a rough comparison is intended.
In that case, we coarsen the PEON -- merging classes, for instance $\{light\ green\},\{green\},\{dark\ green\} \Rightarrow \{green\}$, or even entire classifications, $\{dog\},\{cat\}\Rightarrow \{animal\}$, see Section~\ref{sec:Refinements} -- and compute the corresponding coarsened distribution by integration.
The randomly selected training data then only needs to be labeled against this coarser classification.

Once these preparations are complete, evaluating the training data amounts to testing whether the training data sample conforms to the (coarse-grained) PEON distribution.
Depending on the data type, various classical tests can be used for this purpose, see \cite{Bijma.2017}.

\section{Conclusion and Outlook}\label{sec:Conclusion}

This technical report has introduced the concept of a \textit{probabilistically extended ontology} (PEON). With its help, even highly complex environments -- such as operational design domains (ODDs) -- can be modelled statistically.

We start from an acyclic graph, an ontology that captures the entities of an environment together with their properties and instances, and extend this model by descriptions of occurrence probabilities and probabilistic dependencies. Consider, for example, how often a person appears in a scene, and note that if this person is carrying an umbrella, it is likely raining. Modelled graphically, such a description remains readable even without in-depth prior knowledge. From it, we derive a Bayesian network, which serves as our statistical model, and with its help we can sample arbitrary abstract, ontological descriptions of the environment.

To demonstrate the scalability of our approach, we chose the example of a simple railway environment. To this end, various route configurations were modelled probabilistically, along with trackside signals and further points of interest -- such as forests or train stations -- that may occur along the route. The underlying weather conditions, too, were modelled probabilistically. Besides such static objects, the approach also allows us to model actors, e.g.\ vehicles on intersecting roads, see Section~\ref{sec:SimpVersATO}.
As a case in point: a railway track, complete with a train and two crossing roads with cars in motion -- a rudimentary scenario, albeit merely sketched out, see fig.~\ref{fig:railwaytrack}.

The motivation for this new concept was the question of how classification systems, and feedforward networks more generally, can be tested. Consider, as an example, the task of testing an AI-based obstacle detection system for the railway domain. To define and verify suitable quality criteria for such a system, a sound statistical foundation must first be established for the various metrics, such as accuracy and precision. Every statistic, after all, refers to a well- and clearly-defined set -- so this set must first be described in detail. This is exactly what a PEON provides: as an ontology, it captures all entities and their properties as they may occur in the environment; as a derived statistical model, the Bayesian network; and, taken together, the complete statistical basis against which the results of our tests can be evaluated.

Moreover, the same model can be used to specify abstract test cases. Agreed-upon test criteria and statistical metrics determine both the minimum number of test cases derived from the abstract ones that need to be executed and the required success rate, see \cite{TDGenRailwayDomain}.

It should have become evident that the approach also generalizes to other areas in which system quality is to be evaluated statistically, by tests or other knowledge-based methods.
Wherever a comprehensive and lucid understanding of the statistical interrelationships within a specific domain is imperative, this methodology can be employed to articulate and identify appropriate products.

As a next step, the new technique will be examined in depth on a concrete AI test example, followed by a comparative analysis against conventional methods.

\section*{Acknowledgments}
I would like to take this opportunity to thank Sadegh Sadeghipour and Nicolas Grube for their support and tireless interest.
This work was conducted as part of the \textit{SysTestKI }research project (16KN117756), which was funded by the Federal Ministry of Research, Technology and Space.

\bibliographystyle{unsrtnat}  
\bibliography{ProbModellingODD.bib}  


\section{Appendix}\label{sec:Appendix}
\subsection{Modelling Tool, Freeplane}
``A picture is worth a thousand words'' -- the adage is often cited to emphasize the value of visual communication, and graphical representations are indeed a natural choice for documenting, describing, and modelling complex statistical relationships.
The individual selection options (classifications) are depicted by nodes, their concrete characteristics (classes) by child nodes.
Behind the selection options lie probability distributions recording the occurrence probabilities of the individual characteristics; the values can be listed explicitly or described functionally.
Behind the abstract identifiers of the characteristics, in turn, there may be concrete values or value ranges.

To facilitate the transition to Bayesian networks for computing complex probability densities, it is helpful for the nodes to be structured as a tree or, more generally, a forest.
Probabilistic dependencies are represented graphically by links, and each link carries a description of how the dependency is modelled -- so all of this information must be attachable to the link as well.
Since relationships between entities, such as PROPERTIES, are also represented by arrows, the modelling tool must additionally allow custom link types to be defined and displayed in a visually distinguishable way.

To summarise, within the context of graphical modelling, the capacity to assign diverse attributes to the constituent nodes, and the links between these nodes that are designated by specific typologies, must be a possibility.

A graphical tool must meet these requirements to be usable in the context of PEONs.
Fortunately, \textit{Freeplane} fulfills all of them and more, see \cite{Freeplane.17.04.2025}.
In particular, it stores its model data as XML, which makes the models easily accessible from Python and other programming languages.
We used Freeplane for our modelling.

\subsection{Python Implementation}
In our implementation, the description of probability densities is achieved through the utilisation of tensors.
The programming language Python has firmly established itself as a prominent tool in the field of AI research, particularly in conjunction with libraries such as TensorFlow \cite{TensorFlow.09.12.2025} and PyTorch \cite{PyTorch.09.12.2025}, has firmly established itself as the tool of choice.
Our library is compatible with both the implementation of tensors in NumPy and TensorFlow and PyTorch.
The standard python library \textit{graphlib} provides an implementation of Kahn's algorithm for building a Bayesian network.
The \textit{POT: Python Optimal Transport} module supports the computation of suitable couplings from functional descriptions, see \cite{POT.18.05.2025}.

For parsing a PEON model, the utilisation of the python library \textit{lxml} facilitates efficient processing due to the straightforward XML structure of the storage model, see \cite{LXML.18.05.2025}.
The underlying forest is then topologically sorted using Kahn's algorithm; the resulting node order determines the topology of the associated Bayesian network. The calculation of the conditional expectations is achieved by employing the Bayes rule (see equ. \ref{Equ:BayesProdRule}), thus resulting in the construction of a complete Bayesian network.

\subsection{On some Basics in Probability Theory}\label{sec:Probability}

Probability theory allows us to analyze uncertain data and to draw valid conclusions from it, even where values are imprecise.
Uncertainty in data can have a multitude of causes, among them:
\begin{itemize}
	\item the data under analysis is subject to stochastic disturbances, such as measurement noise;
	\item only limited aspects of the phenomenon are observable -- one cannot, for instance, measure electrical currents in the human brain directly and has to resort to functional magnetic resonance imaging (fMRI);
	\item the modelling is incomplete, a notorious challenge for complex subjects such as economics or behavioural research.
\end{itemize}
This list could easily be continued. Modelling a complex ODD certainly belongs to it.

We assume that the reader possesses basic knowledge of probability theory; for an introduction see \cite{Pollard.2001}.
In this section, probability spaces are denoted by capital letters ($X, Y, Z$), the sigma algebra of measurable sets by $\Omega$, and measurable sets (events) by capital letters ($A,B,\dots \in \Omega$).
Points or elements are denoted by $x_1,x_2,\dots \in X$.
Probability measures are denoted by the letter $p$, indexed by the underlying set where helpful, e.g.\ $p_X,p_Y$, while measurable functions are denoted by lowercase letters such as $f, g$.
A function $f$ is called image measurable if the images of measurable sets are again measurable.

For probability spaces over an Operational Design Domain (ODD), the event spaces are typically discrete and finite.
The collection of all subsets then constitutes the corresponding sigma algebra, and any mapping between event spaces is trivially measurable as well as image measurable.
Furthermore, we assume that all events in the probability model of an ODD have probability $>0$:
if this is not the case for some event, the event does not occur in the ODD and is to be disregarded.

Since our objective is to model the probabilities of situations and events, rather than to record them descriptively, we adopt a Bayesian perspective on probability, see \cite{jaynes2003probability}.

A brief preliminary remark is in order.
Probability theory can be developed on very general spaces, see \cite{Bogachev.2007}, where probability distributions are referred to as measures.
In this article, the terms ``distribution'' and ``measure'' are used interchangeably: for finite sets with the power set as sigma algebra, the more sophisticated theory collapses.
In this case, a probability distribution can be described as a list of nonnegative real numbers indexed by the elements of the underlying set.
This list can equally be read as a measurable function on the finite probability space, namely as the density of the distribution with respect to the counting measure -- and the counting measure itself corresponds, after normalization, to the uniform distribution on the finite set.
A measurable set $A\in \Omega_X$ is simply any subset $A\subset X$, and the integral over a probability measure becomes a plain sum; we use both symbols, $\int$ and $\sum$, interchangeably.
In the finite setting, many distinctions thus disappear, which simplifies the overall picture considerably.

\subsubsection{Conditional probability (Bayes' theorem)}
As stated in Section~\ref{sec:Probability}, we employ the Bayesian perspective on probability theory:
probability expresses the degree of belief that certain events occur.
The crucial question is how this probability changes in the presence of prior knowledge.

\Ex \label{Ex:Bayes}
The prevalence of a specific disease is $10^{-5}$; one in every hundred thousand individuals is afflicted.
Suppose a test decides with error probability $10^{-4}$ whether an individual has the disease: an afflicted person receives a false negative result with probability $10^{-4}$, an unafflicted person a false positive with the same probability.
Now suppose the test comes out positive.
What is the probability of actually having the disease?

Answer: approximately $10\,\%$ -- about one in ten.

Why is that?
Imagine the test carried out on one million people.
About $10=\frac{10^6}{10^5}$ of them are affected by the disease, and essentially all of them will test positive.
The vast majority, roughly $10^6 - 10 = 999{,}990$ people, do not have the disease -- but about $10^2\sim\frac{10^6 - 10}{10^4}$ of them will nevertheless receive a positive result.
A person with a positive test result is therefore one of roughly $110$ positives: one of the $\sim 100$ healthy or one of the $10$ afflicted.
Hence the probability of having the disease is about $\frac{10}{110}\approx 10\,\%$.
\Exa

The theoretical foundation of this -- at first sight perplexing -- example lies in the prior knowledge:
the prevalence of the condition is exceptionally low, one in $100{,}000$, while the error probability of the test, $10^{-4}$, is not negligible against it.

The event space combining test result and disease status is the 2-dimensional Cartesian product $X\times Y$:
$X$ may be taken as a binary variable indicating the test result, $Y$ as one indicating whether the individual has the disease, with $\dot=1$ standing for a positive result and $\dot=0$ for a negative one.
The probability measure on this space is denoted $p_{X\times Y}$.
Integrating out one variable yields the \textbf{marginal distributions}:
the probability of $x \in X$ without any prior knowledge about $Y$ is given by
$$
p_X (x):= \int_Y p_{X\times Y}(x,y) \ \ \mbox{ marginal distribution over X},
$$
and analogously for $y\in Y$.
Conversely, knowing the distribution $p_{X\times Y}(x,y)$, we can assess the probability of $x\in X$ once the value of $y\in Y$ is known -- as in the example.
This brings us to the central theorem of Bayesian theory.

\textbf{ Bayes' theorem: }
On a probability space $X \times Y$, the conditional probability of an event $A\subset X$, given an event $B\subset Y$, is
\be \label{Eq:Bayes}
p_{X\times Y}(A|B)=\frac{p_{X\times Y}(A,B)}{p_Y(B)}=\frac{p_{X\times Y}(A,B)}{\int_X p_{X\times Y}(x,B)}.
\ee
The conditional probability expresses the belief that $x\in A \subset X$ occurs, assuming the occurrence of $y\in B \subset Y$.

$X$ and $Y$ are called stochastically independent if
\be
p_{X\times Y}(A,B)=p_X(A)\cdot p_Y(B) \ \forall A\in\Omega_X,\  B\in\Omega_Y
\ee
in which case we obtain $P(A|B)=P(A)$, as anticipated:
the probability of event $A$ given event $B$ equals the unconditional probability of $A$, since by stochastic independence the prior knowledge of $B$ provides no information about $A$.

Let us re-examine Example~\ref{Ex:Bayes}.
Set A = \{person has disease\} and B = \{test is positive\}.
Then $p_X(A) = 10^{-5}$, the probability of $A$ and $B$ occurring together is $p_{X\times Y}(A,B) = 10^{-5}\cdot(1-10^{-4})$, and the probability of $B$ is $p_Y(B) = 10^{-5}\cdot(1-10^{-4}) + (1-10^{-5})\cdot 10^{-4}$.
Inserting these values into Bayes' formula yields:
\[
\mbox{The probability of the disease being present, given a positive test result, is } 0.0909\dots
\]

Bayes' formula is easily inverted:
given the conditional probability of $A$ under $B$, the conditional probability of $B$ under $A$ is
$$
p_{X\times Y}(B|A)=\frac{p_{X\times Y}(A|B)\cdot p_Y(B)}{p_X(A)},
$$
as is readily verified, see \cite{Pollard.2001}.
Bayes' theorem thus allows probabilities to be computed from prior knowledge and stochastic dependencies; it is the foundation of probabilistic inference, also known as \textbf{Bayes inference}.

When dealing with complex environmental conditions, Bayes' inversion becomes important.
Consider two probabilistically distributed factors: we do not know their concrete values in a specific situation, but we do know their probabilities.
Temperature and precipitation, for instance, are such factors describing the weather conditions of a driving situation.
What does the distribution look like when both factors are considered jointly -- say, what is the probability that the temperature is below $-3\,^{\circ}$C \emph{and} snow is falling?
From weather data we can determine how likely snowfall is, and how likely a temperature of $-3\,^{\circ}$C is; that is, we obtain the marginal distributions of the factors.
But the available tables do not tell us how likely the two are to occur \emph{simultaneously}.
On the other hand, we frequently have an understanding of the probabilistic interrelationship between such factors -- we can estimate the approximate shape of the conditional probabilities; the distribution of temperatures in the event of snowfall, for instance, is easy to conceive.
Requiring certain properties of the conditional probability, informally specified, and resolving these conditions then allows the distribution on the product space to be computed via
\be
p_{X\times Y}(A,B) = p_{X\times Y}(A|B)\cdot p_Y(B) \ \ \forall A\subset X,\ B\subset Y
\ee
In other words: given the marginal distributions $p_X$ and $p_Y$, the objective is to determine $p_{X\times Y}(A|B)$ in order to obtain the joint distribution $p_{X\times Y}$, see section \ref{sec:PEONModelling}.

\textbf{Chain Rule of Conditional Probabilities: }
Bayes' theorem generalizes readily.
Let the event space be the Cartesian product of $n$ event spaces, $X=X_1\times X_2\times\dots\times X_n$, let $p_X$ be a probability distribution on $X$, and $p_{X_i}$ its marginals on the $X_i$.
Then the probability of an event $A_1\subset X_1,...,A_n\subset X_n$ is given by the \textbf{Bayes product rule}:
\be \label{Equ:BayesProdRule}
p_X(A_1...A_n) = p_{X_1}(A_1) \cdot p_{X_1\times X_2}(A_2|A_1) \cdots p_{X_1\times X_2 \cdots X_n }(A_n| A_1,...,A_{n-1})
\ee
where $p_{X_1\times X_2 \cdots \times X_k}$ denotes the marginal measure of $p_X$, i.e.\ $p_X$ integrated over $X_{k+1}\times \cdots \times X_n$.
The result follows by applying Bayes' theorem iteratively to the factors.

This factorization also enables a straightforward sampling procedure for high-dimensional, complex distributions, provided the conditional probabilities are known.

For independent variables (stochastically independent event spaces), the conditional factors reduce to the underlying marginal distributions, and the rule degenerates to the product measure.

The theorem shows its real significance where the distributions are interdependent:
a strategic reordering of the factors can simplify the representation of the distribution considerably.
This is the fundamental observation behind probabilistic graphical modelling, see the next section, \ref{sec:StructProb}.

\subsubsection{Probabilistic Graphical Modeling}\label{sec:StructProb}
Marginal distributions of the individual selection options -- that is, the distribution of each factor considered independently of the others -- can usually be determined without much effort.
The corresponding statistics can frequently be found on pertinent internet pages or in publications of statistical institutes.
Where necessary, they can also be derived from general statistical laws, as special cases of well-known parameterized distribution families such as the Gaussian normal distribution, the Poisson distribution, and others.

The situation becomes more involved when we try to ascertain the occurrence probabilities of \emph{combinations} of different factors.
The statistical interdependence of such factors is best illustrated by a simple example.

\Ex \label{Ex:Weather}
Consider the distribution of factors that influence object detection in a scene.
The position of the sun significantly impacts the visual perception of the scene.
Elevated temperatures can induce shimmering edges, while in colder conditions the air is typically clearer and contours are more pronounced; the coloration of objects changes as well.
And trees and shrubs along the roadside bend in the wind.
\begin{table}[h!]
	\begin{center}

		\caption{weather conditions}
		\label{tab:weather}
		\begin{tabular}{l|c|r}
			\textbf{Factor}     & \textbf{Values}                        & \textbf{Probabilities}     \\
			\hline
			position of the sun & \{dark,  low, middle, high\}           & \{0.4,0.1,0.35,0.15\}      \\
			Temperature         & \{very cold, cold, chilly, warm, hot\} & \{0.05, 0.15,0.4,0.3,0.1\} \\\
			Wind                & \{quiet, weak,medium, strong \}        & \{0.2,0.25,0.45,0.1\}      \\\
		\end{tabular}
	\end{center}
\end{table}
\Exa

\begin{figure}[ht]
	\centering
	\begin{minipage}{0.45\textwidth}
		\centering
		\includegraphics[width=0.8\linewidth]{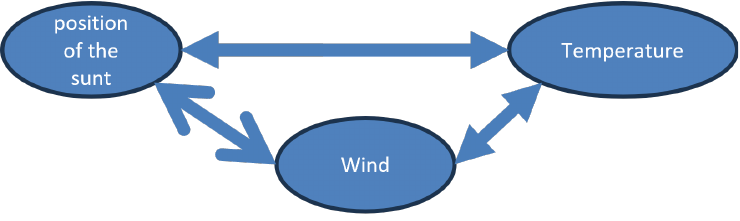}
		\caption{Interdependent probabilistic factors of a scene}
		\label{fig:GraphModelWeather}
	\end{minipage}\hfill
	\begin{minipage}{0.45\textwidth}
		\centering
		\includegraphics[width=0.8\linewidth]{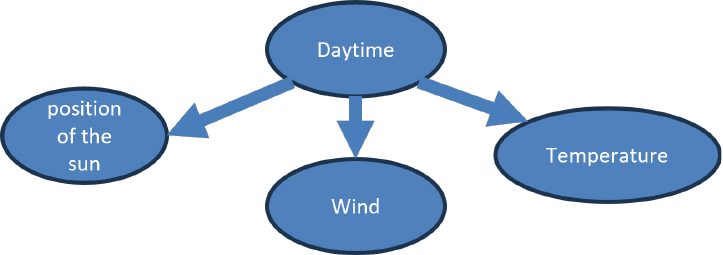}
		\caption{Resolution into an acyclic dependency graph by introducing a common influencing factor}
		\label{fig:GraphModelDaytime}
	\end{minipage}
\end{figure}

As in previous examples, the occurrence probabilities given here are merely exemplary.
When it is hot, there is usually little wind, and the sun is high in the sky.
The wind is typically stronger in the morning than at midday and calms down in the evening; and we expect cooler or warmer temperatures depending on the position of the sun.
The various factors are thus not independent -- but neither are they simply causally linked: their characteristics are probabilistically interwoven.
Modelling these dependencies can become quite complicated.
To obtain a qualitative overview of the various dependencies and influencing factors, the technique of probabilistic graphical modelling has proven itself, see \cite{Koller.2009, Pearl.b, Pearl.2014}.
The various probability spaces are represented as nodes of a graph, with arrows connecting dependent spaces; where the influence is unidirectional, the arrow indicates its direction.
In the present example, a clear direction of influence is hard to discern, see fig.~\ref{fig:GraphModelWeather}.
A closer look, however, reveals a common factor that exerts a substantial influence on all three: the time of day.
The scene transitions from darkness to dawn and from noon to dusk, with the noon period characterized by heightened luminosity and elevated temperatures; mornings, moreover, tend to be windier.
We therefore introduce an additional factor, the time of day, with options \{night, dawn, evening, afternoon, midday, morning, sunrise\} and exemplary probabilities \{0.4, 0.05, 0.1, 0.15, 0.1, 0.15, 0.05\}, respectively, see fig.~\ref{fig:GraphModelDaytime}.

The additional factor allows a dependency diagram to be drawn that is free of cyclic dependencies, offering a much clearer picture of the interdependence of the factors, see fig.~\ref{fig:GraphModelDaytime}.

Probabilistic graphical modelling is a well-proven technique for depicting probabilistic dependencies, see \cite{Koller.2009, Pearl.b, Pearl.2014}, and the introduction of common factors turns the models into acyclic, directed ones.

A second reason for examining the dependencies of the influencing factors is that they significantly reduce the number of required test cases.
Assume, for simplicity's sake, that the dependency between temperature and precipitation is neglected, as discussed in Section~\ref{sec:EnvConditions}.
To cover all potential scenarios combinatorially, we would then be obliged to include scenarios with snowfall at warm and at hot temperatures in the test cases!
Note that dependencies affect not only the combinatorics but also the required statistics: less frequent cases require less testing than normal situations.
In summary: although dependencies do not change the partitioning of the space that a PEON is to model statistically, they do change the statistical relationships on it.

Dependencies can thus both increase and reduce the required effort.
To capture them qualitatively, it is useful to create a probabilistic graphical model of the various factors, see \cite{Koller.2009, Pearl.2014}.
Such a graphical representation already permits in-depth analyses of complex dependency relationships.
In particular, it facilitates the identification of the essential dependencies, and may suggest common factors that turn the qualitative model into an acyclic one.
For an illustration, see Section~\ref{sec:EnvConditions}, where we introduced the supplementary factors season, temperature, and time of day to eliminate cyclic dependencies.

To model an ODD probabilistically, it is therefore advisable to first create a probabilistic graphical model of the essential dependencies:
the essentials are much easier to supplement and restructure, which greatly facilitates the modelling.
For the general theory and techniques, we refer to the excellent books \cite{Koller.2009, Pearl.2014}, which contain many examples and much further information.

\subsubsection{Coupling of measures}\label{sec:coupling}
As explained on several occasions, the marginal distributions of the individual factors involved in the probabilistic modelling of driving situations can usually be determined with little effort, see Sections \ref{sec:EnvConditions} and \ref{sec:Probability}.
The crux of the matter is the determination of their \emph{joint} distributions, i.e.\ the probabilities of specific characteristics of the factors occurring together.
Mathematically: we know the distributions $p_X, p_Y$ on $X$ and $Y$, respectively, but we do not know the distribution $p_{X\times Y}$ on $X\times Y$.
We are thus looking for a distribution $p$ on the Cartesian product $X\times Y$ with the given marginals.

\Def\label{lem:Couplings}
Let $(X,p_X)$ and $(Y,p_Y)$ be probability spaces, i.e.\ event spaces with probability distributions. A distribution $p_{X\times Y}$ on the product space $X\times Y$ is called a \textbf{coupling} of $p_X, p_Y$ if
\be
p_{X}(x) = \int_{y\in Y} p_{X\times Y}(x,y)\ \  \forall\, x\in X, \hspace{0.8cm} p_{Y}(y) = \int_{x\in X} p_{X\times Y}(x,y)\ \ \forall\, y\in Y,
\ee
that is, if it reproduces the specified marginal distributions.
\Defa

\Nb
For finite probability spaces with $\#\{X\}=n\in \N$ and $\#\{Y\}=m\in \N$ a simple count of the constraint equations and parameters shows that the set of couplings of two measures on spaces X, Y has dimension $(n-1)\cdot (m-1)$.

It is also noteworthy that the space of couplings is a convex subset.
\Ne

We can therefore reformulate our problem as follows: given the marginal distributions of the factors, we are looking for a suitable coupling of them, which then defines the distribution on the product space.

Note that the set of all couplings of given marginals is a convex set.

As already outlined in the section on Bayes' theorem, Section~\ref{sec:Probability}, we do not want to specify the distribution $p_{X\times Y}$ on the product space directly, but take the detour via conditional probabilities.

\begin{figure}[ht]
  \centering
  \begin{minipage}{0.45\textwidth}
    \centering
    \includegraphics[width=\textwidth]{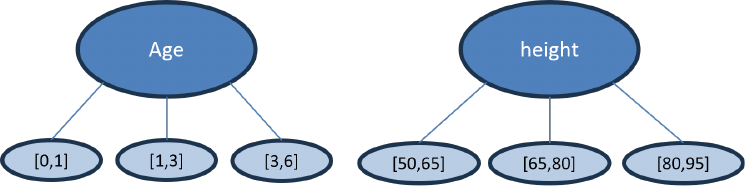}
    \caption{Event spaces 'Age' and 'Height' of Children}
    \label{fig:AgeHeight}
  \end{minipage}\hfill
  \begin{minipage}{0.45\textwidth}
    \centering
    \includegraphics[width=0.5\textwidth]{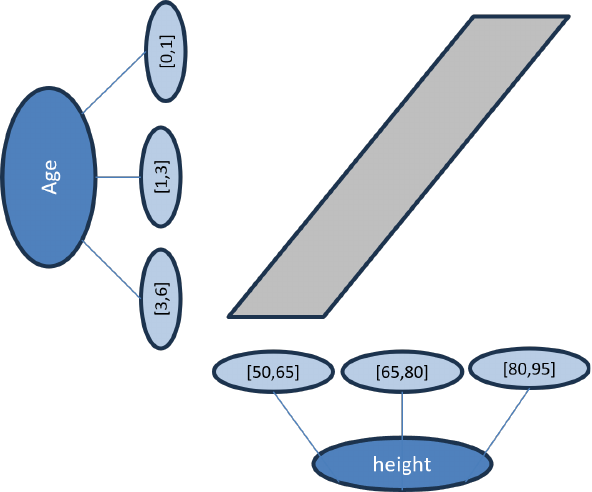}
    \caption{Linear dependency of 'Age' and 'Height'}
    \label{fig:AgeHeightLin}
  \end{minipage}
\end{figure}

\Ex \label{Ex:Children}
The objective of this example is to describe the age and height of children between 0 and 6 years of age, see fig.~\ref{fig:AgeHeight}.
The distributions of children's ages and of children's heights are comprehensively documented in data of the Federal Statistical Office.
What the data does not provide is the \emph{joint} distribution of age and height.
Empirically, however, growing up is accompanied by a corresponding increase in height, so it is reasonable to make statistical inferences about the relationship between age and height under the assumption that older children tend to be taller than younger ones.
This suggests anticipating a linear relationship between age and height: the essential support of the conditional probability should be close to a line $h\sim const.\cdot a$,
$$
  p_{Height,Age}(h | a) = \frac{p_{Height,Age}(h , a)}{p_{Age}(a)} \mbox{ has, for given age $a$, its support in $h$ close to $const.\cdot a$.}
$$
Consequently, the support of the coupling is predominantly concentrated around the diagonal, as illustrated by the gray bar in fig.~\ref{fig:AgeHeightLin}.

Informally: if we anticipate a linear relationship for the conditional probability of $h$ given $a$, the essential support of $p_{X\times Y}(a,h)$ is constrained to a neighbourhood of $\{(a, const.\cdot a)\}$ -- a stringent restriction on the coupling.
This suggests attacking the problem of defining the joint distribution $p_{X\times Y}$ by imposing suitable restrictions on the support of the coupling.
\Exa

Indeed, by the definition of the conditional probability, equ.~\ref{Eq:Bayes}, the statistical behaviour of the conditional probabilities can be inferred from the support properties of a coupling:
if $supp\  p_{Height,Age}(h , a)\sim \{(a,const.\cdot a)\}$, then
$$
  p_{Height,Age}(h | a) = \frac{p_{Height,Age}(h , a)}{p_{Age}(a)} \ \ \mbox{ i.e., given } a \Rightarrow h\sim const.\cdot a.
$$
In this manner, the support of a coupling encodes the statistical relationship between the factors -- and the same holds for the conditional probability.

This observation suggests formulating dependencies between factors informally via the conditional probability, i.e.\ via requirements on the support, and determining the joint distribution from these requirements together with the prescribed marginal distributions.

\subsubsection{Linear Couplings and the Transport Problem}\label{sec:TransportProblem}
As outlined in the preceding section (see Section~\ref{sec:coupling}), our objective is to pin down the support of a coupling by imposing suitable requirements on it.
We first look for ``optimal'' solutions -- couplings that satisfy the imposed support constraints as tightly as possible -- and we begin with couplings whose support is to lie as close as possible to the diagonal.

To ease the way into the subsequent considerations, it is worth starting once more with an example, following Villani \cite{Villani.2009}.

\Ex \label{Ex:TransportProblem}
Bakeries and bakery stores are distributed throughout a city.
Bread is produced in the bakeries and transported to the stores for sale, and the transportation is to be organized so that the associated costs are minimized (Monge's transport problem).
Denote by $p_{bakery}$ the distribution of the produced loaves over the bakeries and by $p_{bakery stores}$ the distribution of the loaves sold over the stores (both normalized by the total number of loaves).
We assume that, ideally, all bread finds a buyer.
An assignment of how many loaves of each bakery are transported to which store can be expressed as a matrix $n_{i,j}$, with $n_{i,j}$ the number of loaves produced in bakery $i$ and transported to the $j$-th store.
Normalizing the matrix by the total count of loaves yields $p_{\{bakery\},\{bakery store\}}(i,j) := n_{i,j}/n$, which is interpretable as a distribution over the Cartesian product of the sets of bakeries and stores.
The constraints that all bread is produced in some bakery and sold in some store state precisely that this distribution is a coupling of $p_{bakery}, p_{bakery store}$.
The transport problem thus amounts to identifying an optimal coupling with respect to the transport costs.

\begin{figure}[htbp]
  \centering
  \includegraphics[width=0.4\textwidth]{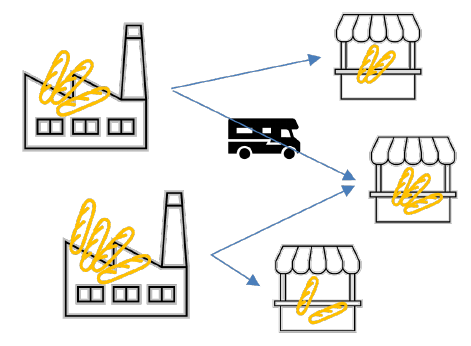}
  \caption{Transport Problem: Breads from Bakeries to Bakery Stores}
  \label{fig:TPBread}
\end{figure}

If the bakeries and the stores are lined up along two parallel lines, a solution is readily found by proceeding greedily from one end:
the lowest store is supplied by the lowest bakery; if that bakery has loaves left over, they go to the next store up, and if the store's demand is not yet met, it is supplied by the next bakery up -- and so on, matching supply and demand along the line.
See fig.~\ref{fig:TPBread} for a visual representation.

For a comprehensive study of the transport problem and its history, see \cite{Villani.2009}.
\Exa

Assume now, for simplicity, that the transportation cost of a loaf is proportional to the distance from the bakery to the store.
The transport problem of Example \ref{Ex:TransportProblem} can then be formulated mathematically as follows:
find a coupling $p_{bakery,\ bakery\ stores}$ that minimizes the cost
\be
\sum_{i,j} dist(i,j)\  p_{bakery,\ bakery\ stores}(i,j)\ \Big(\sim \sum_{i,j} dist(i,j)\  n_{i,j}\Big)
\ee
where $dist(i,j)$ denotes the distance of the $i$-th bakery to the $j$-th store.

\begin{figure}[ht]
  \centering
  \begin{minipage}{0.45\textwidth}
    \centering
    \includegraphics[width=\textwidth]{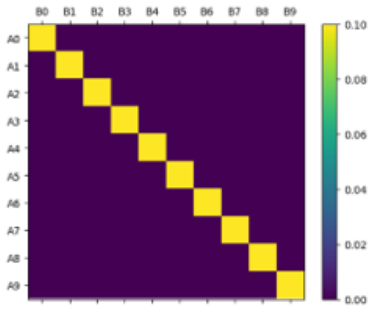}
    \caption{Optimal linear coupling over $10\times 10$ points}
    \label{fig:UniformTP}
  \end{minipage}\hfill
  \begin{minipage}{0.45\textwidth}
    \centering
    \includegraphics[width=\textwidth]{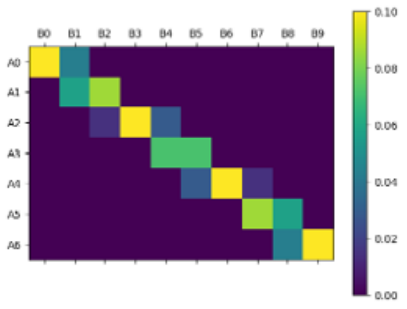}
    \caption{over $7\times 10$ points}
  \end{minipage}
\end{figure}

In Example \ref{Ex:TransportProblem}, a solution was constructed under the assumption of a linear ordering of bakeries and stores, and of equal production in bakeries and sales in stores.
It can be shown that optimal solutions always exist.
The Sinkhorn algorithm, see \cite{Peyre.}, provides a highly efficient numerical method for computing optimal couplings, and thus for solving the transport problem.
Fig.~\ref{fig:UniformTP} presents the matrix of the optimal linear coupling as a heat map; the marginal distributions are uniform over $10\times 10$ resp.\ $7\times 10$ points.
The optimality of these solutions is plainly visible.

Evidently, by choosing the distance function $dist(i,j)$ judiciously, a wide range of constraints can be imposed on the support.
Before delving into this, we introduce an alternative approach that yields general solutions to the coupling problem.

\subsubsection{Polynomial Couplings}\label{sec:PolyCouplings}
A closer look at the constraints a coupling must satisfy reveals that the marginal conditions can be met in a very general manner.
Consider two probability spaces $X$ and $Y$ with distributions $p_X$ and $p_Y$.
For finite sets, a coupling can always be represented by a measurable function $f$ on $X\times Y$ with

\bea
p_{X\times Y}(x,y) &=& (1+f(x,y)) \cdot p_X(x) \cdot p_Y(y), \ \  x\in X, \ y\in Y \nonumber \\
\mbox{with } 1+f(x,y) \geq 0\ \  \forall  x\in X, \ y\in Y  &\mbox{ and }& \ \int_X f(x,y) p_X(x)=\int_Y f(x,y) p_Y(y) =0
\ea
This representation makes it easy to identify simple solutions.
We index the finite elements, i.e.\ we embed the probability spaces into the real numbers and consider finite subsets of $\R$.
For example, let
\be\label{equ:LocNoise}
f(x,y) = \lambda \cdot (x-<x>_X)\cdot (y-<y>_Y)\ \ \ \mbox{with } <x>_X = \int_X x \cdot p_X(x), \ \ <y>_Y = \int_Y y \cdot p_Y(y),
\ee
with $\lambda$ chosen small enough in absolute value to guarantee $1+f(x,y) \geq 0\ \forall x \in X,\ y\in Y$.

This indeed defines a coupling:
the density is nonnegative by the choice of $\lambda$, and the centered factors integrate to zero, $\int_{X} (x-<x>_X)\, p_X =0$, and likewise for $Y$.
Hence
\be
\int_{X} f(x,y) \cdot p_X(x) \cdot p_Y(y) = 0 = \int_{Y} f(x,y) \cdot p_X(x) \cdot p_Y(y) = \int_{X\times Y} f(x,y) \cdot p_X(x) \cdot p_Y(y),
\ee
so all constraints required of a coupling are fulfilled by $(1+f(x,y)) \cdot p_X(x) \cdot p_Y(y)$.
(The product measure $p_X\times p_Y$ itself, i.e.\ $f\equiv 0$, is the trivial coupling.)

Evidently, the monomials may be replaced by any measurable functions in $x$ or $y$ -- for instance by $x^n-<x^n>_X$ with integer $n$.
This motivates the following definition of a family of couplings.

\paragraph{General Functional Couplings}\label{sec:FuncCoupl}
The polynomial construction above admits a far-reaching generalization, which we sketch only briefly.
Let $\mathcal{X}_i \subset \R$, $i \in \{1,\dots,m\}$, be probability spaces with measures $p_i$, and let $\mathcal{X} := \prod_{i=1}^m \mathcal{X}_i$ carry the product measure $p = \otimes_{i=1}^m p_i$.
The centered monomials $b_i^n(x) := x^n - \langle x^n \rangle_{p_i}$, $n \geq 1$, span the subspace of $L^2(\mathcal{X}_i, p_i)$ orthogonal to the constants, so that any coupling of the marginals $p_i$ can be written as
\be
p_{coupl} = \Big(1 + \sum_{n_1,\dots,n_m} \lambda_{n_1 \dots n_m}\, \omega_{n_1 \dots n_m}\Big)\, p,
\ee
where the $\omega_{n_1 \dots n_m}$ denote the Gram--Schmidt orthonormalization of the tensor products $b_1^{n_1} \otimes \cdots \otimes b_m^{n_m}$ with respect to $p$.
The polynomial couplings of section \ref{sec:PolyCouplings} are precisely the lowest-order truncations of this expansion.

Conversely, a desired shape of a coupling, given as a nonnegative, normalized density $g$ on $\mathcal{X}$ with the correct marginals, can be approximated by expanding $g$ in this system; the coefficients are obtained as the scalar products $\lambda_{n_1 \dots n_m} = \int_{\mathcal{X}} g\, \omega_{n_1 \dots n_m}\, dp$.

\subsection{Noise Coupling}\label{sec:NoiseModelling}
Dependencies are not always sharp, i.e.\ optimal in the sense of the transport problem of Section~\ref{sec:TransportProblem}; they may be ``smeared'', noisy.
Many types of noise are known -- thermal noise, white noise, and others -- and our objective is a model that accommodates noisy dependencies.
Consider again the dependency between the age and the height of a child, Example \ref{Ex:Children}: it is not strict or causal, but a tendency, and we want to model the linear relation in a noisy manner.
Before doing so, let us briefly discuss what ``noise'' should mean in this context.
The noise blurs the possible values, depending on the \textit{temperature}.
In other words, the higher the temperature, the taller younger children will be, and the shorter older children will be.
This noise should apply equally to all events.

These examples cover the salient categories of noise, which we will use exclusively for our modelling:
we describe ``global'' noise by the product measure of the distributions, and ``local'' noise as defined in \ref{def:ThermalNoise} below.

\Def \label{def:ThermalNoise}
Let $(X,p_X)$ and $(Y,p_Y)$ be finite ordinal probability spaces, $X=\{0,\dots,m\}$, $Y=\{0,\dots,n\}$, with strictly positive densities $p_X \in \mathbb{R}_{>0}^{X}, p_Y \in \mathbb{R}_{>0}^{Y}$.
For $\beta>0$ define the \textbf{thermal kernel}
\be
K_\beta(i,j) := \exp\!\big(-\beta\,(i-j)^2\big), \hspace{0.6cm} i\in X,\ j\in Y.
\ee
Since $K_\beta$ has strictly positive entries, Sinkhorn's matrix-scaling theorem \cite{Cuturi.c} guarantees the existence of positive vectors $u\in\mathbb{R}_{>0}^{X}$, $v\in\mathbb{R}_{>0}^{Y}$ -- unique up to the rescaling $(u,v)\mapsto(c\,u,\,v/c)$, $c>0$ -- such that
\be
p^{temp}_{X\times Y,\beta}(i,j) := u_i\,K_\beta(i,j)\,v_j
\ee
is a coupling of $p_X,p_Y$ in the sense of Def.~\ref{lem:Couplings}, i.e.\ $\sum_j p^{temp}_{X\times Y,\beta}(i,j)=p_{X,i}$ and $\sum_i p^{temp}_{X\times Y,\beta}(i,j)=p_{Y,j}$.
We call $p^{temp}_{X\times Y,\beta}$ the \textbf{temperature (thermal) noise coupling} of $p_X,p_Y$ at inverse temperature $\beta$.
\Defa

The scaling vectors $u,v$ are computed as the fixed point of the Sinkhorn / iterative-proportional-fitting recursion
\be \label{equ:Sinkhorn}
u \leftarrow p_X \oslash (K_\beta v), \hspace{0.6cm} v \leftarrow p_Y \oslash (K_\beta^{\!\top} u),
\ee
where $\oslash$ denotes entrywise division, started at $u^{(0)}=v^{(0)}=\1$; the iteration converges geometrically to $u,v$ up to the scaling ambiguity above, see \cite{Cuturi.c}.

\Nb
The parameter $\beta$ interpolates between two extreme cases: as $\beta\to\infty$, $K_\beta$ degenerates to the identity kernel and, for $X=Y$, $p^{temp}_{X\times Y,\beta}$ tends to the exact, noise-free matching $p(i,i)=p_X(i)$; as $\beta\to 0$, $K_\beta$ becomes constant and the Sinkhorn scaling of a constant kernel recovers exactly the product coupling $p_X\otimes p_Y$. Thus $p^{temp}_{X\times Y,\beta}$ reflects the physical picture of thermal noise for ordinal classes.
\Ne

The heat map of a linear noise coupling exhibits the expected symmetry, and its noisy character in the linear sense is plainly visible, see fig.~\ref{fig:NoiseCoupling} or fig.~\ref{fig:NoisyAbsXmin5}.

\begin{figure}[htbp]
  \centering
  \includegraphics[width=0.4\textwidth]{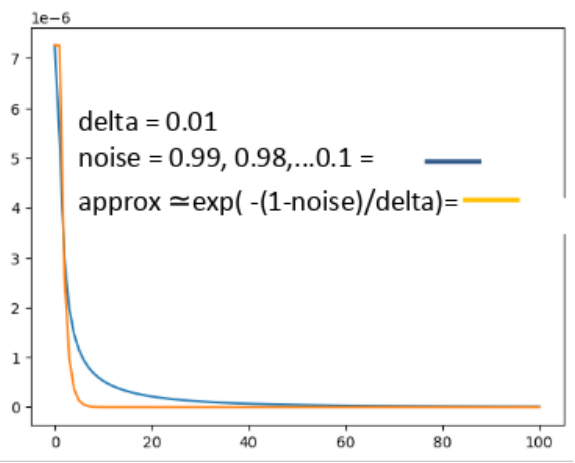}
  \caption{covariance of a linear coupling with noise is related to an exponential decrease}
  \label{fig:Cov2Noise}
\end{figure}

Unsurprisingly, the correlation coefficient in the presence of noise is lower than for the pure transport solution.
This also provides a way of determining a suitable thermal factor $\beta$ from experimental data:
the covariance of $p^{thermal}_{X\times Y,\beta}$ depends on the parameter $\beta$, so computing the correlation coefficients of $p^{lin}_{X\times Y }\ \ (\beta=1)$ and $p_X \cdot p_Y , \ \  (\beta=0)$ and matching them against the empirical correlation coefficient, or a statistical estimate thereof, determines the proper $\beta$, see figure \ref{fig:Cov2Noise}.

With these constructs, linear statistical dependencies with noise can be modelled very accurately.
In the present implementation, the rescaled quantity $\beta \rightarrow \frac{1}{1-\beta}$  is utilised, in conjunction with the parameter interval $[0, 1]$. In this context, $\beta$ equals 1 at zero temperature and 0 at infinite temperature, i.e. global noise.
\subsubsection{Reparametrization and Coupling}\label{sec:Reparam}
Many dependencies, however, are not simple linear relationships, see Section~\ref{sec:FuncDep}.
With specific choices of distance functions (Section~\ref{sec:TransportProblem}) we have a flexible way of modelling more complex dependencies -- but these are somewhat unwieldy when it comes to graphical modelling.

On the other hand, we have already indexed the various classes and characteristics, both to create linear dependencies and to describe their marginal distributions by lists.
This suggests considering more general indexings, or re-indexings, of the classes:
suppose two classifications (two factors) are connected in a simple linear way -- what does the coupling look like if we merely rearrange the classes?

Let us first recall some general facts about variable transformations and measurable functions on probability spaces.
The following is a standard result of measure theory, see \cite{Villani.2009}.
\Lem \label{lem:pushforward_measure}
\textbf{(pushforward measure)} Let $X,Y\subset \R^n, n\in \N$ be measurable spaces and $f:X\rightarrow Y$ measurable.
Then any measure $p_X$ on $X$ induces a unique measure $p_Y=f_{\#}\circ p_X$ on $Y$, defined by $p_Y (A) := p_X(f^{-1}(A))$ for all measurable $A\subset Y$ -- the \textbf{pushforward measure}.
\Lema

A second important construction is the pullback.
As shown in \cite{Bogachev.2007}, the theory can be developed in far greater generality; for our purposes, the finite case suffices.

\Lem \label{lem:pullback_coupling}
\textbf{(pullback coupling)} Let $X,Y,Z$ be finite probability spaces with measures $p_X, p_Z$ on $X,Z$, respectively.
Let $f:X\rightarrow Y$ be measurable, and equip $Y$ with the pushforward measure $p_Y=f_{\#}(p_X)$.
Let a coupling $p_{Y\times Z}$ of $p_Y, p_Z$ be given.
Then
\be
f^\#(p_{Y \times Z})(x , z) := \frac{p_X(x)}{f_{\#}(p_X) (f(x))} \cdot p_{Y\times Z}(f(x) , z) \ \ \ \forall \ \ x\in X, z\in Z
\ee
defines the density (with respect to the counting measure) of a coupling of $p_X, p_Z$ on $X \times Z$, called the \textbf{pullback coupling} along $f$.
\Lema

Note that $0 \leq p_X(\{x\}) \leq p_X(f^{-1}(f(x)))=f_{\#}(p_X) (f(x))$ for all $x\in X$, so on null sets of $f_{\#}(p_X)$ the numerator vanishes as well; with the convention $0/0=0$, the density is well defined.

\noindent \textbf{proof:}\hspace{0.1cm}
For every $y$ in the support of $f_\#(p_X)$ we have $\sum_{x:f(x)=y} p_X(x) = p_X(f^{-1}(y)) = f_{\#}(p_X)(y)$, i.e.\ the fibre weights $\frac{p_X(x)}{f_{\#}(p_X)(y)}$ sum to $1$.
For the first marginal we therefore obtain
\bea
\sum_x \frac{p_X(x)}{f_{\#}(p_X) (f(x))} \cdot p_{Y\times Z}(f(x) , z) &=&
\sum_y \sum_{x:f(x)=y}  \frac{p_X(x)}{f_{\#}(p_X) (y)} \cdot p_{Y\times Z}(y , z)
\nonumber \\
= \sum_y p_{Y\times Z}(y , z) \cdot \sum_{x:f(x)=y}  \frac{p_X(x)}{f_{\#}(p_X) (y)}
&=& \sum_y p_{Y\times Z}(y , z) = p_Z(z)
\ea
The second marginal property holds by construction:
$$
  \sum_{z\in Z} f^\#(p_{Y\times Z})(x , z) = \frac{p_X(x)}{f_{\#}(p_X) (f(x))} \cdot p_{Y\times Z}(\{f(x)\} \times Z) =  \frac{p_X(x)}{f_{\#}(p_X) (f(x))} \cdot f_{\#}(p_X) (f(x)) = p_X(x) \ \forall \ x\in X
$$ \hfill $\Box$

Sampling from the pullback coupling on $X\times Z$ proceeds in two steps:
first sample a pair $(y,z)$ according to $p_{Y\times Z}$, then sample $x$ within the fibre $f^{-1}(y)$ according to the multinomial weights $\frac{p_X(x)}{f_{\#}(p_X) (y)}$.
The construction generalizes as follows.

\Cor\label{cor:Reparam}
Let $X, p_X$ and $W,p_W$ be finite probability spaces and $f:X\rightarrow Y, g:W\rightarrow Z$ functions.
Then any coupling $p_{Y\times Z}$ of $f_\#(p_X),g_\#(p_W)$ can be pulled back along $f\times g$ to a coupling of $p_X, p_W$ with density
\be
(f\times g)^\#(p_{Y \times Z})(x,w) := \frac{p_X(x)}{f_{\#}(p_X) (f(x))} \cdot \frac{p_W(w)}{g_{\#}(p_W) (g(w))}\cdot p_{Y \times Z}(f(x),g(w)) \ \ \ \forall \ \ x\in X, w\in W
\ee
with respect to the counting measure.
\Cora

With these two results at hand, complex probabilistic dependencies can be formulated in a very general yet manageable way.

\begin{figure}[ht]
  \centering
  \begin{minipage}{0.45\textwidth}
    \centering
    \includegraphics[width=0.4\textwidth]{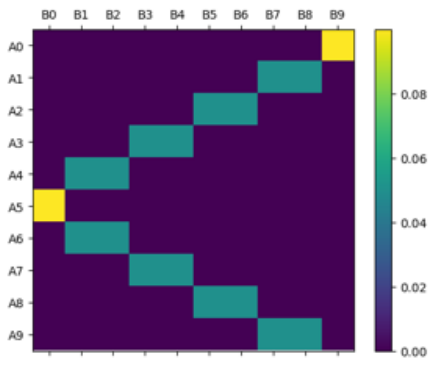}
    \caption{Functional Coupling with $x\mapsto abs(5-x)$}
    \label{fig:FuncConstr}
  \end{minipage}
  \begin{minipage}{0.45\textwidth}
    \centering
    \includegraphics[width=0.5\textwidth]{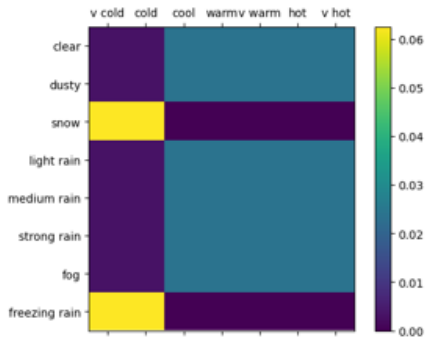}
    \caption{Coupling with logical constraint: if snow falls or freezing rain, the temperature must be cold!}
    \label{fig:LogConstr}
  \end{minipage}\hfill
\end{figure}

As a first, rudimentary application, suppose we want to model a functional dependency of the type
\be
p_{X\times Z}(z|x) \ \mbox{ concentrated near } \ z= abs(5-x).
\ee
Let $X$ and $Z$ be the set $\{0,1,\dots 9\}$ with the counting measure, and let $Y =\{0,1,\dots,5\}$.
Define $f:X\rightarrow Y, x\mapsto abs(5-x)$.
The pushforward construction, Lemma~\ref{lem:pushforward_measure}, yields a measure on $Y$; we take the optimal linear coupling on $Y\times Z$ for $f_\#(p_X), p_Z$ and pull it back to $X\times Z$, which defines the final coupling for $p_X,p_Z$ as stated in Lemma~\ref{lem:pullback_coupling}.
Fig.~\ref{fig:FuncConstr} shows the heat map of this construction: the essential support of the coupling is clearly the graph of the function $f$.

This example illustrates how, by strategically selecting a ``reparametrization'', complex couplings with functional dependencies can be modelled in a very general and accessible manner.

As a further salient example, let us reconsider logical constraints.
\Ex \label{ex:log_constr}
Think of temperature and precipitation: it only snows when it is cold.
This is not a linear constraint but a Boolean one, imposed on the combination of the temperature and precipitation distributions. We use a simplified version.
\begin{table}[h!]
  \begin{center}

    \caption{temperature, precipitation}
    \label{tab:precipitation_temp}
    \begin{tabular}{l|c|r}
      \textbf{Factor} & \textbf{Values}                                                                   & \textbf{Probabilities} \\
      \hline
      Temperature     & \{very cold, cold, cool, warm, very warm, hot, very hot\}                         & \{0.1,0.1,\dots\}      \\\
      precipitation   & \{clear, dusty, snow, light rain, medium rain, strong rain, fog, freezing rain \} & \{0.1,0.1,\dots\}      \\\\
    \end{tabular}
  \end{center}
\end{table}

If it is snowing or there is freezing rain, it must be cold or very cold!
The resulting coupling should look like the one in fig.~\ref{fig:LogConstr}.

\Exa

This can be modelled through a simple reparametrisation and utilisation of Corollar \ref{cor:Reparam}.
Let $X=\{Temperature\}, W=\{precitipation\}$ and $Y=\{Y_{cold},Y_{warm}\}, Z=\{Z_{snow}, Z_{not freezing}\}$

\bea
f: X  \rightarrow  Y &\hspace{1cm} &g:W  \rightarrow  Z \nonumber \\
f(\mbox{very cold})= f(\mbox{cold})  =&  Y_{cold} \hspace{1cm} g(\mbox{snow})&=g(\mbox{freezing rain}) = Z_{snow} \nonumber\\
f(\mbox{warm})= f(\mbox{warm}) \cdots =&  Y_{warm} \hspace{1cm} g(\mbox{clear})&=g(\mbox{dusty}) \cdots= Z_{not freezing}
\ea

These function define pushforward measures on $Y$ and $Z$. It can be demonstrated that, by retracting the linear coupling of the push-forward measures, that is to say, by applying Corollary \ref{cor:Reparam}, the coupling displayed in Figure fig.~\ref{fig:LogConstr} is obtained.

There is a general mechanism behind that construction of logical constraint couplings which will be shown next.

\paragraph{Equivalence Relation}\label{sec:EquivRel}
It is a fundamental observation in mathematics that equal behaviour under certain aspects establishes an equivalence relation on a given set; associated with it is the natural projection onto the equivalence classes,
\bea
X = \{class_1,\dots,class_n\} \mbox{ with equivalence relation } \sim \ \ &\Rightarrow \nonumber \\
Proj_\sim : X \rightarrow X/\sim
\ea
Given a probability distribution $p_X$ on the classification set $X$, we obtain the pushforward measure
\be
(Proj_\sim)_\# (p_X) \mbox{ on } X/\sim,
\ee
which simply restates that the probability of a partition class is the sum of the probabilities of its elements.

Let now a second equivalence relation be given on a set $W$ with measure $p_W$.
Then we have pushforward measures on $X/\sim$ and $W/\sim$, and any coupling of these can be pulled back via the projections to a coupling on $X\times W$ with respect to $p_X,p_W$.

Sampling on $X\times W$ then proceeds by first sampling the pair of partition classes according to the coupling, and then sampling within the classes, weighted by $\frac{p_X(x)}{\sum_{x\sim x'}p_X(x')}$, and similarly for $w$.
With this toolbox, Boolean boundary conditions for couplings are now easily implemented.

\Nb \label{sec:LogConstr}
Let $A\subset X, B\subset Z$ and distributions $p_X, p_Z$ on finite spaces $X,Z$ be given. Define
\be
\chi_A:X \rightarrow \{0,1\}, \ \mbox{ the characteristic function of A, i.e. } \chi_A(x)=1 \ \mbox{ iff } x\in A
\ee
and analogously $\chi_B$. With Lemma~\ref{lem:pushforward_measure} we obtain two pushforward measures $\chi_{A_{\#}}\circ p_X, \chi_{B_{\#}}\circ p_Z$ on $\{0,1\}$.
Let $p^{TP}_{\{0,1\}\times \{0,1\}}$ be the solution of the transport problem for these two measures with respect to the discrete distance on $\{0,1\}$ (generically the unique optimal coupling).
Via Lemma~\ref{lem:pullback_coupling} we then define
\be
p_{X,Z}^{A\rightarrow B}:= (\chi_A,\chi_B)^\# (p^{TP}_{\{0,1\}\times \{0,1\}})
\ee
as the pullback coupling of $p^{TP}$. One easily checks that this yields an optimal solution among the couplings for the Boolean constraint $x\in A \Rightarrow z \in B$, see fig.~\ref{fig:LogConstr}.

An even simpler, equivalent approach is to define
\be
dist(x,z) =1 \mbox{ iff } (x\in A,\  z\notin B) \mbox{ or } (x\notin A,\  z\in B), \mbox{ else } 0
\ee
and to take the corresponding transport solution directly on $X\times Z$; showing the equivalence is a simple exercise in probability theory.

The construction extends immediately to a list of disjoint Boolean constraints.
Let $A_i \subset X, B_i\subset Z$ be disjoint subsets of $X$ and $Z$, respectively, $i=1,2\dots k$, and let the constraints be $A_i \mapsto B_i$. Define
\be
\chi_X(x) := i \mbox{ iff } x\in A_i, \mbox{ else }  0
\ee
and similarly $\chi_Z$.
Consider the coupling $p_{lin}$ on $\{0,1,\dots k\}^2$ of the measures $\chi_{X\#}(p_X)$ and $\chi_{Z\#}(p_Z)$ with minimal support close to the diagonal.
The pullback $(\chi_X\times \chi_Z)^\# (p_{lin})$ then models these constraints probabilistically.
\Ne

\Nb \label{sec:GenerelizedConstruction}
Evidently, more complex conditions can be defined in general by choosing appropriate mappings or, better, distances.
As an illustration, consider a situation analogous to Remark~\ref{sec:LogConstr}, except that the $B_i$ are not disjoint -- in the simplest case $k=2$ with $B_2\ne B_1$.
In this case we take the minimal solution of the transport problem with the distance function
\bea
dist(x,z) &=0.5 & \mbox{ if } (x\in A_1,\  z\in B_1) \mbox{ or } (x\in A_2,\  z\in B_2) \nonumber \\
&=1 & \mbox{ if } (x\in A_1\cup A_2,\  z\notin B_1 \cup B_2) \mbox{ or } (x\notin A_1 \cup A_2,\  z\in B_1 \cup B_2) \nonumber \\
&=0 & \mbox{ else }
\ea
In this way, the modelling of such constraints becomes viable as well.
\Ne

\subsection{Iterative Construction of Complex Couplings}
In most cases, we are dealing not with couplings of just a few probability distributions, but with distributions on complex, interconnected classification spaces.
The following lemma is very useful for constructing such distributions.

\Lem \label{lem:IterativeCoupling}
Let $x$, $y$, and $z$ be random variables with distributions $p_X$, $p_Y$, and $p_Z$ on the probability spaces $X$, $Y$, and $Z$, and let $p_{X\times Z}$ and $p_{X\times Y}$ be couplings of the respective marginal distributions.
Then
\be
p(x,y,z):= \frac{p_{X\times Z}(x,z)\cdot p_{X\times Y}(x,y)}{p_X(x)}\hspace{0.5cm} \forall x\in X,y\in Y,z\in Z
\ee
(with the convention $0/0=0$) defines a coupling on $X\times Y \times Z$ of the marginal distributions $p_X, p_Y, p_Z$.
\Lema

\noindent \textbf{proof:}\hspace{0.1cm}
Since $p_{X\times Y}$ is a coupling, we always have $\frac{ p_{X\times Y}(x,y)}{p_X(x)} \geq 0$, and thus $p(x,y,z) \geq 0\ \ \forall x\in X,y\in Y,z\in Z$; note also that $p_X(x)=0$ forces $p_{X\times Y}(x,y)=0$, so the convention $0/0=0$ is consistent.
From the coupling property $\int_Y p_{X\times Y}(x,y) = p_X(x)$ -- and similarly for $p_{X\times Z}$ -- it follows that
\be
\int_{X\times Y} p(x,y,z) = \int_X \int_Y p(x,y,z) = \int_X \frac{p_{X\times Z}(x,z)\cdot p_{X}(x)}{p_X(x)} = p_Z(z) \ \ \forall z \in Z
\ee
Similarly, $\int_{Y\times Z} p(x,y,z) = p_X(x)$ for all $x \in X$.
This leaves
\be
\int_{X\times Z} p(x,y,z) = \int_X \int_Z \frac{p_{X\times Z}(x,z)\cdot p_{X\times Y}(x,y)}{p_X(x)} = \int_X \frac{p_X(x) \cdot p_{X\times Y}(x,y)}{p_X(x)} = p_Y(y) \ \ \forall y \in Y
\ee
The normalization then follows from the coupling properties. \hfill $\Box$

\subsubsection{Reconstructing Couplings}\label{sec:Reconstruction}
How can we proceed if only the structural part of a PEON is given?
That is: a forest with probability spaces at the nodes and dependency relations, given explicitly or implicitly -- but without the marginal distributions and without any noise parameters.
Suppose we have random samples from the underlying domain.
Then classical methods yield the marginal distributions together with estimates of their quality: all the tools of descriptive and empirical statistics are available to obtain valid distributions for the individual nodes, see \cite{Bijma.2017, Haslwanter.2022}.
What is still missing are the coefficients of the noise.

To this end, the covariance of the factors in question is estimated from the samples, along with its significance, using methods such as the $\chi^2$ test.
Assuming a linear statistical dependence between the factors, a suitable linear coupling together with its noise parameter $\beta$ can then be determined from the data, see Section~\ref{sec:NoiseModelling}:
the covariances of the optimal transport coupling and of the noise coupling differ in general, and since the covariance of $p^{\beta}$ depends on $\beta$, matching it against the empirical covariance yields an estimate of the noise parameter.
In this way, a set of samples can be represented and analyzed transparently via the reconstruction of a PEON as sketched above.

Conversely, suppose a PEON for an ODD has been designed, and the quality of a set of ODD samples is to be evaluated.
Using descriptive statistics on a sufficiently large sample, we can check whether the marginal distributions, variances, and covariances of the various factors agree with the estimates from the PEON.
This provides a development-independent evaluation of the balance of the examples -- exactly what is needed for their use as training data for an AI, see section \ref{sec:TestTD}.

\subsubsection{Refinements of PEONs}\label{sec:Refinements}
In some situations, a set of labeled data is given and is to be analyzed and visualized by means of a PEON.
How can this be done?

Suppose we already possess a mature, exhaustive PEON $P_{ODD}$ that can serve as a reference.
The labelling, however, will typically not be comprehensive enough, and the data will not suffice to reconstruct a PEON for the full structural part of $P_{ODD}$, i.e.\ to recover its marginal distributions and noise parameters.

The natural first step is to organize the labels within a PEON of their own:
we group suitable kinds of labels into the classes of classifications, guided by the classifications of $P_{ODD}$, and arrange these classifications into a -- rather coarse -- probabilistic graphical model, see Section~\ref{sec:StructProb}.
We do not add any dependencies.
The resulting PEON is modelled after $P_{ODD}$, but will usually be much coarser.
We can then attempt to reconstruct this coarse PEON from the labeled data, as in Section~\ref{sec:Reconstruction}, arriving at a further PEON, $P_{data}$.

The pivotal element of this construction is the aggregation of the labels into classifications \emph{according to the classifications of $P_{ODD}$} -- in other words, the partition classes of $P_{ODD}$ refine the groupings.
(Note that dropping all dependencies does not alter the partitioning: although the dependencies of $P_{ODD}$ are discarded in $P_{data}$, what matters here is the partitioning of the ODD, respectively of the labeling set, see Section~\ref{sec:StructProb}.)
Under this assumption, the following lemma applies.

\Lem \label{lem:Refinement}
Let $P_{ODD}$ be a PEON and let $\mathcal{L}$ be a set of labellings that can be grouped such that the groups are refined by the partition classes of $P_{ODD}$.
Then the probability distribution of $P_{ODD}$ defines a coarse-grained PEON $P_{\mathcal{L}}$ for $\mathcal{L}$.
\Lema

\noindent \textbf{proof:}\hspace{0.1cm}
The procedure sketched above shows how the coarse-grained PEON is obtained structurally.
Let $\mathcal{P}_{\mathcal{L}}$ denote the partitioning belonging to $\mathcal{L}$, and $\mathcal{P}_{ODD}$ the one belonging to $P_{ODD}$.
The refinement assumption states
\be
s\in \mathcal{P}_{\mathcal{L}} \Rightarrow s = \cup_{i} s_i \hspace{0.5 cm} \mbox{with suitable } s_i \in \mathcal{P}_{ODD}
\ee
Defining
\be
p_{\mathcal{L}}(s) := \sum_{s_i \subset s,\ s_i \in \mathcal{P}_{ODD} } p_{ODD}(s_i),
\ee
where $p_{\mathcal{L}}$, $p_{ODD}$ denote the probability distributions on the respective partitionings, yields a probability distribution on $\mathcal{P}_{\mathcal{L}}$.
Since all dependencies have been discarded, only marginal distributions need to be specified in the coarse PEON, and these are obtained by summing over the partition classes as above.
We thus obtain a unique coarse-grained reference PEON for $\mathcal{L}$, denoted $P_{\mathcal{L}}$.

For the $\sigma$-algebras we evidently have the relation $\Sigma_{\mathcal{L}}\subset \Sigma_{ODD}$, which gives rise to the Radon--Nikodym derivative
\be \label{eq:Radon-Nikodym}
\Delta(\frac{p_{ODD}}{p_{\mathcal{L}}})(s) = \frac{p_{ODD}(s)}{\sum_{s_i \in \mathcal{P}_{ODD},\ s_i \equiv_{\mathcal{L}} s} p_{ODD}(s_i)}
\ee
where $s_i \equiv_{\mathcal{L}} s$ indicates that $s$ and $s_i$ belong to the same partition class over $\mathcal{L}$.

Obviously $\Delta(\frac{p_{ODD}}{p_{\mathcal{L}}})(s) \geq 0$; and under our standing assumption that all partition classes have occurrence probability $> 0$, the Radon--Nikodym derivative is even strictly positive.
\hfill $\Box$

\Nb
Let $P_{ODD}$ be a refinement of $P_{\mathcal{L}}$, with $\mathcal{L}$ a labelling set as above.
The natural projection
\bea
\iota : \mathcal{P}_{ODD}&\rightarrow & \mathcal{P}_{\mathcal{L}} \nonumber \\
s' &\mapsto & s, \ \ \ \  s'\subset s
\ea
is measurable, and the induced measure on $\mathcal{P}_{\mathcal{L}}$ is precisely the pushforward measure $p_{\mathcal{L}} = \iota_{\#} \circ p_{ODD}$, see Lemma~\ref{lem:pushforward_measure}.
Even if the labeling set $\mathcal{L}$ is not refined by the partition set of the PEON $P_{ODD}$, but merely related to it by a mapping $f: \mathcal{P}_{ODD} \rightarrow \mathcal{P}_{\mathcal{L}}$, we still obtain a probabilistic extension of $\mathcal{L}$ via the pushforward construction.
\Ne

In such situations, the expectation values of functions over the partitioning of $\mathcal{L}$ are easily related via the substitution rule (change of variables).

Let $T$ be a function on the partitioning of $\mathcal{L}$ -- for example, the mean test result per common labelling.
Denote by $P_{\mathcal{L}}$ the coarse-grained PEON of a refinement $P_{ODD}$.
We extend $T$ to the partitioning of $P_{ODD}$ by setting $\hat{T}(s') := T(s)$ iff $f(s')=s$, for all $s'\in\mathcal{P}_{ODD}$.
Then the expectations with respect to the two PEONs are related by
\bea \label{eq:ExpectationRefinement}
<\hat{T}>_{p_{ODD}}&:=& \sum_{\mathcal{P}_{ODD}} \hat{T}(s')\cdot p_{ODD}(s') \nonumber \\
&=& \sum_{\mathcal{P}_\mathcal{L}} T(s)\cdot f_{\#} \circ p_{ODD}(s) =: <T>_{p_{\mathcal{L}}}
\ea
The equality $<\hat{T}>_{p_{ODD}} = <T>_{p_{\mathcal{L}}}$ follows by grouping the sum over the fibres of $f$.
The construction thus ensures consistent expectation values across the ontologies; in particular, quality statements can be transferred directly, i.e.\ related to the mature PEON.

It can thus be concluded that statistical criteria, such as the accuracy or precision of test results in relation to the labelling,$\mathcal{L}$ can be suitably reweighted to obtain the characteristic numbers in relation to the matured PEON $P_{ODD}$.

\subsubsection{Comparing two PEONs over a common ontology}\label{sec:2PEONs2Onts}
We now consider two distinct occurrence probability distributions over one and the same labeling ontology:
one may be defined by the pushforward construction of a refinement, the other by a reconstruction from training data as outlined in Section~\ref{sec:Refinements}.
We assume that the set of partition classes that never occur is the same in both cases -- mathematically, that the null sets of the two distributions coincide, so that the distributions are mutually absolutely continuous.
Then a unique Radon--Nikodym derivative exists; for a partition class it is computed as the quotient of its occurrence probabilities under the two distributions, and expectation values with respect to the two distributions are related by appropriate weighting with this derivative.
Broken down to the labeling sets, this reads as follows.

We employ a slightly more general setting.
Let a labeling set $\mathcal{L}$ be given, together with two different probabilistic extensions, i.e.\ two PEONs -- for instance $P_{data}$ and the pushforward of $P_{ODD}$, see Section~\ref{sec:TestCorrection}.
Denote by $p_{\mathcal{L}}, \hat{p}_{\mathcal{L}}$ the two measures.
Assume that $p_{\mathcal{L}}$ is absolutely continuous with respect to $\hat{p}_{\mathcal{L}}$ as measures on the labeling set:
if a partition class cannot occur with respect to $\hat{p}_{\mathcal{L}}$, i.e.\ its occurrence probability is $\hat{p}_{\mathcal{L}}(s)=0$, then it cannot occur with respect to $p_{\mathcal{L}}$ either.
As shown in Section~\ref{sec:TestCorrection}, this holds when $p_{\mathcal{L}}$ is the distribution of the test data and $\hat{p}_{\mathcal{L}}$ the pushforward measure of $P_{ODD}$, provided the underlying PEON is complete.
Then the Radon--Nikodym derivative
\be
\Delta \frac{p_{\mathcal{L}}}{\hat{p}_{\mathcal{L}}}(s) \geq 0 \ \ \forall s \in \mathcal{P}_{\mathcal{L}}
\ee
is well defined, with $\mathcal{P}_{\mathcal{L}}$ the set of all partition classes of $\mathcal{L}$;
it is strictly positive if, in addition, $\hat{p}_{\mathcal{L}}$ is absolutely continuous with respect to $p_{\mathcal{L}}$, in which case $\Delta \frac{\hat{p}_{\mathcal{L}}}{p_{\mathcal{L}}}(s) = (\Delta \frac{p_{\mathcal{L}}}{\hat{p}_{\mathcal{L}}}(s))^{-1}$.

\Lem\label{lem:Jacobian}
Let $p_{\mathcal{L}}, \hat{p}_{\mathcal{L}}$ be two measures on the set of partition classes $\mathcal{P}_{\mathcal{L}}$ of a labeling set $\mathcal{L}$.
Let $p_{\mathcal{L}}$ be absolutely continuous with respect to $\hat{p}_{\mathcal{L}}$, and let $T:\mathcal{P}_{\mathcal{L}} \rightarrow \R$ be any real-valued function. Then
\bea
<T>_{p_{\mathcal{L}}} &:=& \sum_{s\in \mathcal{P}_{\mathcal{L}}} T(s)\cdot p_{\mathcal{L}}(s) \nonumber \\
&=& \sum_{s\in \mathcal{P}_{\mathcal{L}}} T(s)\cdot \Delta \frac{p_{\mathcal{L}}}{\hat{p}_{\mathcal{L}}}(s) \cdot \hat{p}_{\mathcal{L}}(s) = <T\cdot \Delta \frac{p_{\mathcal{L}}}{\hat{p}_{\mathcal{L}}} >_{\hat{p}_{\mathcal{L}}}
\ea
The expectation value of $T$ with respect to $p_{\mathcal{L}}$ can thus be computed as a weighted sum of the function values over $\hat{p}_{\mathcal{L}}$, with weights given by the Radon--Nikodym derivative -- which, as shown above, can be computed explicitly (substitution rule).
\Lema
In particular, Lemma~\ref{lem:Jacobian} allows quality criteria stated for known labeling sets to be converted to a mature PEON.

\subsubsection{Closing remarks}
Most of the facts collected in this chapter carry over to more general, non-finite and non-discrete probability spaces.
Doing so, however, requires the sophisticated machinery of measure theory -- existence of measurable mappings, and considerably more involved proofs.
Since this paper is concerned exclusively with the discrete modelling of ODDs, we have refrained from that level of generality.


\end{document}